\documentclass{article}

\usepackage{epsfig}
\usepackage{graphics}
\usepackage{graphicx}

\usepackage{url}
\usepackage{subfigure}
\usepackage{hyperref}
\usepackage{multirow}

\usepackage{booktabs}
\usepackage{amsmath}

\usepackage{xcolor}
\usepackage{listings}

\title{Computing with Modular Robots}

\author{Genaro J. Mart{\'i}nez$^{1,2}$, Andrew Adamatzky$^{2}$, Ricardo Q. Figueroa$^{1}$, \\ Eric Schweikardt$^{3}$, Dmitry A. Zaitsev$^{4}$, Ivan Zelinka$^{5}$, \\ Luz N. Oliva-Moreno$^{6}$}

\begin{document}

\maketitle

\begin{centering}
$^1$ Artificial Life Robotics Laboratory, Escuela Superior de C\'omputo, Instituto Polit\'ecnico Nacional, M\'exico. \\
\url{gjuarezm@ipn.mx}, \url{qf7.ricardo@gmail.com } \\
$^2$ Unconventional Computing Laboratory, University of the West of England, Bristol, United Kingdom. \\
\url{andrew.adamatzky@uwe.ac.uk} \\
$^3$ Modular Robotics, Denver, United States of America. \\
\url{eric@modrobotics.com} \\
$^4$ Odessa State Environmental University, Odessa, Ukraine. \\
\url{daze@acm.org} \\
$^5$ Fakulta Elektrotechniky a Informatiky, Technick\'a Univerzita Ostrava, Czechia. \\
\url{ivan.zelinka@vsb.cz} \\
$^6$ Unidad Profesional Interdisciplinaria de Ingenier{\'i}a Campus Hidalgo, Instituto Polit\'ecnico Nacional,  M\'exico. \\
\url{loliva@ipn.mx } \\
\end{centering}

\begin{abstract}
\noindent
Propagating patterns are used to transfer and process information in chemical and physical prototypes of unconventional computing devices. Logical values are represented by fronts of traveling diffusive, trigger or phase waves. We apply this concept of pattern based computation to develop experimental prototypes of computing circuits implemented in small modular robots. In the experimental prototypes the modular robots Cubelets are concatenated into channels and junction. The structures developed by Cubelets propagate signals in parallel and asynchronously. The approach is illustrated with a working circuit of a one-bit full adder. Complementarily a formalization of these constructions are developed across Sleptsov nets. Finally, a perspective to swarm dynamics is discussed.

\vspace{3mm}

\noindent
\emph{Keywords:} robotics, unconventional computing, cellular automata, competing patterns, Cubelets, binary adder, Sleptsov net, networks.
\end{abstract}

\noindent{\footnotesize \it International Journal of Unconventional Computing, 17(1-2), 31-60, 2022. \url{https://www.oldcitypublishing.com/journals/ijuc-home/ijuc-issue-contents/ijuc-volume-17-number-1-2-2022/}}

\newpage

\section{Introduction}

One of the aims of the unconventional computing is to uncover novel computing substrates and protocols of computation in these~\cite{adamatzky2016advances}. Not necessarily the substrates and protocols should be efficient or optimal but they might open our horizons of understanding on how chemical and physical system process information and how a computation can be embedded into novel smart materials and intelligent structures.  The unconventional computation can be applied in the intelligent structures --- modular designs of buildings where a computation is implemented by building blocks, their ensembles and parts of the building~\cite{OnBuildings}. The smart building structures do echo up to some degree von Neumann original ideas on {\it machines constructing machines}~\cite{neumann1966theory}.

In experiments we used modular robots Cubelets. These robots are flexible, modular and easy to assemble and reconfigure~\cite{schweikardt2008learning,schweikardt2011modular}. The robots are capable for transmitting information between each other and performing a wide range of actions. The contribution in this paper is to make the robotic structures to implement computation by using a framework of computation by propagating and competing patterns. Implementation of Boolean gates with propagating patterns have been explored before in a context of precipitating chemical computers~\cite{RDC}, slime mould and plant gates~\cite{adamatzkyAdvPhysarum,adamatzky2017plantgates}, am gate implemented with swarms of soldier crabs~\cite{gunji2011}. The competing patterns were first introduced in \cite{martinez2010majority, martinez2010computation} in a model of a chaotic Life-like cellular automaton. The propagating patterns are used to represent binary information, and the computation is implemented by the patterns at the meeting sites. In the experimental setups presented concatenations of Cubelets represent communication channels --- where patterns propagate, and the patterns are represented by illumination of the robots. As a complementary contribution in parallel study we simulate Cubelets robots by Sletptsov nets \cite{SleptsovNet} to verify their behavior and formalism. Finally, we explore some aspect where Cubelets behavior could walk to global swarm phenomena \cite{wanka2019swarm}.

\section{Modular robots Cubelets}

Cubelets are modular robots~\cite{schweikardt2008learning,schweikardt2011modular}. Theoretically, we can have an infinite number of robots which can be constructed. Therefore we can say that Cubelets define a formal language as follows.

\begin{table}
\small
\caption{Classes of Cubelet robots.}
\centering
\begin{tabular}{ |l|l|l| }
\hline
\multicolumn{3}{ |c| }{Cubelets} \\
\hline
\multirow{4}{*}{sense} & distance & $di$ \\
 & brightness & $br$ \\
 & knob & $kn$ \\
 & temperature & $te$ \\ \hline
\multirow{8}{*}{think} & battery & $ba$ \\
 & Bluetooth & $bl$ \\
 & passive & $pa$ \\
 & blocker & $bo$ \\
 & inverse & $in$ \\
 & minimum & $mi$ \\
 & maximum & $ma$ \\
 & threshold & $th$ \\ \hline
\multirow{5}{*}{action} & rotate & $ro$ \\
 & drive & $dr$ \\
 & bar graph & $bg$ \\
 & speaker & $sp$ \\
 & flashlight & $fl$ \\
\hline
\end{tabular}
\label{symbolCubeletsTable}
\end{table}

\begin{figure}
\centering
\includegraphics[width=0.85\textwidth]{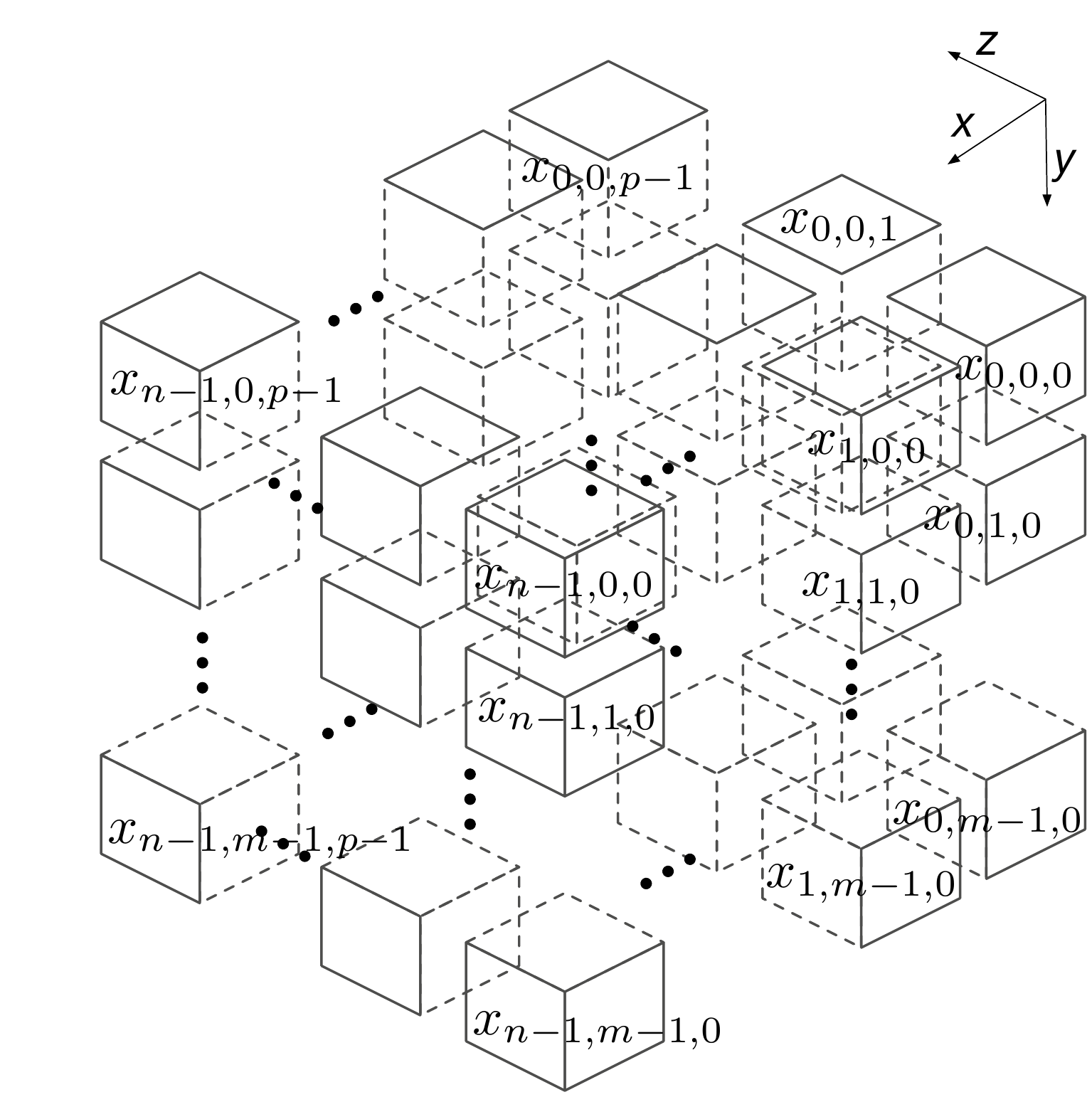}
\caption{Three-dimensional array that is represented as a string.}
\label{3D-diagram}
\end{figure}

\begin{figure}
\begin{center}
\subfigure[]{\scalebox{0.46}{\includegraphics{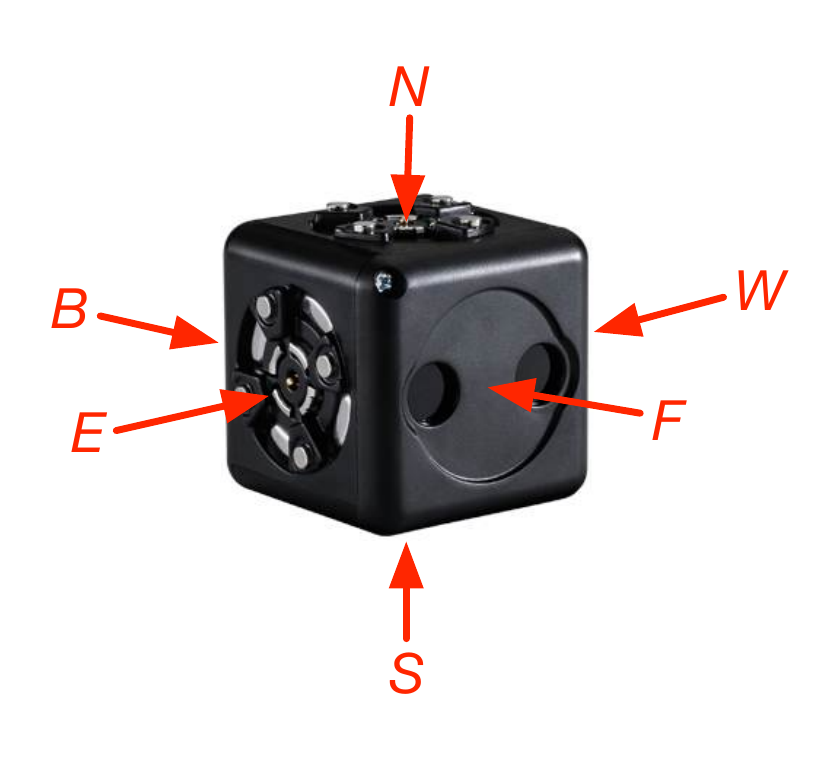}}} 
\subfigure[]{\scalebox{0.46}{\includegraphics{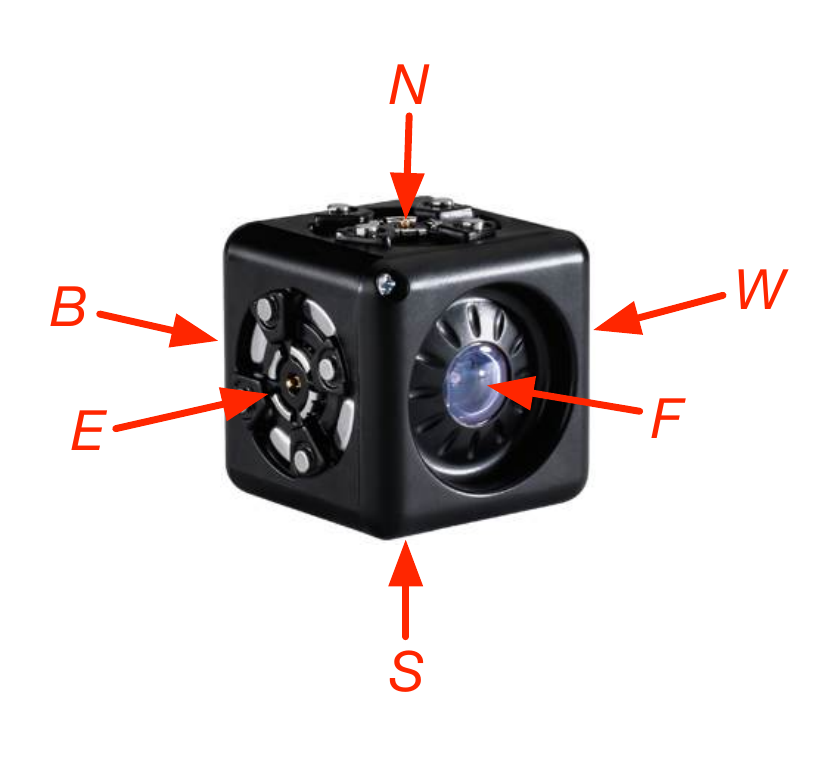}}} 
\subfigure[]{\scalebox{0.46}{\includegraphics{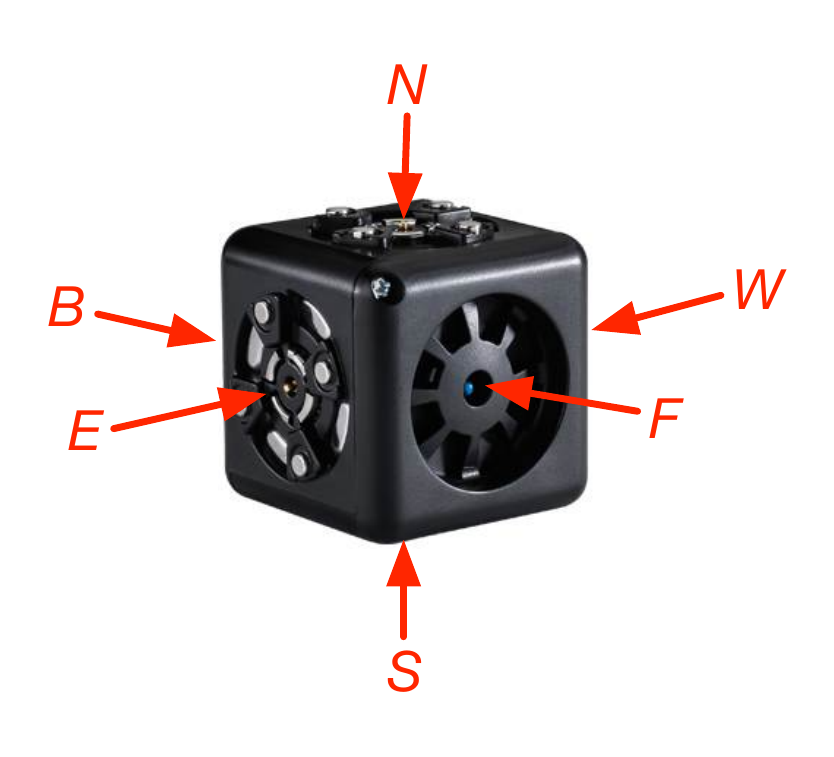}}} \hspace{1.2cm}
\subfigure[]{\scalebox{0.46}{\includegraphics{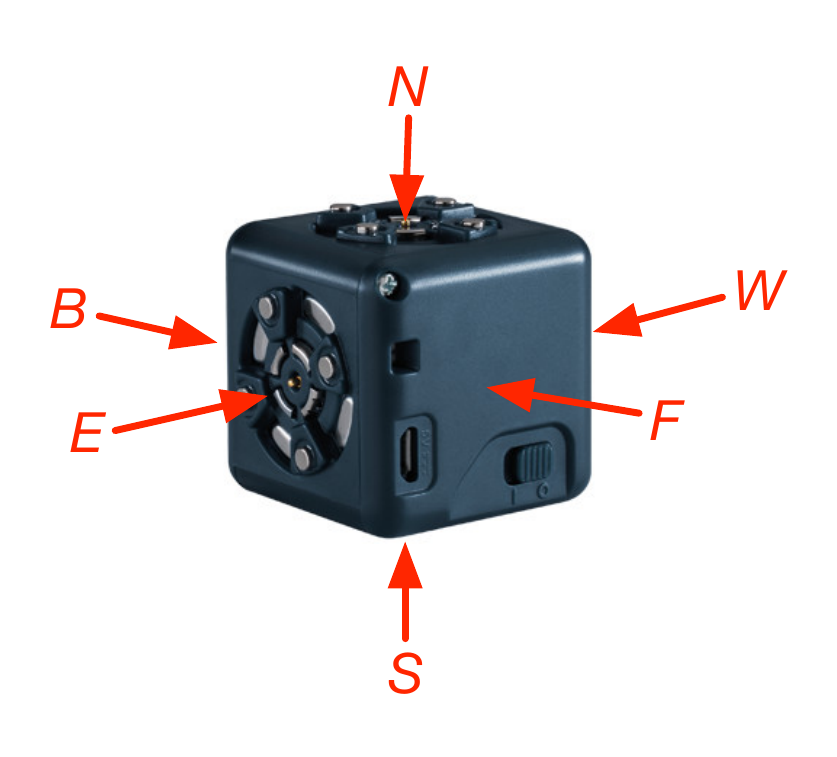}}} 
\subfigure[]{\scalebox{0.46}{\includegraphics{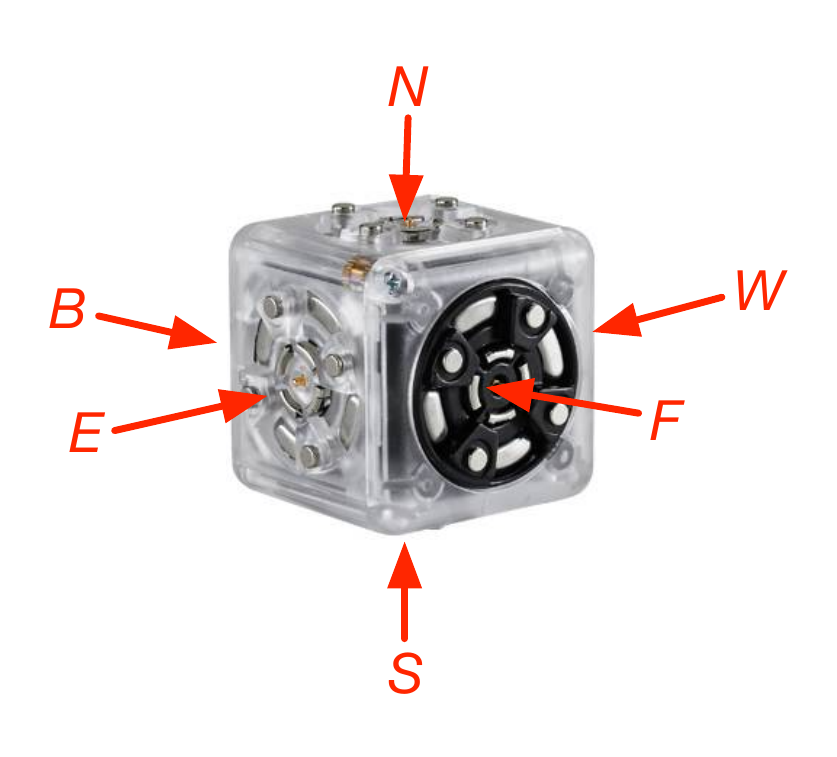}}} 
\subfigure[]{\scalebox{0.46}{\includegraphics{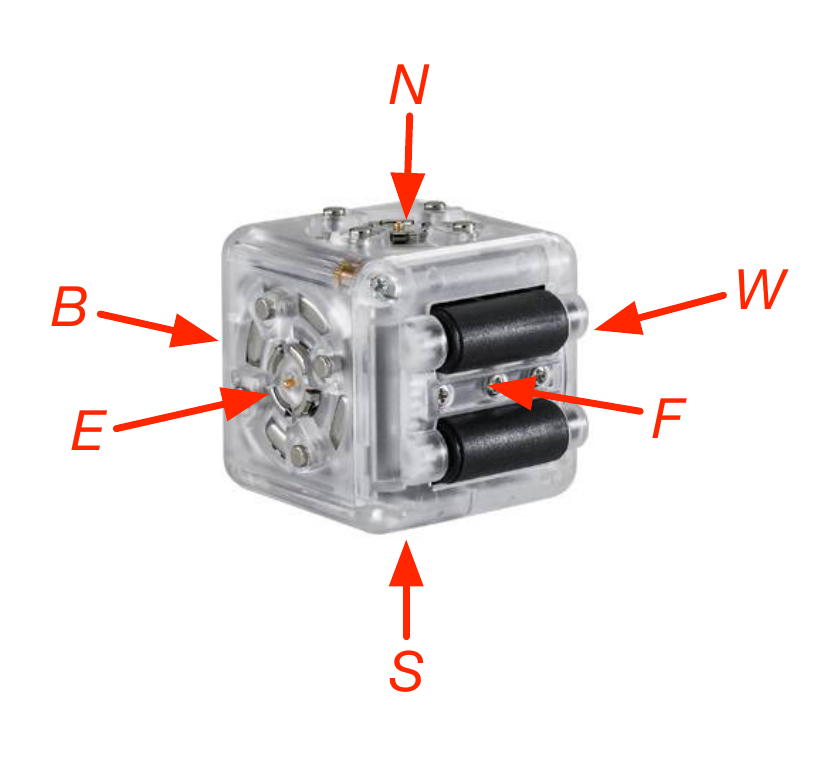}}} \hspace{1.2cm}
\subfigure[]{\scalebox{0.46}{\includegraphics{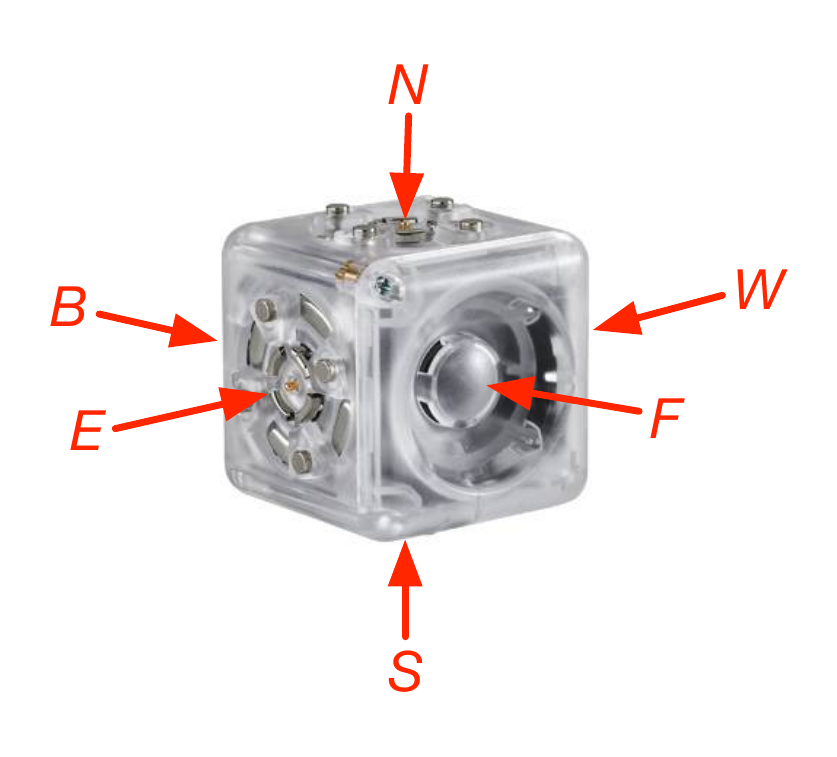}}} 
\subfigure[]{\scalebox{0.46}{\includegraphics{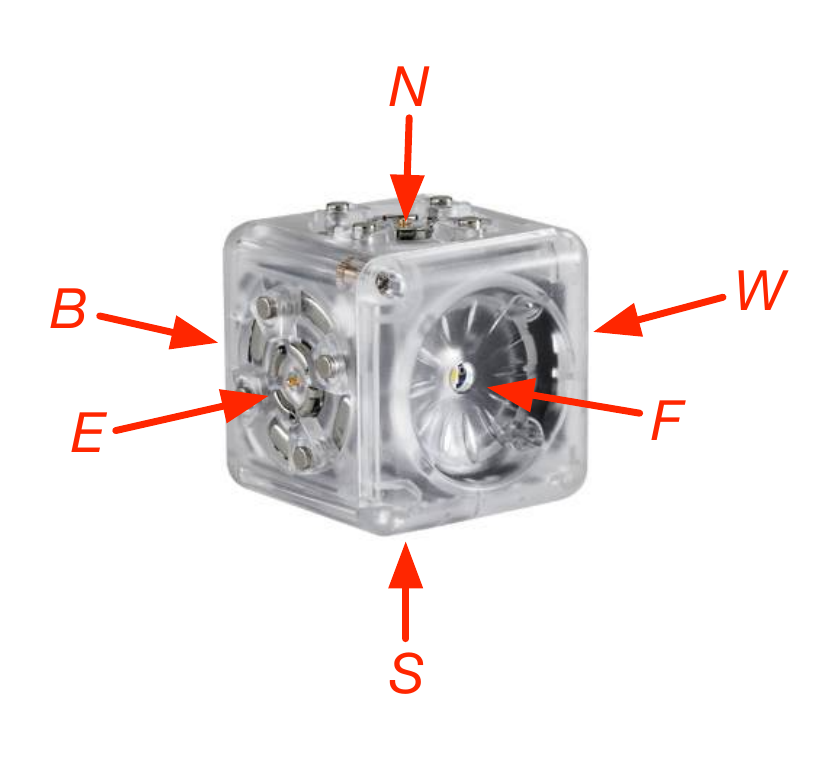}}} 
\subfigure[]{\scalebox{0.46}{\includegraphics{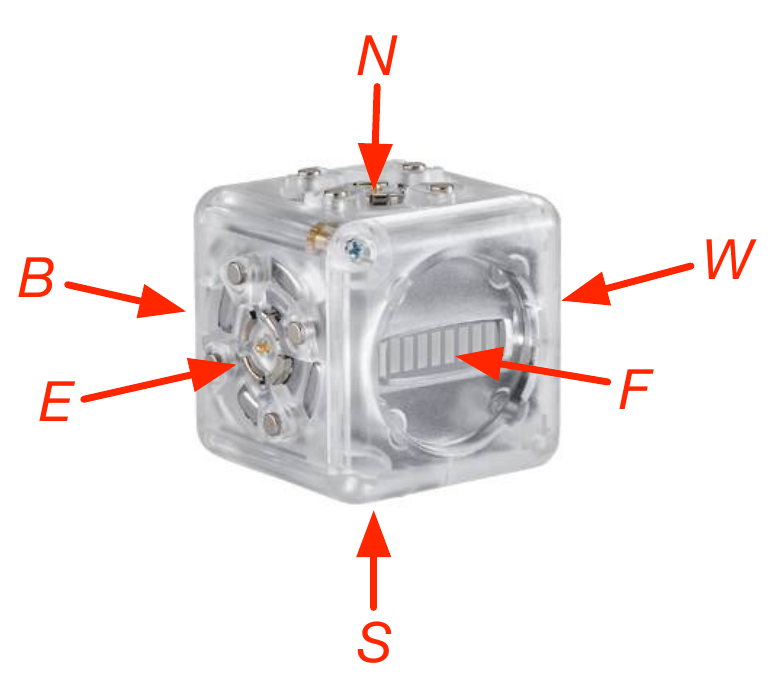}}} \hspace{1.2cm}
\subfigure[]{\scalebox{0.46}{\includegraphics{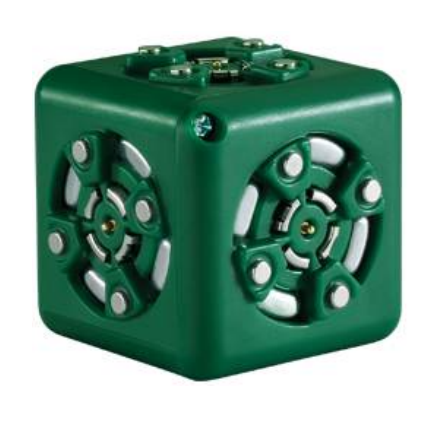}}} \hspace{1.4cm}
\subfigure[]{\scalebox{0.46}{\includegraphics{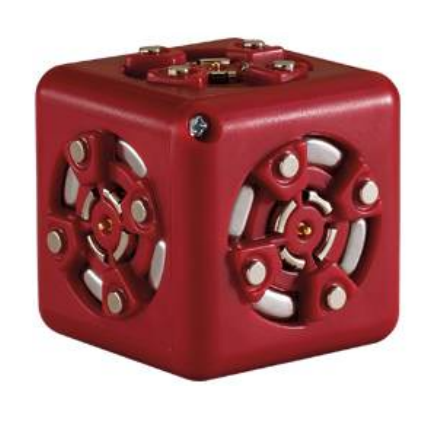}}} \hspace{1.8cm}
\subfigure[]{\scalebox{0.46}{\includegraphics{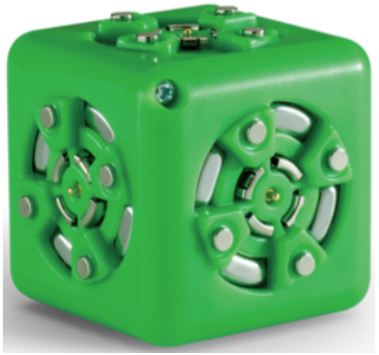}}}
\end{center}
\caption{Labelling faces in Cubelets. They are represented following the three-dimensional cardinal points as: $N$-north, $S$-south, $E$-east, $W$-west, $B$-back, $F$-front. Thus we have (a) distance, (b) brightness, (c) temperature, (d) battery, (e) rotate, (f) drive, (g) speaker, (h) flashlight, and (i) bar graph. For the case (j) blocker, (k) inverse, and (l) passive, here is not necessary label any face because the whole cube is symmetric.}
\label{CubeletsLabel}
\end{figure}

Let $\Sigma = \{0,\ldots, 17\}$ the alphabet of Cubelets robots (Table~\ref{symbolCubeletsTable}), where each of one defines a robot by itself. We can say that $\Sigma^*$ represents a universe of Cubelets. Thus, we have two disjoint sets of robots: $\Sigma_{F}$ and $\Sigma_{NF}$. $\Sigma_{F} \subset \Sigma^*$ is a set of functional robots and $\Sigma_{NF} \subset \Sigma^*$ is a set of non functional robots, e.g. robots without a battery. Functional robots are assembled in a three ${\mathcal Z}^3$ and encoded as a sequence of symbols. Thus, a functional robot can be expressed as a one-dimensional robot $w \in \Sigma^*$. As known, Cubelets work in three dimensions and for to get a practical representation and codification we will concatenate every coordinate as a string. This way, we will to codify each robot as a symbolic equation and this regular expression shall be practical for construct a robot. Hence every Cubelet is defined recursively and therefore the union, juxtaposition and Kleene closer of every cube is a regular expression \cite{mcintosh1990programming, minsky1967computation}.

In Fig.~\ref{3D-diagram}, we can see how this tree-dimensional space is labelled and can be represented as a string, as follows:

\begin{align*}
w &= x_{0,0,0} x_{1,0,0} \cdots x_{n-1,0,0} x_{0,1,0} \cdots x_{n-1,1,0} x_{0,m-1,0} x_{1,m-1,0} \cdots x_{n-1,m-1,0} \\
   & \cdots x_{0,0,1} \cdots x_{n-1,m-1,p-1}.
\end{align*}

Later, we will identify every face to each cube. It is illustrated in Fig.~\ref{CubeletsLabel}, where we can see mainly the front position which is the most important to fix and to get a correct position of the Cubelet which is recognized for the front $(F)$ orientation. In this case, we show this labelling for Cubelets that are used to construct our circuit. They are Cubelets flashlight $(fl)$, distance $(di)$, brightness $(br)$, battery $(ba)$, inverse $(in)$, blocker $(bl)$. This way, a codification base is denoted as $(F,N,W)$. If the position change hence we will rotate the cube and update these three parameters. If a coordinate has not a cube then we will represent this one with a blank ({\bf B}) which represents an empty string of $\Sigma^*$.

This way, the next string determines a robot walking $w_1$ when the distance Cubelet find an object close. To represent a robot with a string we should select two first letter from the Tab.~\ref{symbolCubeletsTable}, for example, 

\begin{itemize}
\item The robot that moves as a car $w_{car}$ (see Fig.~\ref{w_scar}). It has a volume (total area) of $3 \times 1 \times 1$ cubes with a mass (total number of Cubelets) of 3 cubes. The expression to construct this robot is:

$$
w_{scar} = dr_{(2,0,0)}^{(N,B,W)} \cdot ba_{(1,0,0)}^{(F,N,W)} \cdot di_{(0,0,0)}^{(S,F,W)}.
$$

\begin{figure}[th]
\centering
\includegraphics[width=0.5\textwidth]{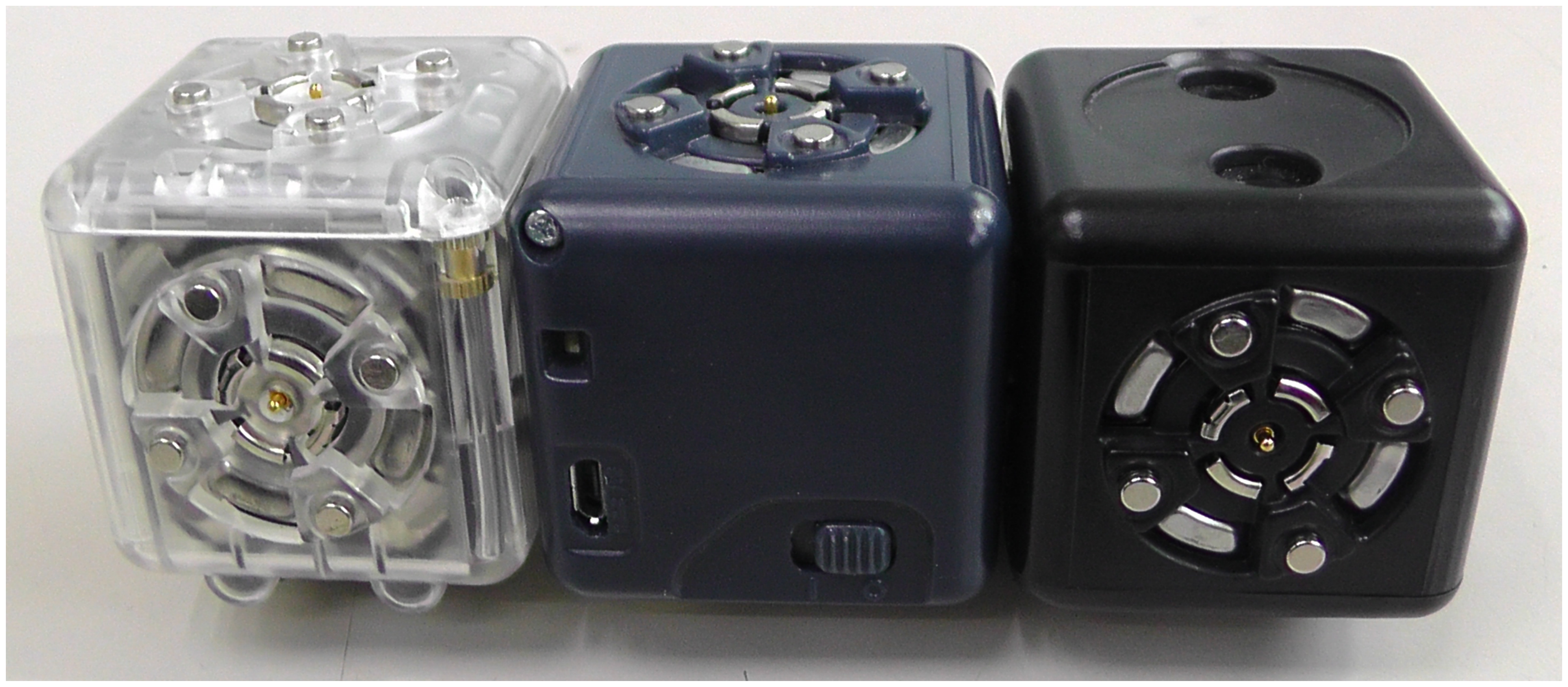}
\caption{Cubelet robot simple car $w_{scar}$.}
\label{w_scar}
\end{figure}

\item A robot that detects a fire (see Fig.~\ref{w_fire}). It has a volume of $3 \times 1 \times 1$ cubes with a mass of 3 cubes. The expression to construct this robot is:

$$
w_{fire} = fl_{(2,0,0)}^{(S,F,W)} \cdot ba_{(1,0,0)}^{(F,N,W)} \cdot te_{(0,0,0)}^{(S,F,W)}.
$$

\begin{figure}[th]
\centering
\includegraphics[width=0.5\textwidth]{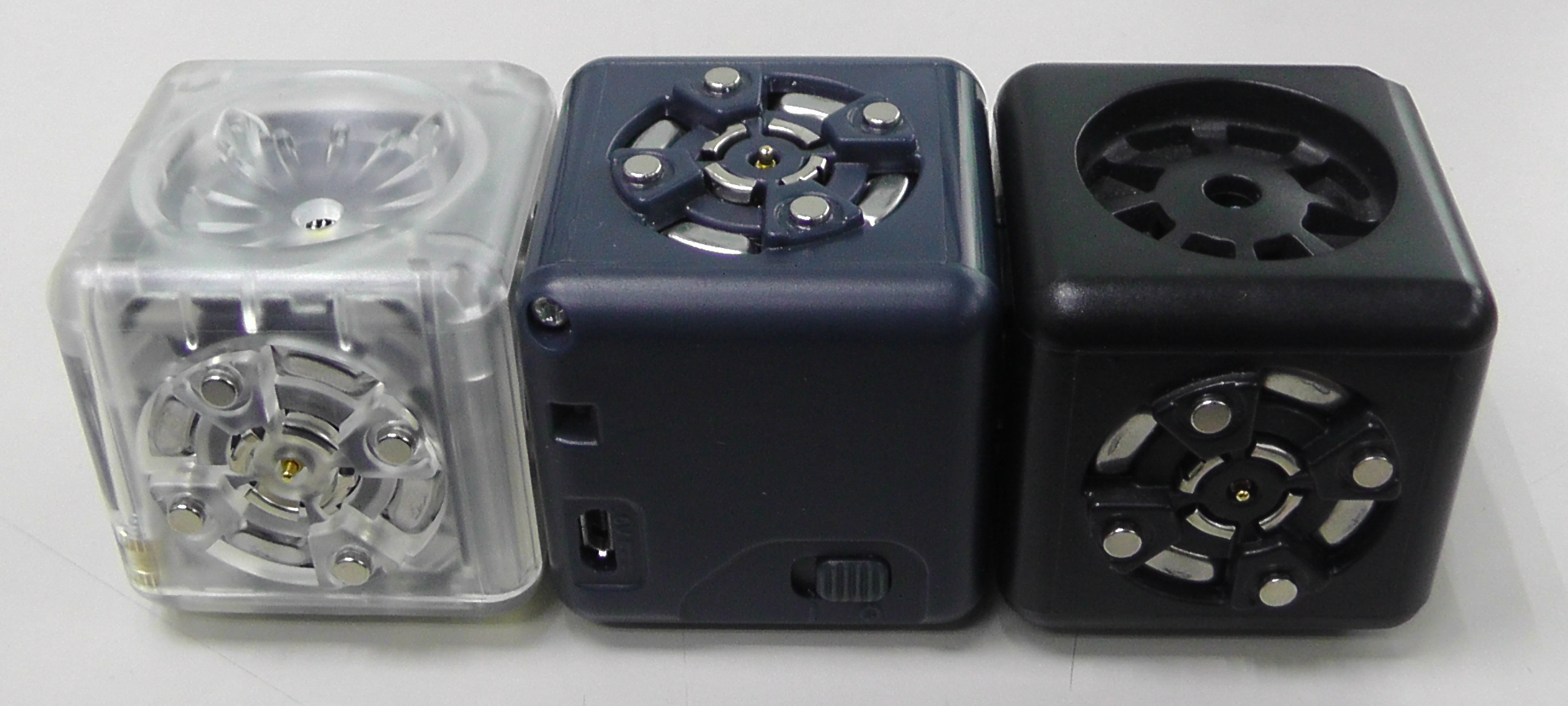}
\caption{Cubelet robot detector $w_{fire}$.}
\label{w_fire}
\end{figure}

\item A robot that acts as an autonomous car\footnote{Autonomous car with Cubelets. \url{https://youtu.be/Q5ueOMFNhwQ}} avoiding obstacles and changing its direction of movement (see Fig.~\ref{w_autonomouscar}). It as a volume of $3 \times 2 \times 3$ cubes with a mass of 8 cubes. The expression to construct the robot is:

\begin{figure}[th]
\centering
\includegraphics[width=0.5\textwidth]{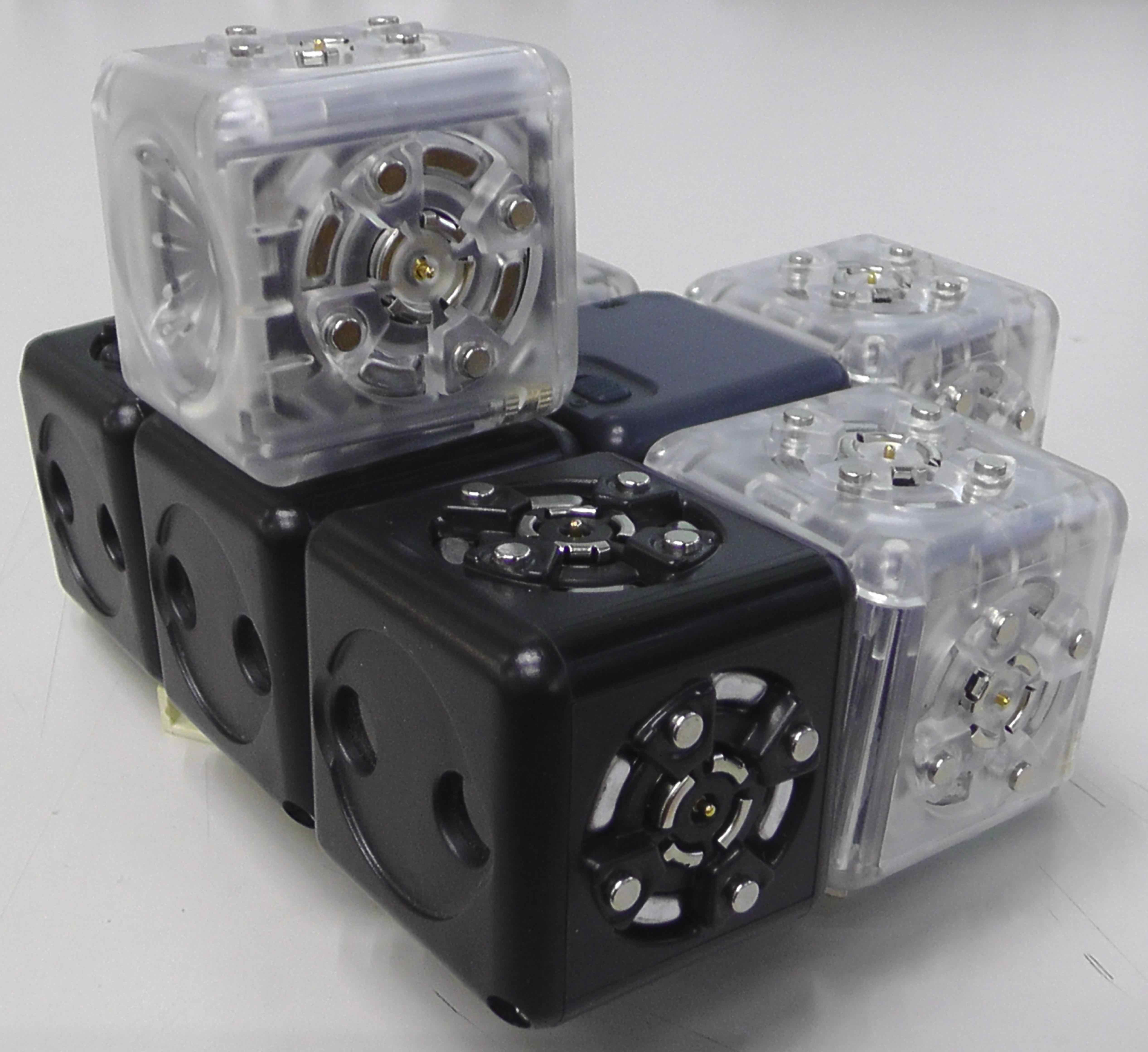}
\caption{Cubelet robot autonomous car $w_{acar}$.}
\label{w_autonomouscar}
\end{figure}

\begin{eqnarray*}
w_{acar} = (({\bf B} \cdot fl_{(1,0,0)}^{(F,N,W)}) \cdot (di_{(2,1,0)}^{(F,N,W)} \cdot di_{(1,1,0)}^{(F,N,W)} \cdot di_{(0,1,0)}^{(F,N,W)})) \cdot \\ (dr_{(2,1,1)}^{(W,B,N)} \cdot ba_{(1,1,1)}^{(F,N,W)} \cdot dr_{(0,1,1)}^{(W,B,N)}) \cdot ({\bf B} \cdot (dr_{(1,1,2)}^{(S,F,W)})).
\end{eqnarray*}

\item A robot that seems as a caterpillar.\footnote{A robotic caterpillar with Cubelets. \url{https://youtu.be/E9nVEPK3UFY}} This one avoid obstacles and change its direction automatically (see Fig.~\ref{w_caterpillar}). Particularly this robot is extendable and you can concatenate parts of it as long as you want. It as a volume of $2 \times 2 \times 4$ cubes with a mass of $7 \times 2$ cubes. Particularly, the robot $w_{caterpillar}$ is constructed from two independent robots. Also, it is assembled across a reflection and extensions. The basic structure is defined in a volume of $2 \times 2 \times 4$ Cubelets but this basic structure is not the caterpillar form, then we need a reflection of the basic structure and connect both structures with 4 Lego adapters and 2 Lego bars, thus we have $2 \times (7 \times 2)$ Cubelets to yield finally the most small caterpillar. To get extensions and increase the length of the caterpillar we need connect $2 \times 3$ Cubelets for each side, the basic structure for the extension is determined by one battery, a rotator and a light Cubelets, and its reflection will be connected on the other side. So to stabilize the robot we need connect additional Lego adapters in the top. The expression to construct the robot is:

\begin{eqnarray*}
w_{caterpillar} = ({\bf B} \cdot di_{(1,1,0)}^{(F,N,W)}) \cdot (({\bf B} \cdot di_{(1,0,1)}^{(F,N,W)}) \cdot (di_{(0,1,0)}^{(F,N,W)} \cdot pa_{(1,1,1)}^{(F,N,W)})) \cdot \\ (({\bf B} \cdot ba_{(1,1,2)}^{(N,F,W)}) \cdot (fl_{(0,1,3)}^{(E,N,B)} \cdot ro_{(1,1,3)}^{(E,N,B)}))^*.
\end{eqnarray*}

\begin{figure}[th]
\centering
\includegraphics[width=1\textwidth]{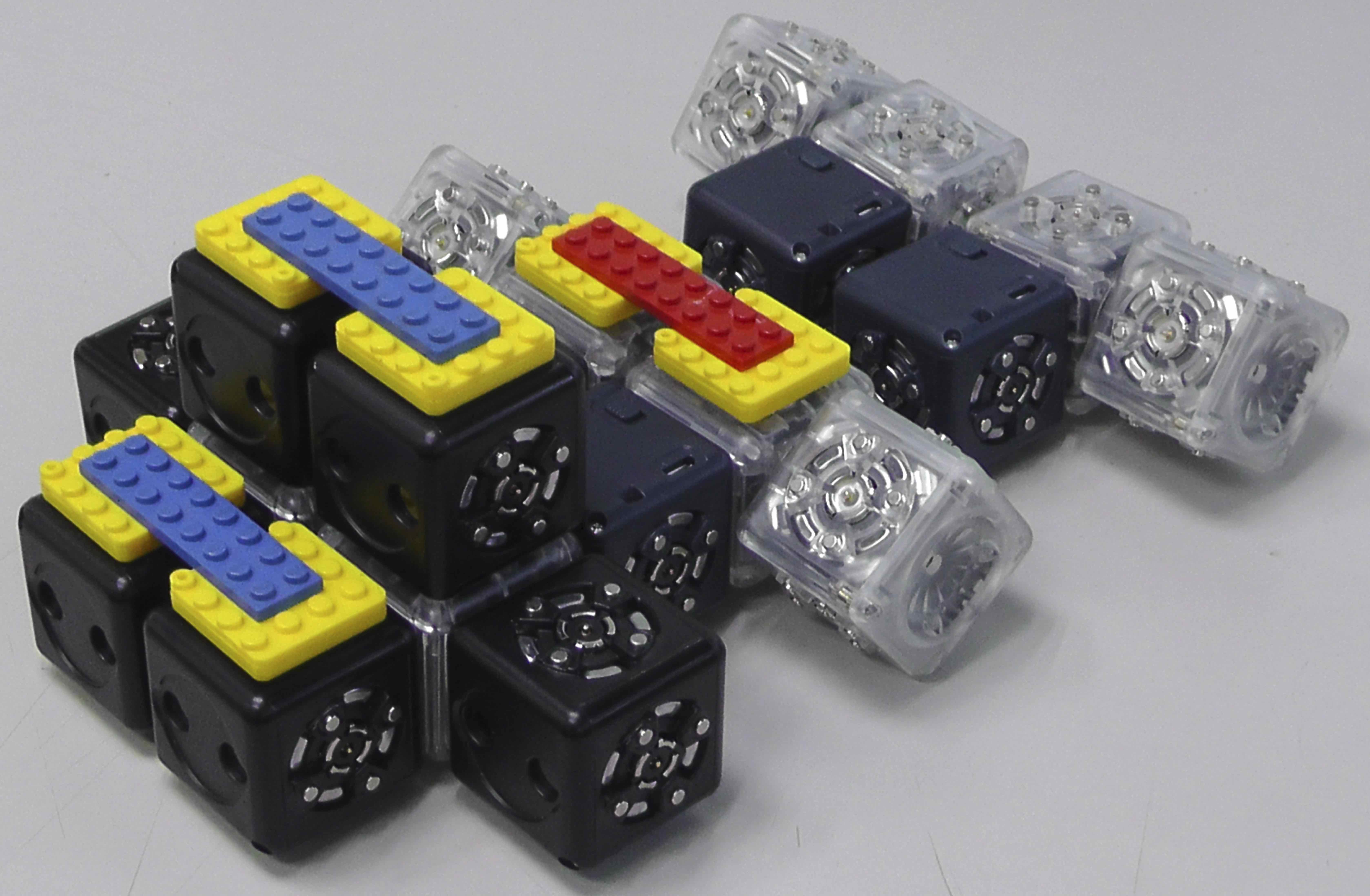}
\caption{Cubelet robot caterpillar $w_{caterpillar}$.}
\label{w_caterpillar}
\end{figure}

\item A robot capable of simulate logical gates\footnote{Logic gates with Cubelets: {\sc NAND} and {\sc XOR} gates. \url{https://youtu.be/O2XSK33VjTU}} just with Cubelets (Fig.~\ref{w_logicgates}). This one use a head that read values from other Cubelets moving in two directions. It is a robot $w_{lg}$ with a volume of $3 \times 3 \times 3$ cubes with a mass of 9 cubes. The robot $w_{lg}$ is defined for the next expression:

\begin{eqnarray*}
w_{lg} = ({\bf B} \cdot di_{(1,1,0)}^{(N,B,E)}) \cdot (({\bf B} \cdot bg_{(1,0,1)}^{(S,F,E)}) \cdot ({\bf B} \cdot ro_{(1,1,1)}^{(S,F,E)}) \cdot (dr_{(0,2,1)}^{(E,B,S)} \cdot \\ ba_{(1,2,1)}^{(F,N,E)} \cdot dr_{(2,2,1)}^{(E,B,S)})) \cdot (dr_{(0,2,2)}^{(E,B,S)} \cdot ba_{(1,2,2)}^{(F,N,E)} \cdot dr_{(2,2,2)}^{(E,B,S)}).
\end{eqnarray*}

But the  robot $w_{lg}$ needs another robot which represent the `tape' and it is represented by a second robot $w_{lg\_tape}$ with a volume of $10 \times 1 \times 1$ cubes and a mass of 10 cubes. To get this robot we need the next expression:

\begin{figure}[th]
\centering
\includegraphics[width=1\textwidth]{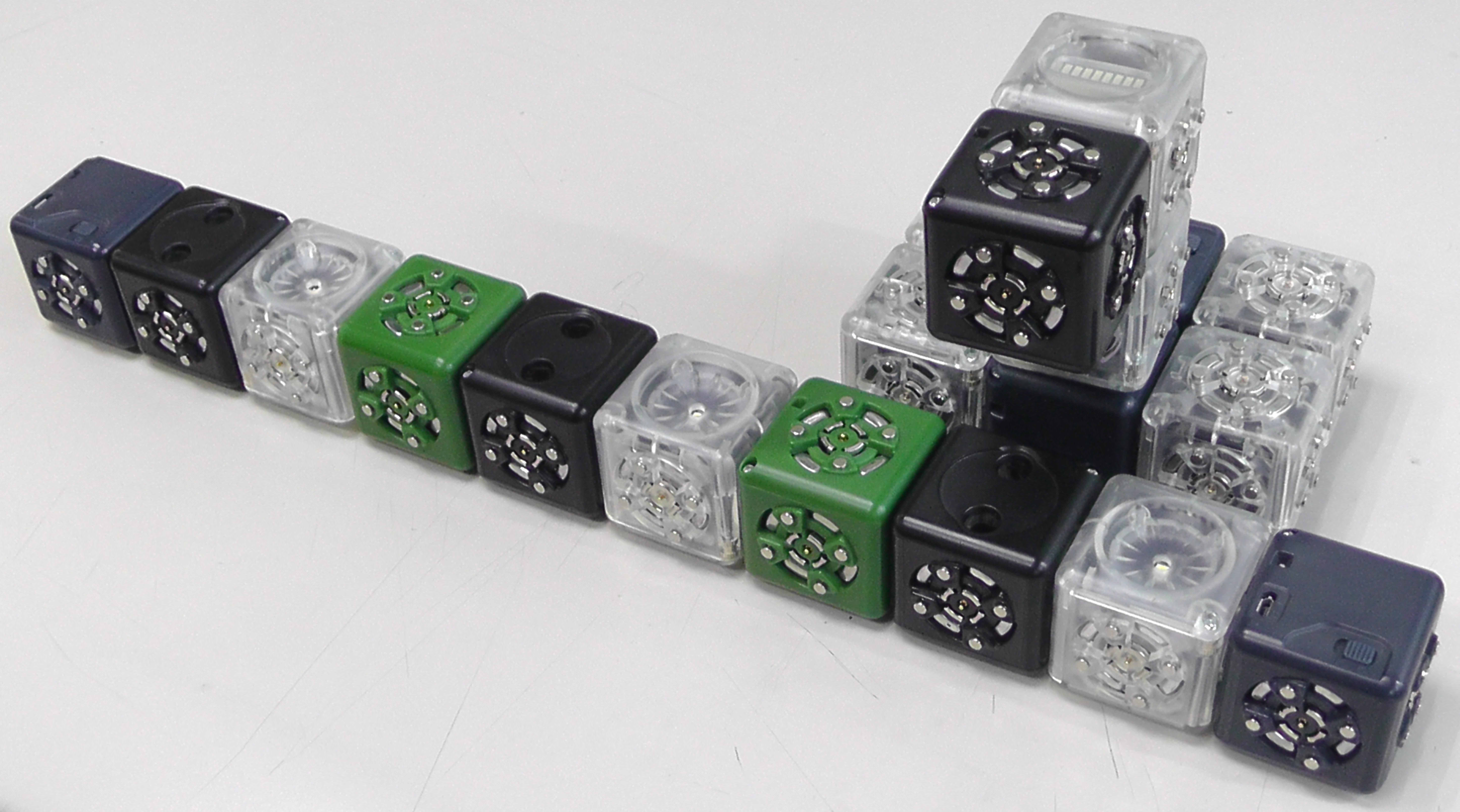}
\caption{Cubelet robot performing logic gates.}
\label{w_logicgates}
\end{figure}

\begin{eqnarray*}
w_{lg\_tape} = (ba_{(0,0,0)}^{(S,F,E)} \cdot fl_{(1,0,0)}^{(S,F,E)} \cdot di_{(2,0,0)}^{(E,F,B)} \cdot bo_{(3,0,0)} \cdot fl_{(4,0,0)}^{(S,F,E)} \cdot di_{(5,0,0)}^{(E,F,B)} \cdot \\ bo_{(6,0,0)} \cdot fl_{(7,0,0)}^{(S,F,E)} \cdot di_{(8,0,0)}^{(E,F,B)} \cdot ba_{(9,0,0)}^{(S,F,E)}).
\end{eqnarray*}

\item In \cite{figueroa2019turing}, we show a robotic Turing machine constructed from Cubelets and a lot of Lego bricks.\footnote{CULET a robotic Turing machine. \url{https://www.comunidad.escom.ipn.mx/ALIROB/CULET/}}

\begin{figure}[th]
\begin{center}
\subfigure[]{\scalebox{0.025}{\includegraphics{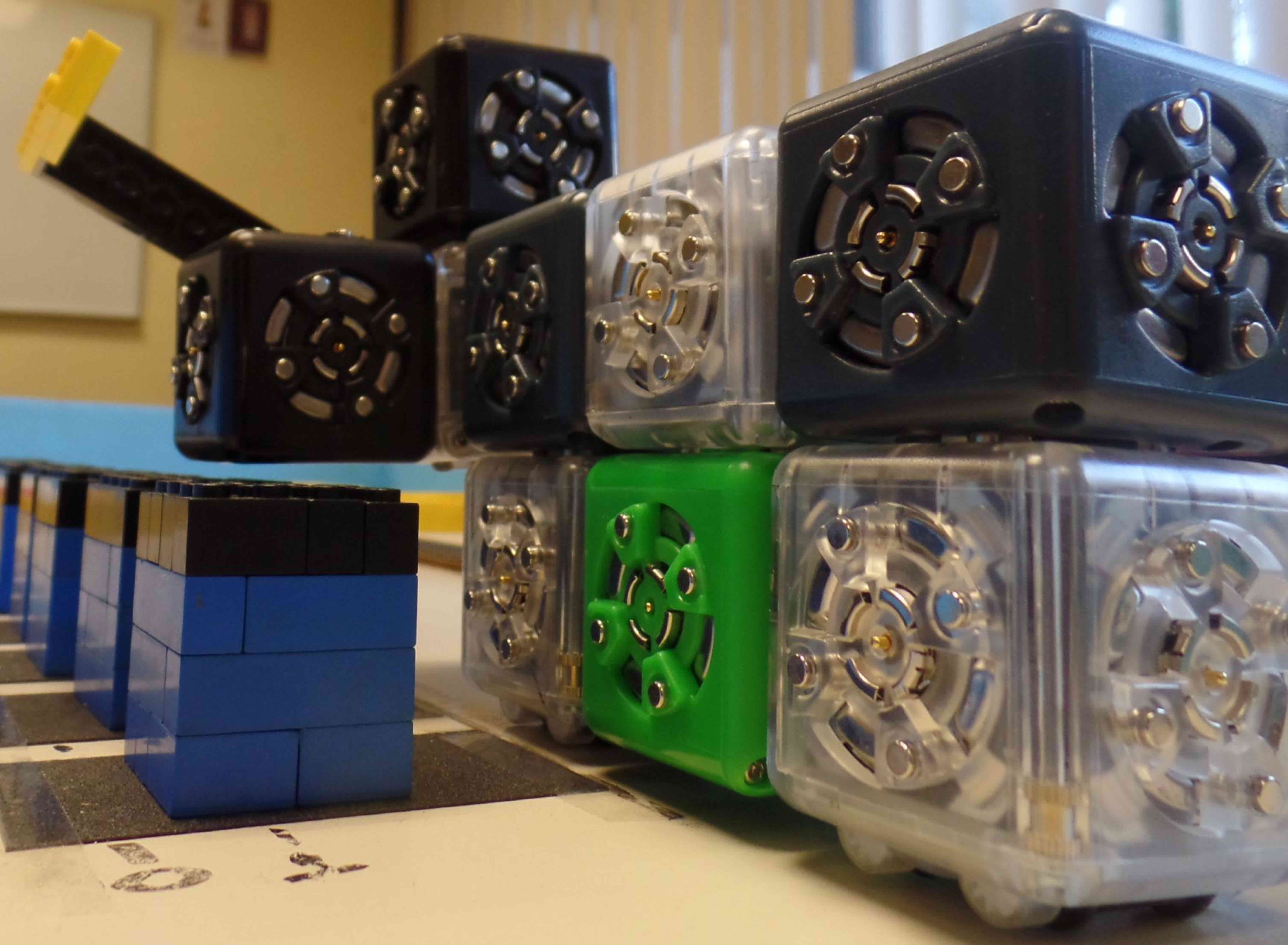}}} 
\subfigure[]{\scalebox{0.075}{\includegraphics{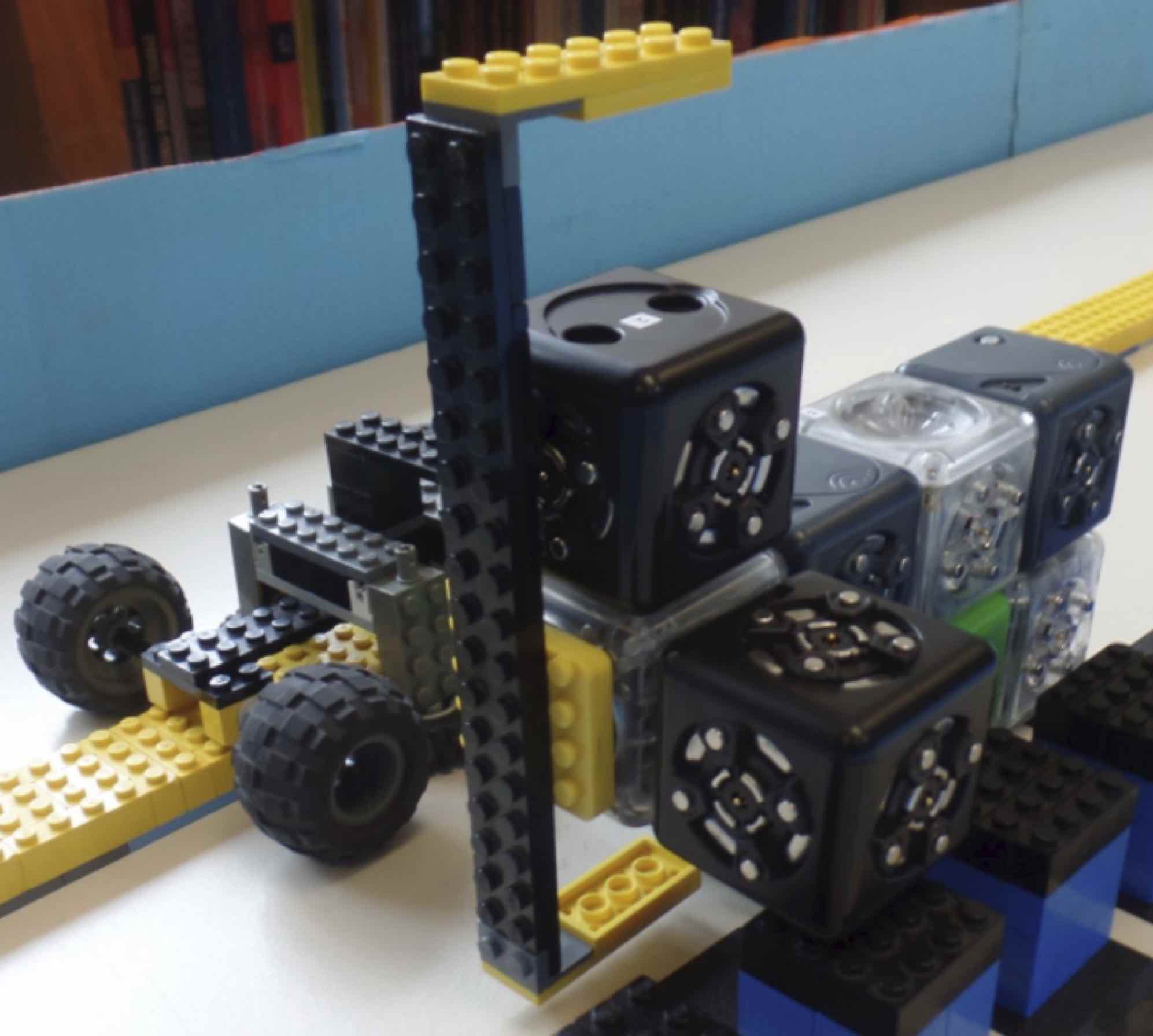}}}
\end{center}
\caption{Robotic Turing machine constructed with Cubelets and Lego bricks.}
\label{w_culet}
\end{figure}

Figure~\ref{w_culet} displays two snapshot of the machine. Particularly, CULET is a robot that needs a lot of different Lego bricks. But, here we explain what is the equation to get this robot $w_{TM}$. It is defined for the next expression:

\begin{eqnarray*}
w_{TM} = ({\bf B}^3 \cdot di_{(3,1,0)}^{(N,B,E)}) \cdot (({\bf B}^3 \cdot di_{(3,0,1)}^{(S,F,E)}) \cdot (ba_{(0,1,1)}^{(S,F,E)} \cdot \\ fl_{(1,1,1)}^{(S,F,E)} \cdot ba_{(2,1,1)}^{(S,F,E)} \cdot ro_{(3,1,1)}^{(W,N,F)}) \cdot (dr_{(0,2,1)}^{(E,B,S)} \cdot pa_{(1,2,1)} \cdot dr_{(2,2,1)}^{(E,B,S)})).
\end{eqnarray*}

\end{itemize}

Our goal is to find a string $w$ such that $w$ belongs to an equivalent computable system made only with Cubelets. We proof how logical circuits can be constructed with Cubelets and the computation is performed by  propagating patterns.

\section{Computing with propagating patterns}

In 2008 we constructed a Life-like cellular automaton capable for simulating a propagating of patterns on a feedback of channels and implemented a majority gate in this automaton~ \cite{martinez2008logical}. Later we designed additional sets of gates and a developed a binary full adder with the propagating patterns~\cite{martinez2010majority, martinez2010computation}.

Cellular automata are often used to design unconventional computing in several ways. Historically, several protocols of handling signals propagating in cellular automata have been proposed by von Neumann era \cite{neumann1966theory}. These were advanced by a two-dimensional cellular automaton with less states by Codd \cite{codd1968cellular, hutton2010codd}, the WireWorld invented by Silverman \cite{dewdney1990column}, and the reversible computers designed by Morita \cite{morita2001simple}. With regards to one-dimensional cellular automata, the exhaustive analysis of signal interactions was done by Mazoyer \cite{mazoyer1996computations} and some rules working with simple and compact signals have been reported by Mitchell {\it et al.} \cite{mitchell1994evolving}. Recently Mart{\'i}nez {\it et al.} \cite{martinez2018conservative, martinez2018logical} proposed a cellular automaton with memory capable of simulating logic gates with just one particle. A complimentary version of representing information in cellular automata space is based on a composition of signals interpreted as particles (gliders, waves, mobile-self localisation)~\cite{adamatzky2002collision, griffeath2003new, adamatzky2010game, adamatzky2016advances}.

\begin{figure}[th]
\begin{center}
\subfigure[]{\scalebox{0.35}{\includegraphics{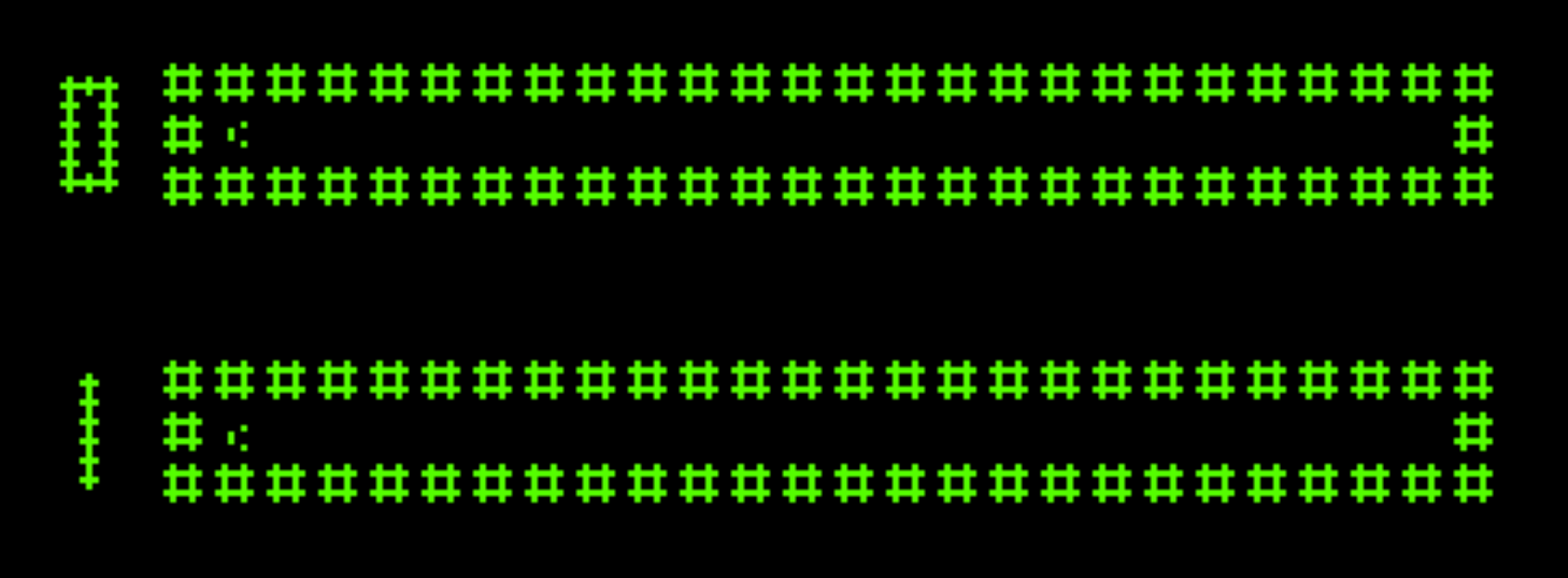}}} 
\subfigure[]{\scalebox{0.35}{\includegraphics{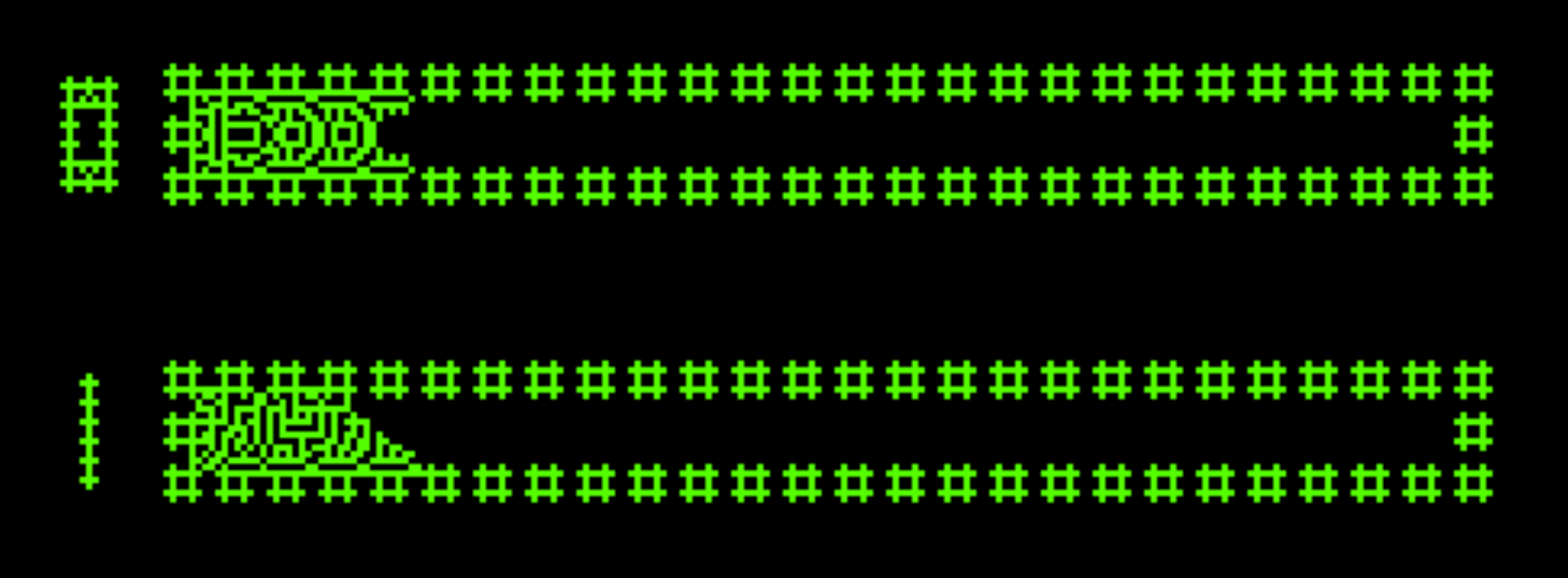}}} 
\subfigure[]{\scalebox{0.35}{\includegraphics{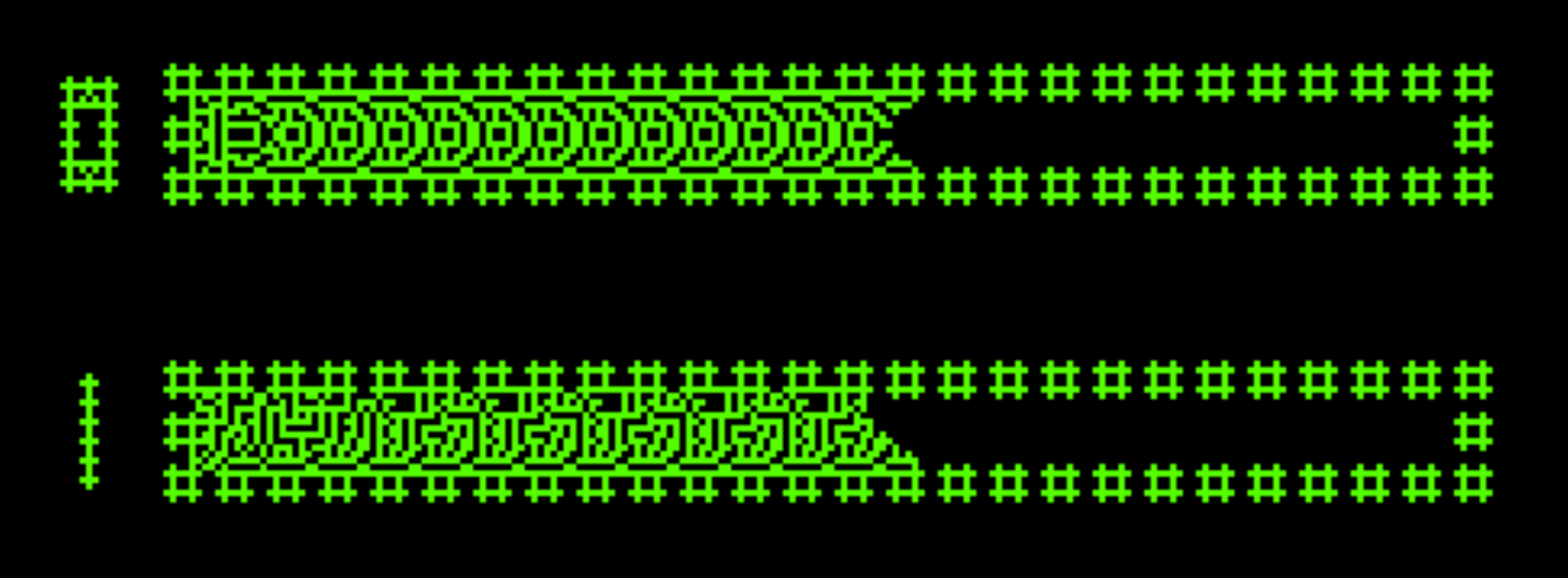}}} 
\subfigure[]{\scalebox{0.35}{\includegraphics{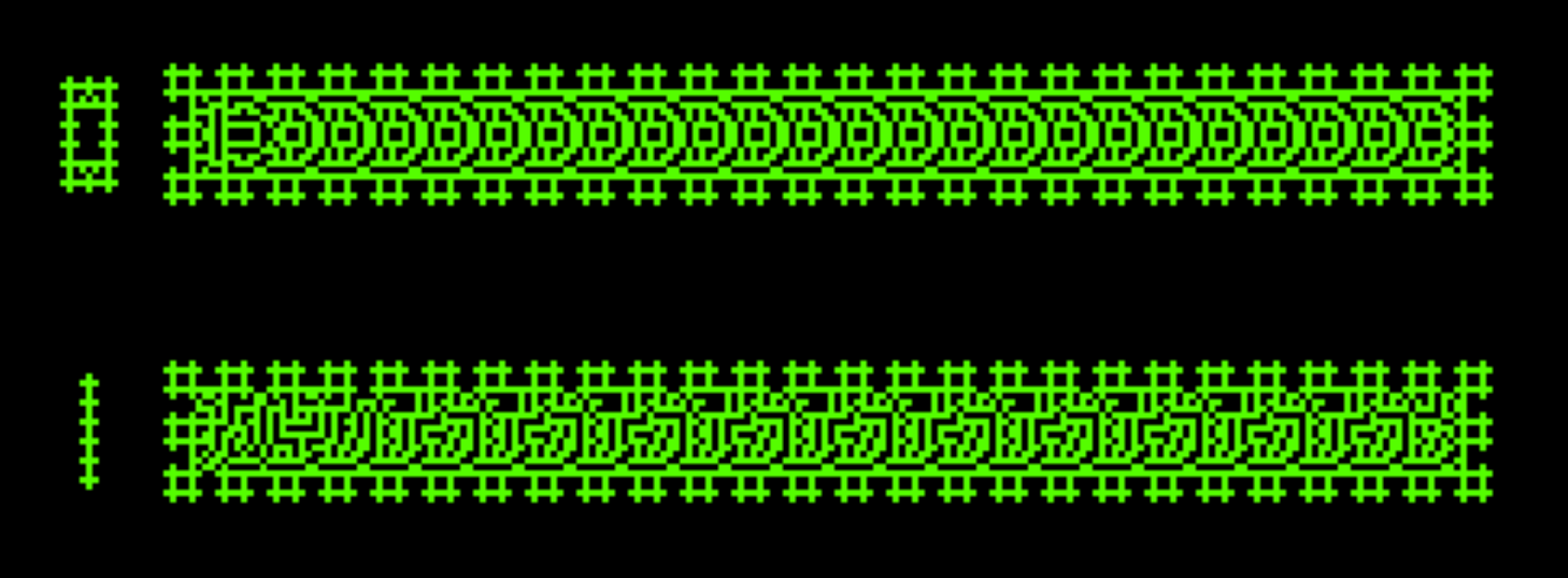}}}
\end{center}
\caption{Patterns propagating by particle reaction in information channels with the Life-like rule $B2/S2345$. (a) The initial condition with two particles before they collide. This way, snapshots (b,c,d) beginning of the propagation of symmetric and asymmetric patterns in (b) 37 iterations, (c) 114 iterations, and (d) 208 iterations after the collision.}
\label{signals}
\end{figure}

Computing with propagation patterns exploits the dynamics of propagating patterns that encode binary value, these patterns emerge as chaotic regions. They are constructed by streams of particles traveling and colliding with a wall, defining symmetric or asymmetric patterns propagating and competing for the space. To geometrically constrain the patterns and time their collisions we design a network of channels, typically using arrangements as a $n$-junction function. Fronts of propagating phase (excitation) or diffusive waves represent signals, the values of logical variables. When fronts interact at the junctions some fronts annihilate or new fronts emerge. The patterns propagating in the output channels represent results of the computation. Using this paradigm a range of logic gates have been implemented in \cite{martinez2008logical} and a binary full adder constructed in \cite{martinez2010majority, martinez2010computation}. To design a logic circuits with Cubelets we will use the Life rule $B2/S2345$\footnote{Majority adder implementation in Life rule B2/S2345. \url{https://comunidad.escom.ipn.mx/genaro/Diffusion_Rule/Life_B2-S2345.html}}, the cellular automata which exhibits a chaotic behaviour, and is governed by a semi-totalistic function described as follows. Each cell takes two states `0' (`dead') and `1' (`alive'), and updates its state depending on the states of its eight closest neighbour as follows,

\begin{enumerate}
\item {\it Birth}: a central cell in state 0 at time step $t$ takes state 1 at $t+1$ if it has exactly two neighbours in state 1.
\item {\it Survival}: a central cell in state 1 at time $t$ remains in the state 1 at time $t+1$ if it has two, three, four or five live neighbours.
\item {\it Death}: any other case.
\end{enumerate}

\begin{figure}[th]
\begin{center}
\subfigure[]{\scalebox{0.19}{\includegraphics{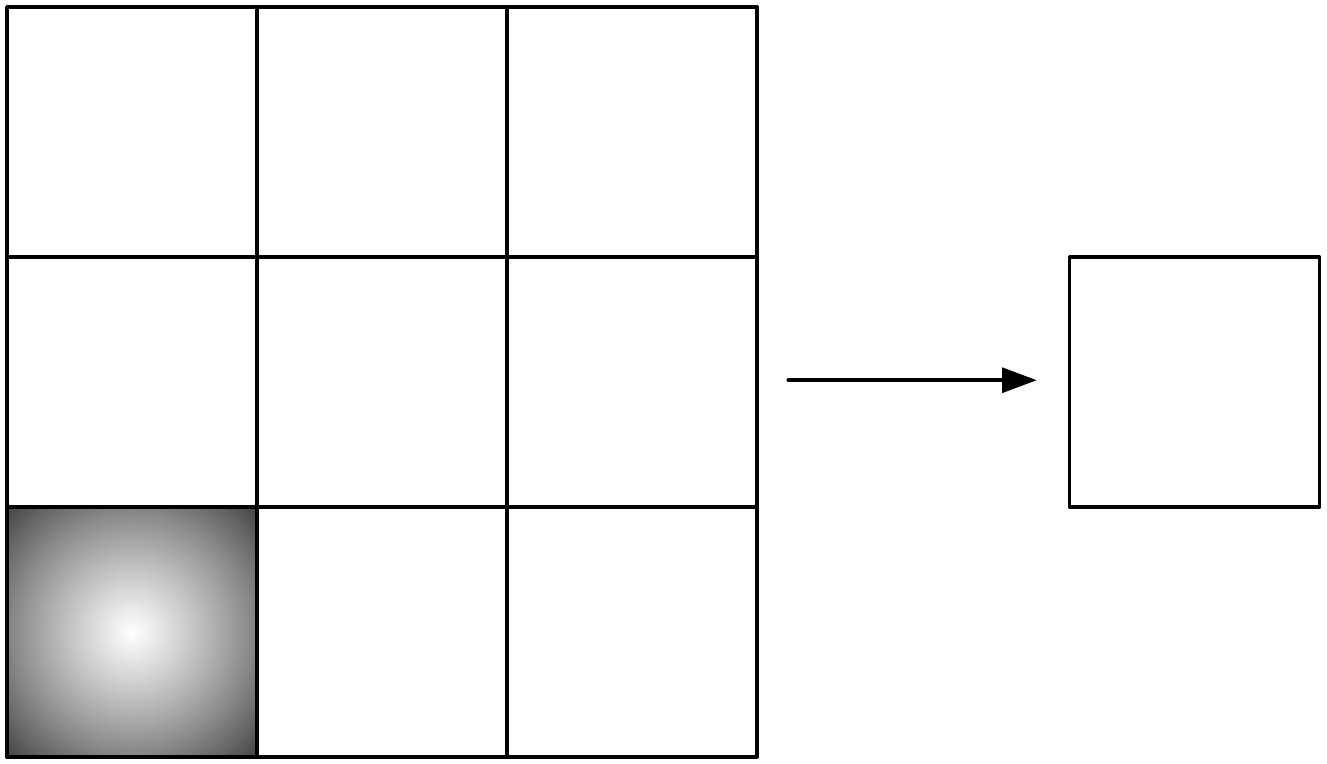}}} \hspace{0.3cm}
\subfigure[]{\scalebox{0.19}{\includegraphics{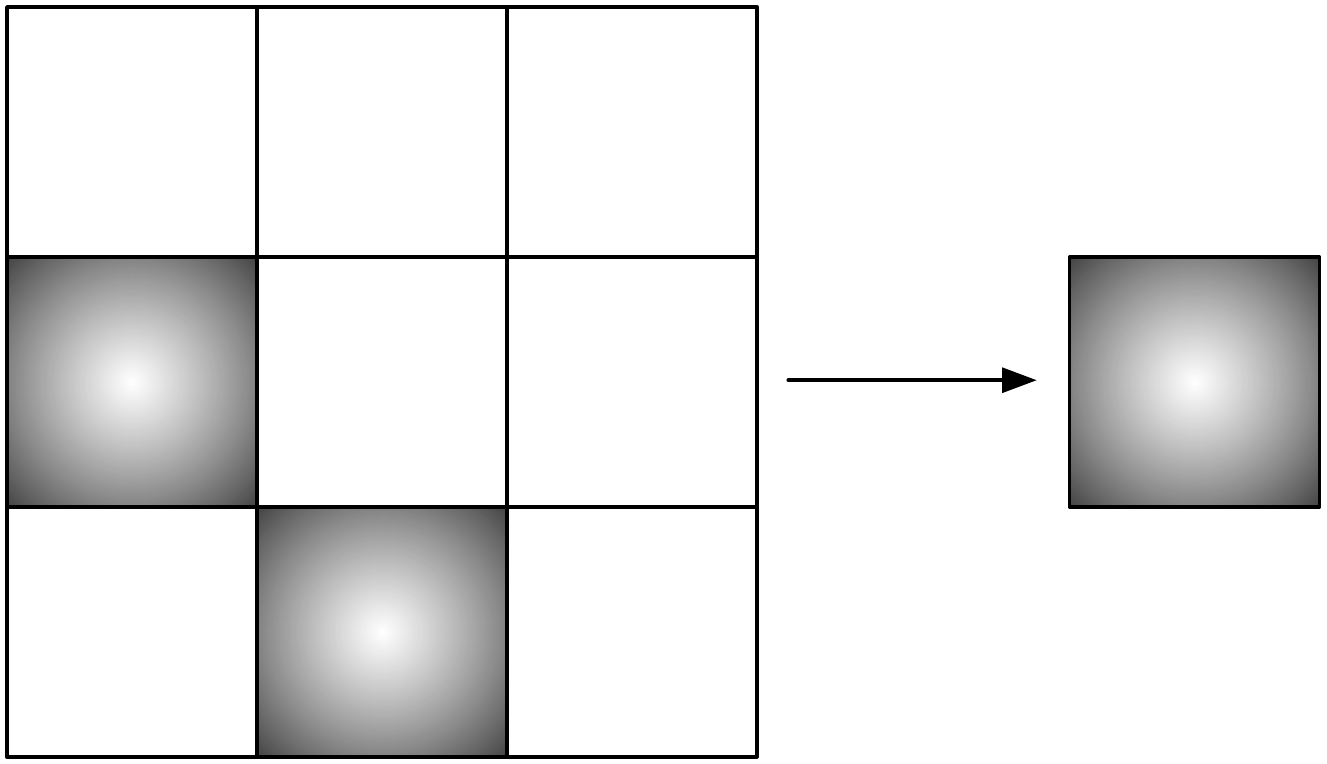}}} \hspace{0.3cm}
\subfigure[]{\scalebox{0.19}{\includegraphics{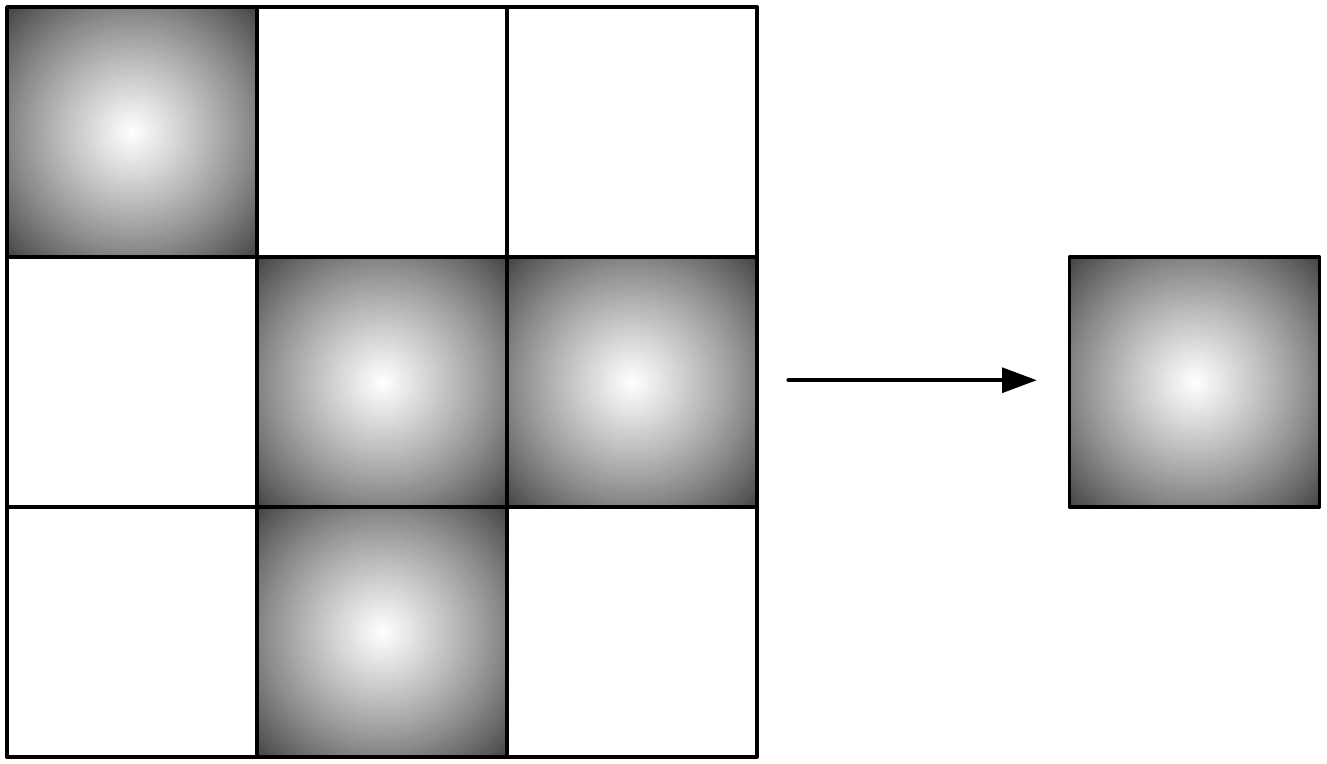}}} \hspace{0.3cm}
\subfigure[]{\scalebox{0.19}{\includegraphics{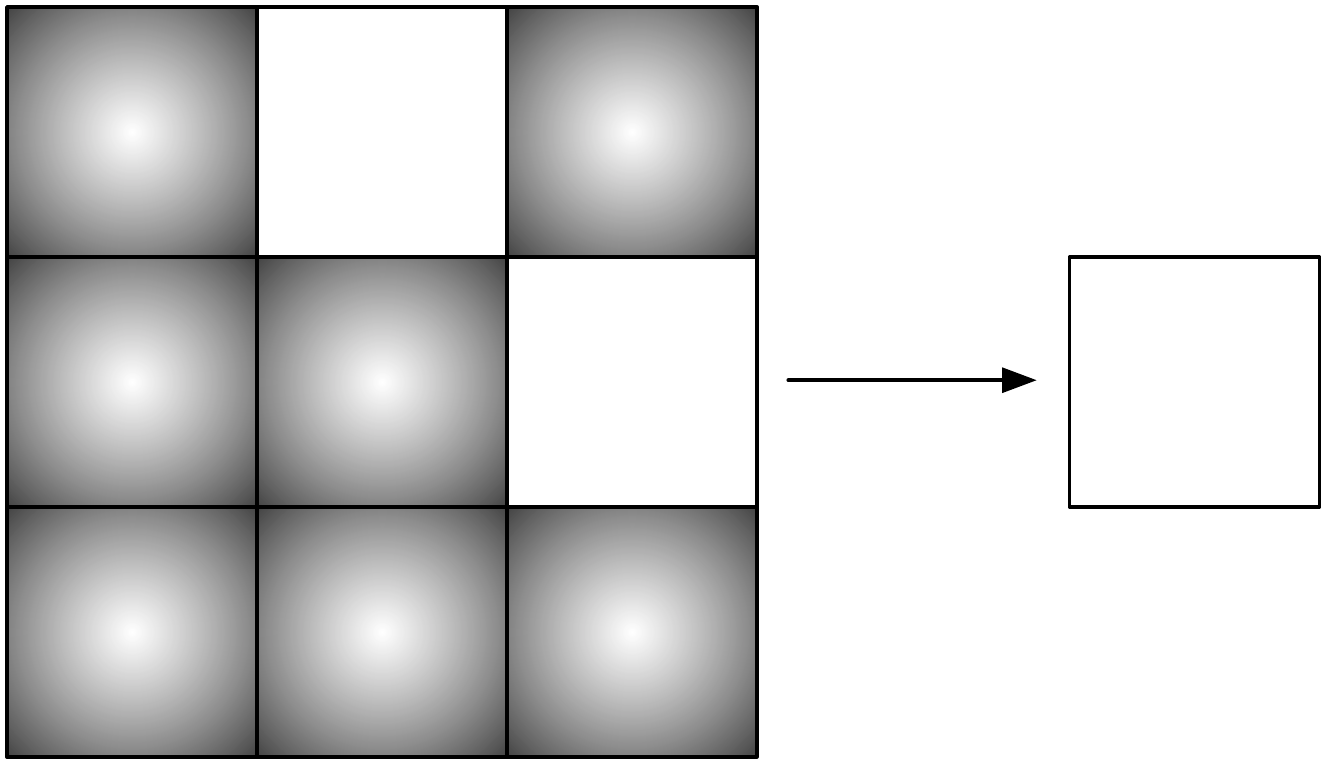}}}
\end{center}
\caption{The Life-like rule $B2/S2345$ evolves on a Moore neighbourhood. (a) Shows a condition where the sum of the neighbours do not permit a birth in the next time when the central cell is 0, it is the function $f(0,0,0,0,0,0,1,0,0) \rightarrow 0$. (b) Shows a condition where the sum of the neighbours permit a birth in the next time when the central cell is 0, it is the function $f(0,0,0,1,0,0,0,1,0) \rightarrow 1$. (c) Shows a condition where the sum of the neighbours permit a survival in the next time when the central cell is 1, it is the function $f(1,0,0,0,1,1,0,1,0) \rightarrow 1$. (d) Shows a condition where the sum of the neighbours do not permit a survival in the next time when the central cell is 1, it is the function $f(1,0,1,1,1,0,1,1,1) \rightarrow 0$.}
\label{MooreNeighborhood}
\end{figure}

Figure~\ref{MooreNeighborhood} illustrates some relations to get a birth, a survival and death, in the Moore neighbourhood. Cells in color white represent the symbol 0 and cells in color grey represent the symbol 1. Every relation determines the central cell in the present time with their neighbours and the value for the next time. The total number of relations is determined by $2^{2^9}$.

Once a resting lattice is perturbed, patterns of states 1 emerge, grow and propagate on the lattice (Fig.~\ref{signals}). Boolean values are represented by reaction of particles, positioned initially in the middle of channel, value 0 (Fig.~\ref{signals}a up), or slightly offset, value 1 (Fig.~\ref{signals}a down). The initial position of the particles determine outcomes of their reaction. Particle corresponding to the value 0 is transformed to a regular symmetric pattern (Fig.~\ref{signals}b,c,d up), similarly to `frozen' waves of excitation activity \cite{adamatzky2009hot}. Particle representing signal value 1 is transformed to transversally asymmetric patterns (Fig.~\ref{signals}b,c,d down). Both patterns propagate inside the channel with constant speed, advancing unit of the channel length per one step of a discrete time.

\begin{figure}[th]
\centering
\includegraphics[width=1.0\textwidth]{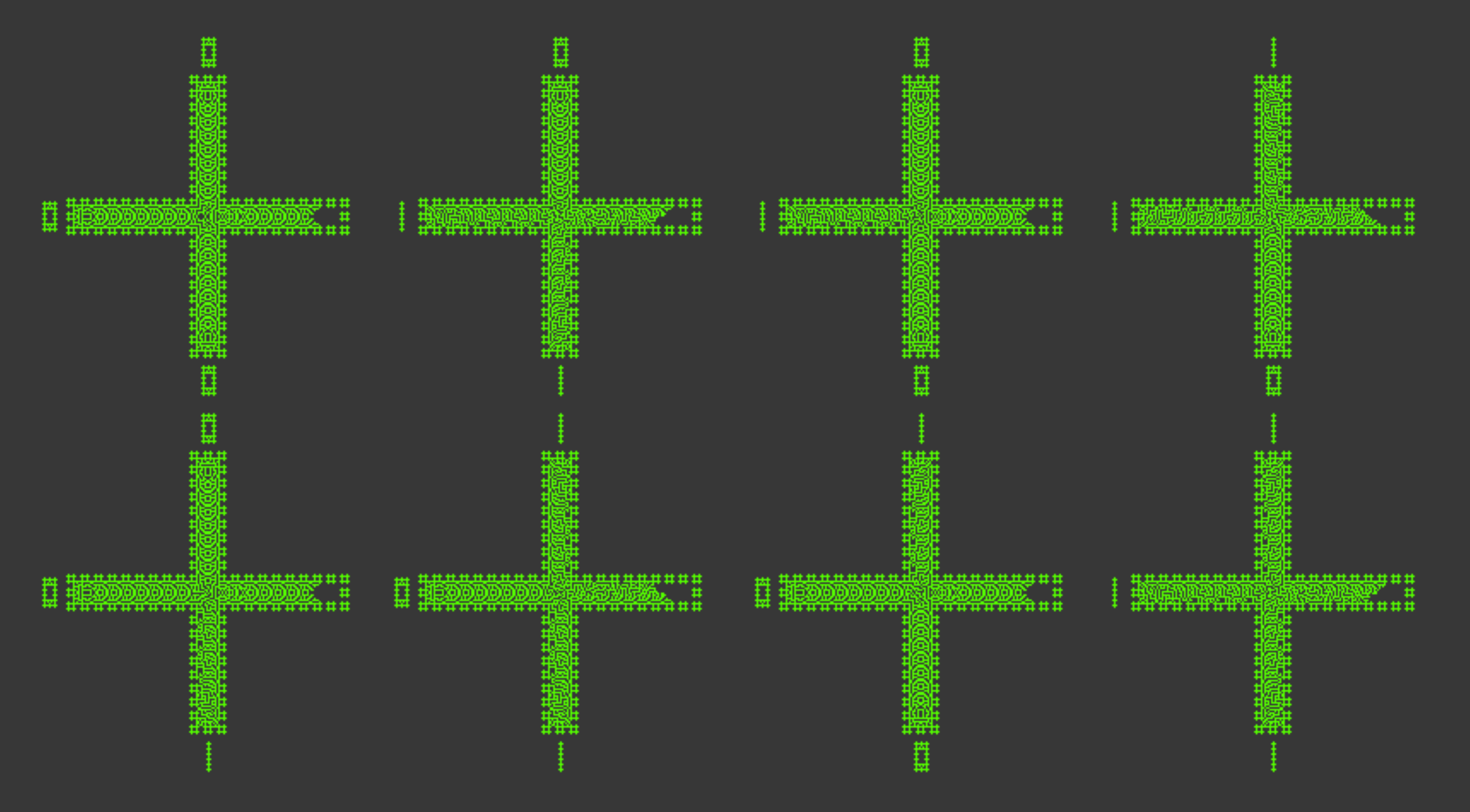}
\caption{Three-valued {\sc majority} gate implemented in the rule $B2/S2345$. From left to right: $MAJ(0,0,0)=0$, $MAJ(0,1,1)=1$, $MAJ(0,1,0)=0$, $MAJ(1,1,0)=1$, $MAJ(0,0,1)=0$, $MAJ(1,0,1)=1$, $MAJ(1,0,0)=0$, $MAJ(1,1,1)=1$.}
\label{majorityGate}
\end{figure}

When patterns, representing values 0 and 1, meet at $T$-junctions they compete for the output channel. Depending on initial distance between particles, one of the patterns wins and propagates along the output channel. By controlling these signals we have implemented a number of logic gates, such as: {\sc and}, {\sc or}, {\sc not}, {\sc fanout}, {\sc delay}, {\sc majority} \cite{martinez2010computation}. Figure~\ref{majorityGate} shows how a {\sc majority} gate has been implemented in the function $B2/S2345$. The {\sc majority} gate can be defined given three inputs $a$, $b$, and $c$ \cite{minsky1967computation}, as follows: $MAJ(a,b,c) = (a \wedge b) \vee (a \wedge c) \vee (b \wedge c)$.

Implementation of {\sc majority} gate is shown in Fig.~\ref{majorityGate}. The gate has three inputs: North, West and South channels, and one output: East channel. Three propagating pattern, which represent inputs, collide at the cross-junction of the gate. The resultant pattern is recorded at the output channel.

In other fields, the majority gates have been selected to design circuits in quantum-dot cellular automata~\cite{snider1999quantum}. On the other hand, Fischer {\it et al.} in \cite{fischer2017experimental} have implemented a spin-wave majority on an electromagnetic device.

\section{Cubelets computing}
\label{cubeletscomputing}

\subsection{Propagating signals with Cubelets}

\begin{figure}[th]
\begin{center}
\subfigure[]{\scalebox{0.48}{\includegraphics{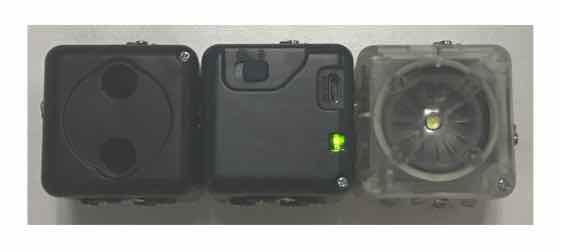}}} \hspace{2.9cm}
\subfigure[]{\scalebox{0.48}{\includegraphics{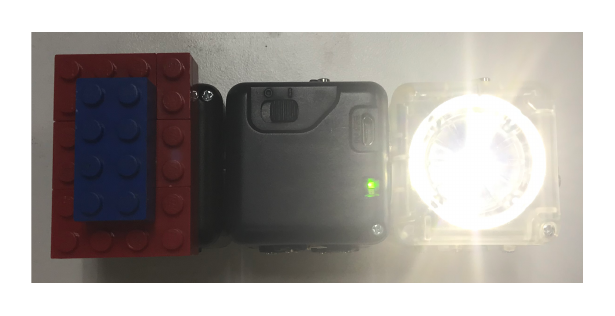}}} \hspace{1.5cm}
\subfigure[]{\scalebox{0.44}{\includegraphics{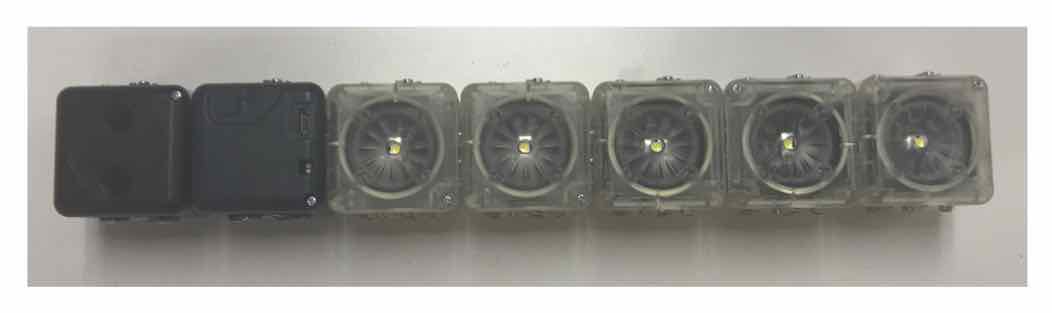}}}
\subfigure[]{\scalebox{0.443}{\includegraphics{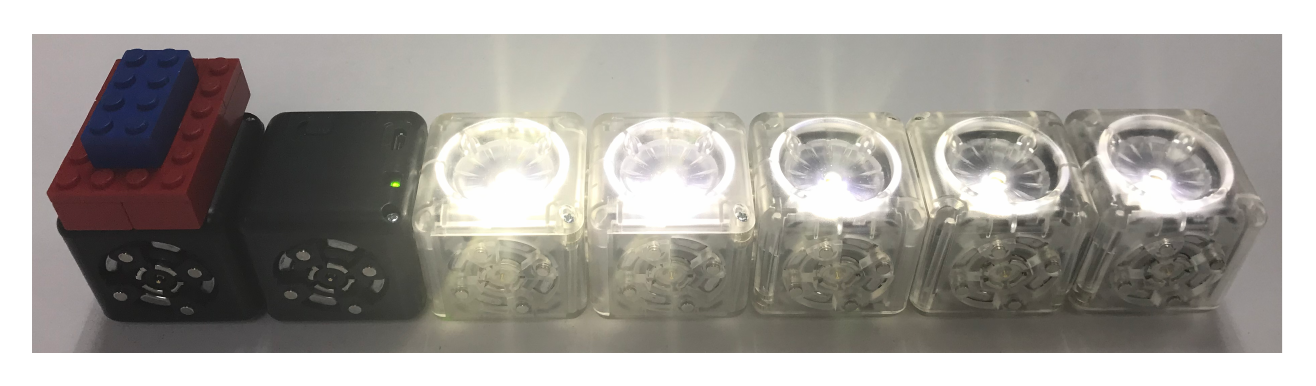}}}
\end{center}
\caption{Activating and propagating signals with Cubelets. Binary values are represented if the lightness Cubelet is in state {\sf off} (as 0) and in state {\sf on} (as 1). The high intensity represents the pattern as 1 and any other intensity as 0. This way, to conserve a high intensity we need to concatenate more battery Cubelets with one or two lightness Cubelets. The binary values are initialized by the value of a distance Cubelet (input data) and the result will be indicated by the value of lightness Cubelets (output data).}
\label{CubeletsLineSimple}
\end{figure}

The transmission of signals in Cubelets is across electrical current, values are traveling when Cubelets are concatenated. This way, the distance Cubelet ($di$ in Tab.~\ref{symbolCubeletsTable}) produces a positive value (255 maximum) if an object is near to it and close to 0 (minimum) if the object is far.\footnote{Cubelets API Documentation. \url{http://api.modrobotics.com/cubelets/index.html}} Figure~\ref{CubeletsLineSimple} displays how we can propagate and recognise signals with Cubelets robots. Figure~\ref{CubeletsLineSimple}a shows the configuration to construct a {\it lamp robot} $w = di \cdot ba \cdot fl$ and in Fig.~\ref{CubeletsLineSimple}b we activate the distance cube that transmit an integer value to the rest of cubes yielding which the lightness cube change of state {\sf off} to {\sf on}. Consequently to propagate this signal we need to concatenate more lightness cubes and produce the configuration $di \cdot ba \cdot (fl)^*$. Figure~\ref{CubeletsLineSimple}c shows the initial configuration in state {\sf off} and Fig.~\ref{CubeletsLineSimple}d shows the configuration in state {\sf on}. We will note that the intensity of light is strong just in the first two cubes whiles the illumination in the rest of the cubes is decreasing in intensity.  Therefore, we consider just high intensity as value 1 and 0 in any other case. The regular expression to expand the signal should include a battery cube $di \cdot (ba \cdot fl)^*$.

\subsection{Constructing circuits with Cubelets}

Now let us show how to construct logic gates from Cubelets using the Life-like rule $B2/S2345$. The first and crucial element in our construction is the implementation of a {\sc majority} gate. Following the same topology as in the Life-like rule $B2/S2345$ (see Fig.~\ref{majorityGate}) the {\sc majority} gate is defined as the robot:

\begin{eqnarray*}
w_{MAJgateI} = ({\bf B}^4 \cdot ba_{(4,0,0)}^{(F,N,W)}) \cdot ({\bf B}^4 \cdot di_{(4,1,0)}^{(F,N,W)}) \cdot ({\bf B}^4 \cdot fl_{(4,2,0)}^{(F,N,W)}) \cdot \\ ({\bf B}^4 \cdot fl_{(4,3,0)}^{(F,N,W)}) \cdot (ba_{(0,4,0)}^{(F,N,W)} \cdot di_{(1,4,0)}^{(F,N,W)} \cdot fl_{(2,4,0)}^{(F,N,W)} \cdot fl_{(3,4,0)}^{(F,N,W)} \cdot fl_{(4,4,0)}^{(W,N,F)}) \cdot \\ ({\bf B}^4 \cdot fl_{(4,5,0)}^{(F,N,W)}) \cdot ({\bf B}^4 \cdot fl_{(4,6,0)}^{(F,N,W)}) \cdot ({\bf B}^4 \cdot di_{(4,7,0)}^{(F,N,W)}) \cdot ({\bf B}^4 \cdot ba_{(4,8,0)}^{(F,N,W)}).
\end{eqnarray*}

So, the complementary robot necessary to process the output is defined by the next expression:

\begin{eqnarray*}
w_{MAJgateO} = {\bf B} \cdot br_{(6,4,0)}^{(W,N,F)} \cdot fl_{(7,4,0)}^{(F,N,W)} \cdot fl_{(8,4,0)}^{(F,N,W)} \cdot ba_{(9,4,0)}^{(F,N,W)}.
\end{eqnarray*}

For design a {\sc not majority} gate it is constructed by adding an inverse Cubelet in the output wire. It is defined by the next expression::

\begin{eqnarray*}
w_{NMAJgateO} = {\bf B} \cdot br_{(6,4,0)}^{(W,N,F)} \cdot in_{(7,4,0)} \cdot fl_{(8,4,0)}^{(F,N,W)} \cdot fl_{(9,4,0)}^{(F,N,W)} \cdot ba_{(10,4,0)}^{(F,N,W)}.
\end{eqnarray*}

\begin{figure}
\begin{center}
\subfigure[]{\scalebox{0.083}{\includegraphics{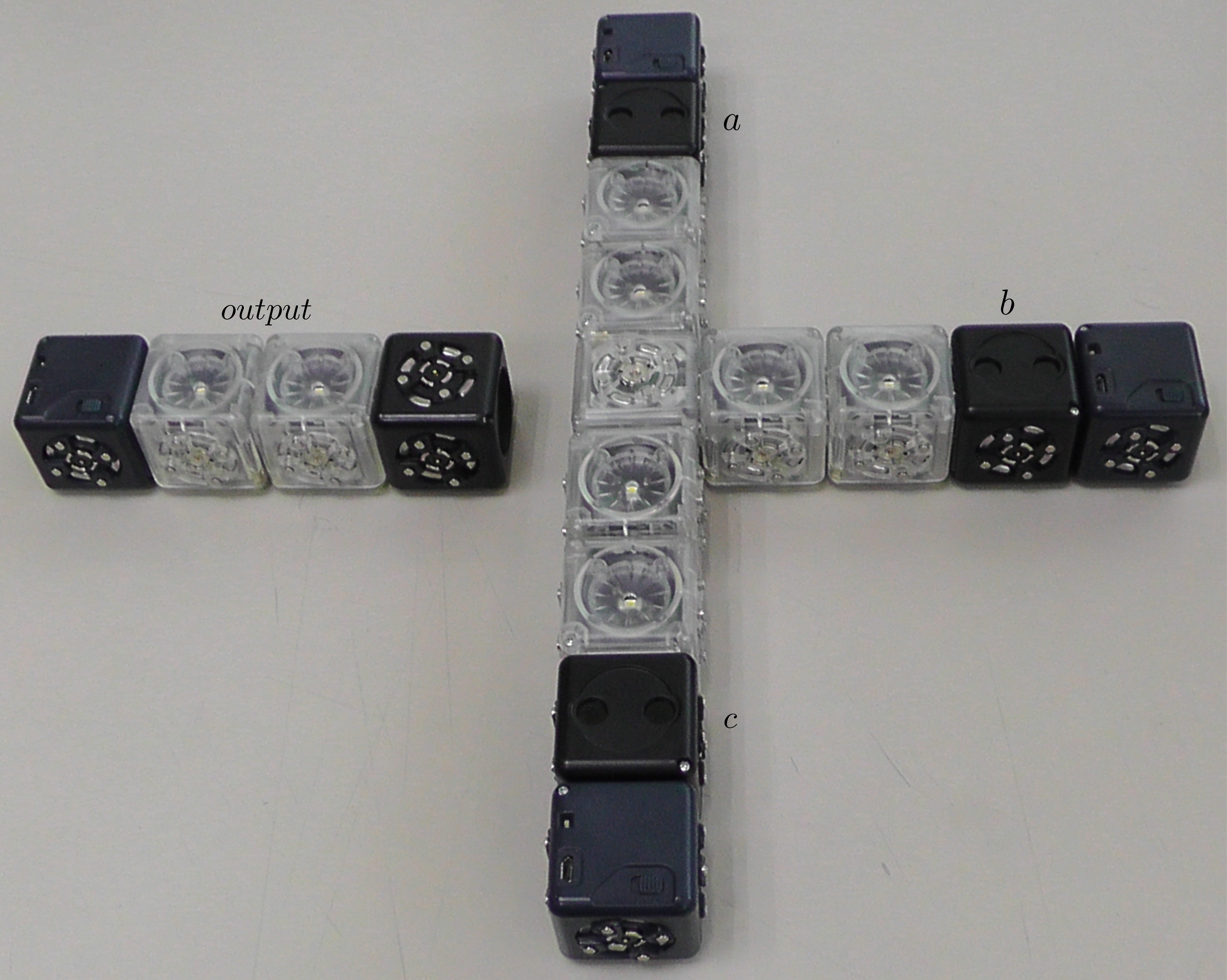}}} \hspace{0.4cm}
\subfigure[]{\scalebox{0.172}{\includegraphics{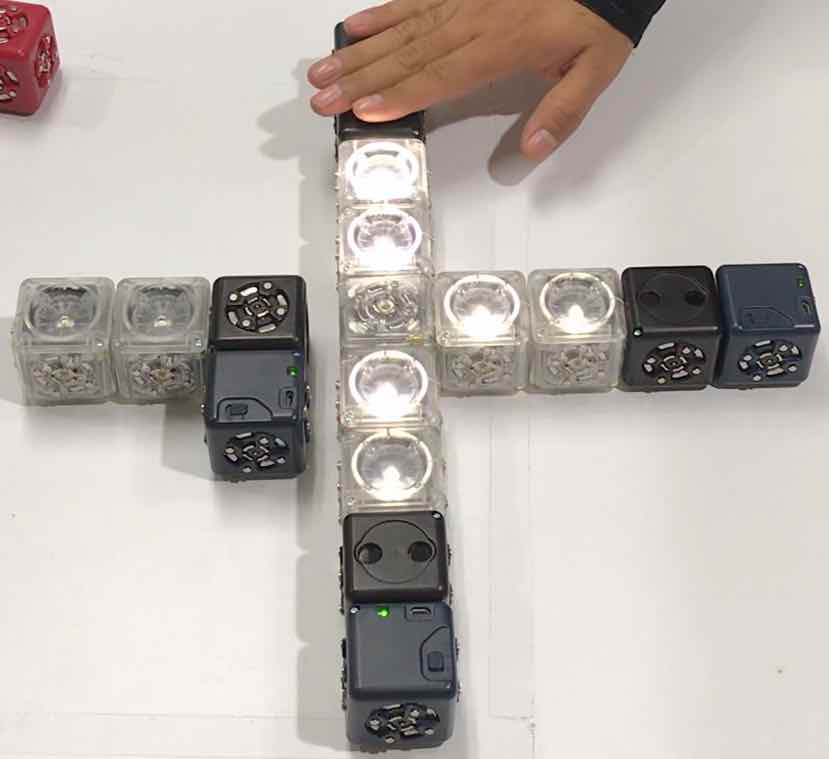}}} \hspace{0.8cm}
\subfigure[]{\scalebox{0.081}{\includegraphics{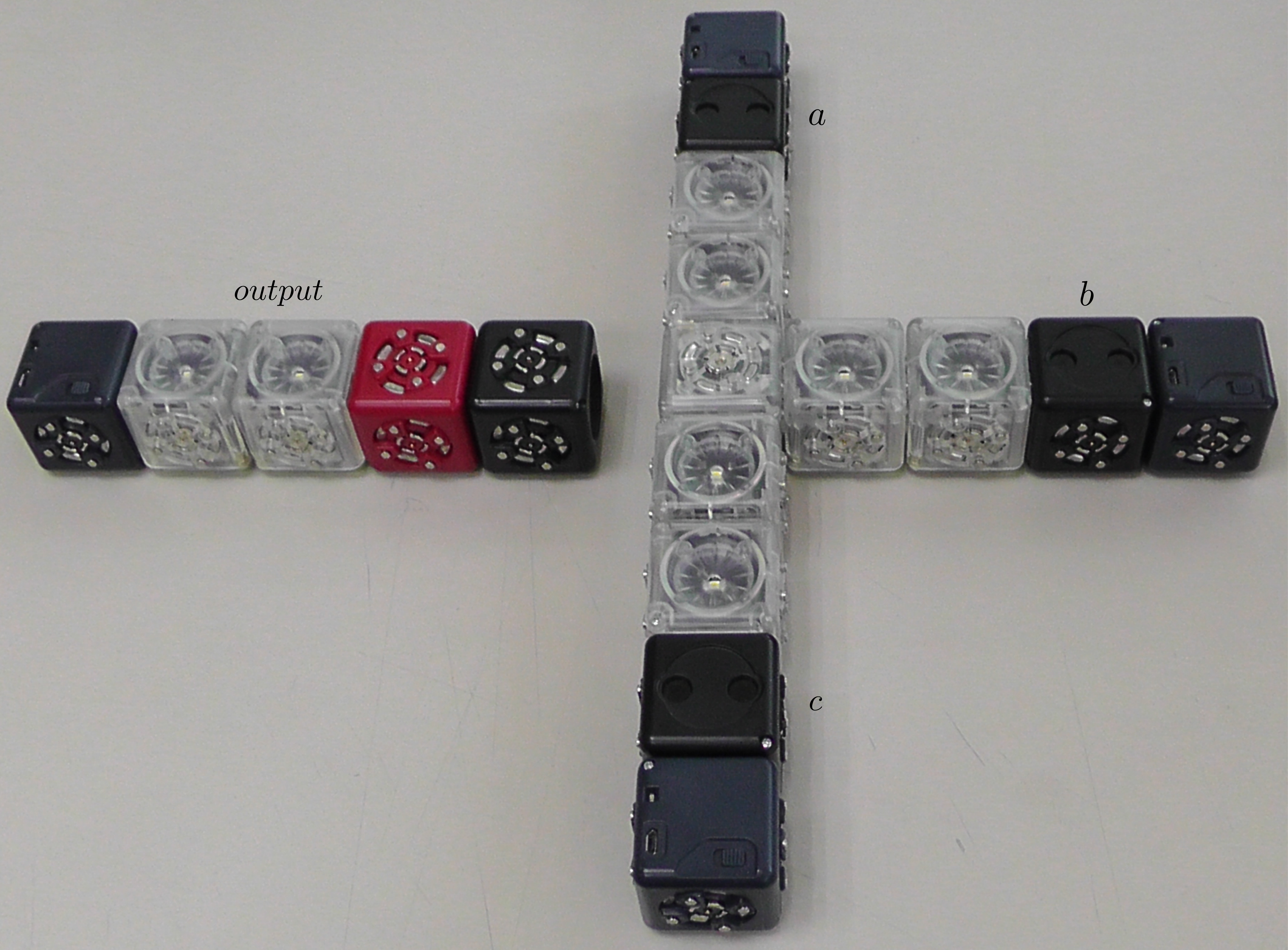}}} 
\subfigure[]{\scalebox{0.163}{\includegraphics{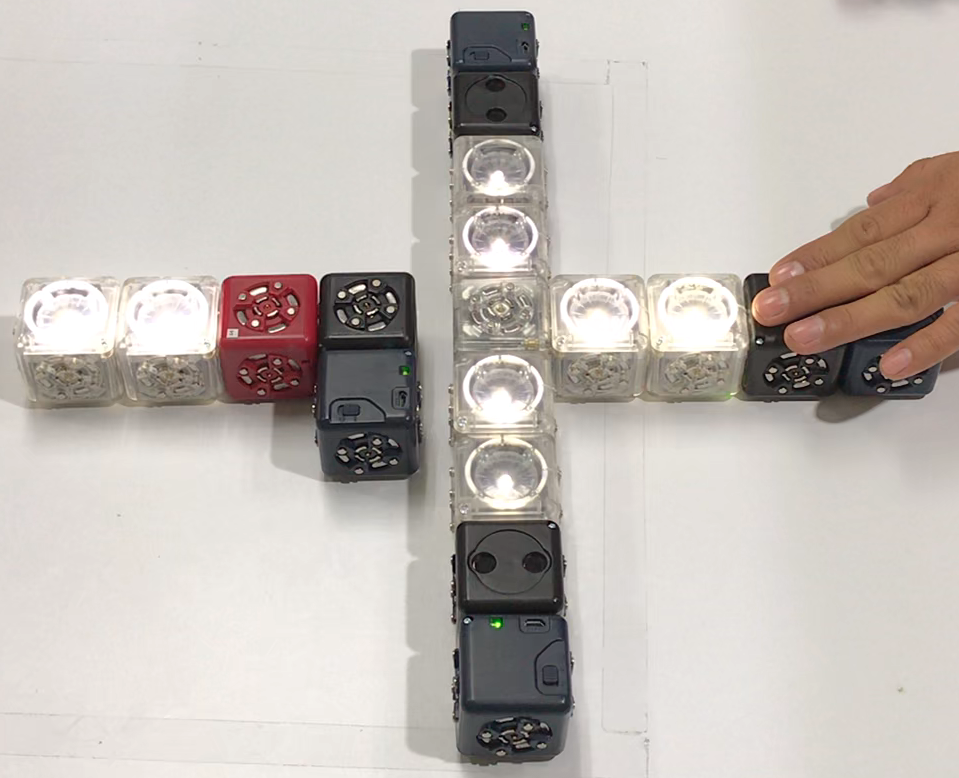}}}
\end{center}
\caption{Configurations of Cubelets robots that simulate a {\sc majority} gate $w_{MAJ}$. (a) Displays the main configuration. (b) Shows an operation in $w_{MAJ}$ with input $a=1$, $b=0$, and $c=0$. (b) Displays the configuration for a {\sc not majority} gate $w_{NMAJ}$. (c) Shows an operation in $w_{NMAJ}$ with input $a=0$, $b=1$, $c=0$.}
\label{CubeletsMAJgate}
\end{figure}

Figure~\ref{CubeletsMAJgate} shows the position of each Cubelet encoded in the strings $w_{MAJ}$ and $w_{NMAJ}$. The configuration works identically as cellular automaton rule $B2/S2345$. We have three input channels (North, East, South) and one output (West).\footnote{A video showing all these operations is available in: \url{https://youtu.be/v6O4-1iGDk0}. The source code to implement the {\sc MAJ} gate is available in: \url{https://gist.github.com/RQF7/87b89a3cf21bf2794a77112ea2bb6542}.}

\begin{figure}[th]
\centering
\includegraphics[width=1\textwidth]{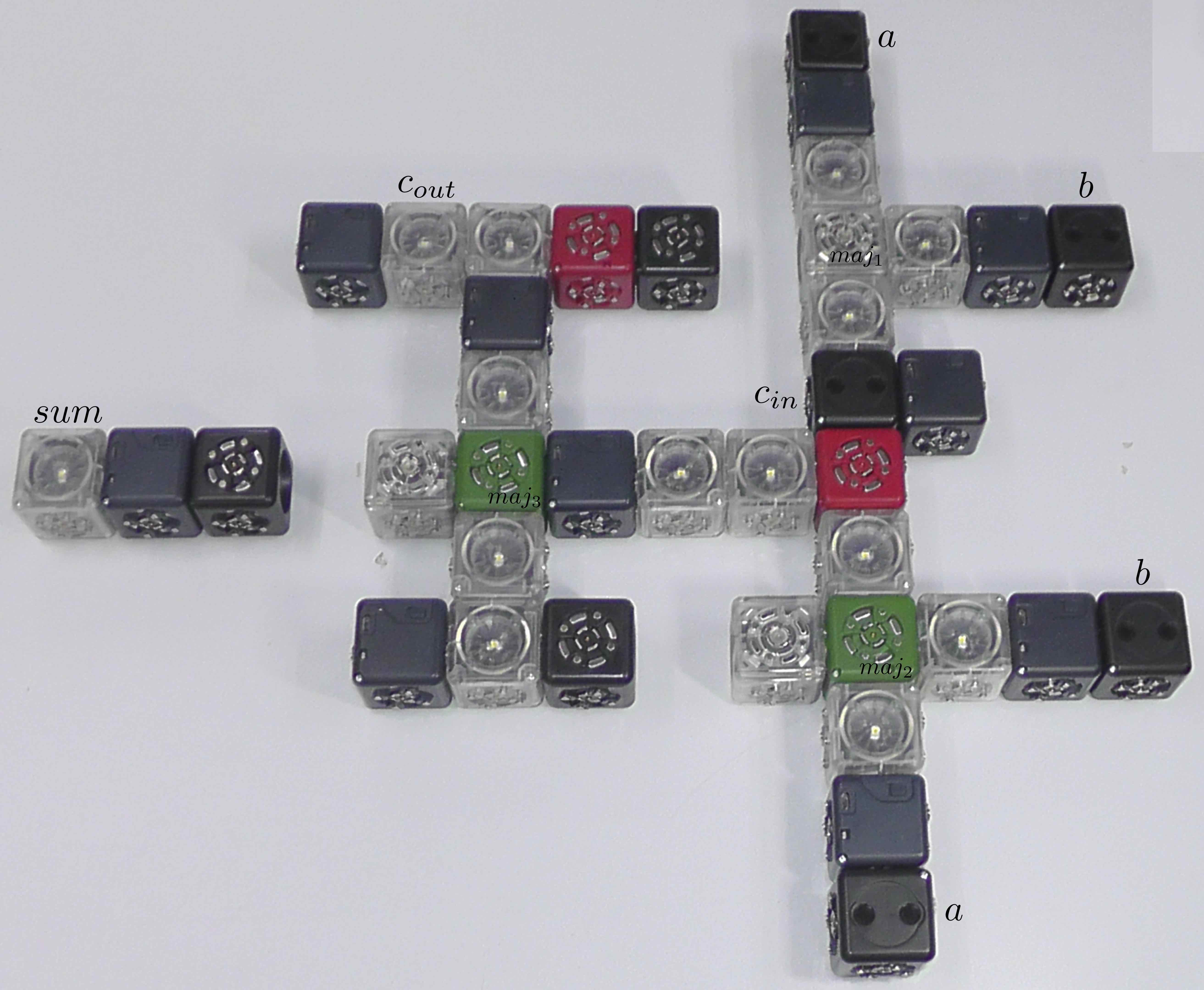}
\caption{A binary full adder made of Cubelets.}
\label{CubeletsCircuitAdder1}
\end{figure}

Logical universality with Cubelets across this {\sc majority} gate is shown as follows. If we fix input $c =0$ then we get the {\sc and} gate and for $c=1$ we get the {\sc or} gate (Tabs.~\ref{and_table} and~\ref{or_table}, respectively). To get a {\sc nand} or {\sc nor} gate we concatenate an inverse Cubelet (see Tab.~\ref{symbolCubeletsTable}), that represents the {\sc not} gate.

\begin{table*}[th]

\setlength{\tabcolsep}{7pt}
\begin{minipage}{0.5\linewidth}
\centering
\caption{{\sc and} gate, $c=0$}
\label{and_table}

\medskip
\begin{tabular}{ccc|c}
\toprule
$a$ & $b$ & $c$ & \\
\midrule
0 & 0 & . & 0 \\
0 & 1 & . & 0 \\
1 & 0 & . & 0 \\
1 & 1 & . & 1 \\
\bottomrule
\end{tabular}   

\end{minipage}\hfill
\begin{minipage}{.5\linewidth}
\centering

\caption{{\sc or} gate, $c=1$}
\label{or_table}

\medskip

\begin{tabular}{ccc|c}
\toprule
$a$ & $b$ & $c$ & \\
\midrule
0 & 0 & . & 0 \\
0 & 1 & . & 1 \\
1 & 0 & . & 1 \\
1 & 1 & . & 1 \\
\bottomrule
\end{tabular}   
\end{minipage}\hfill

\end{table*}

\begin{figure}[th]
\begin{center}
\subfigure[]{\scalebox{0.48}{\includegraphics{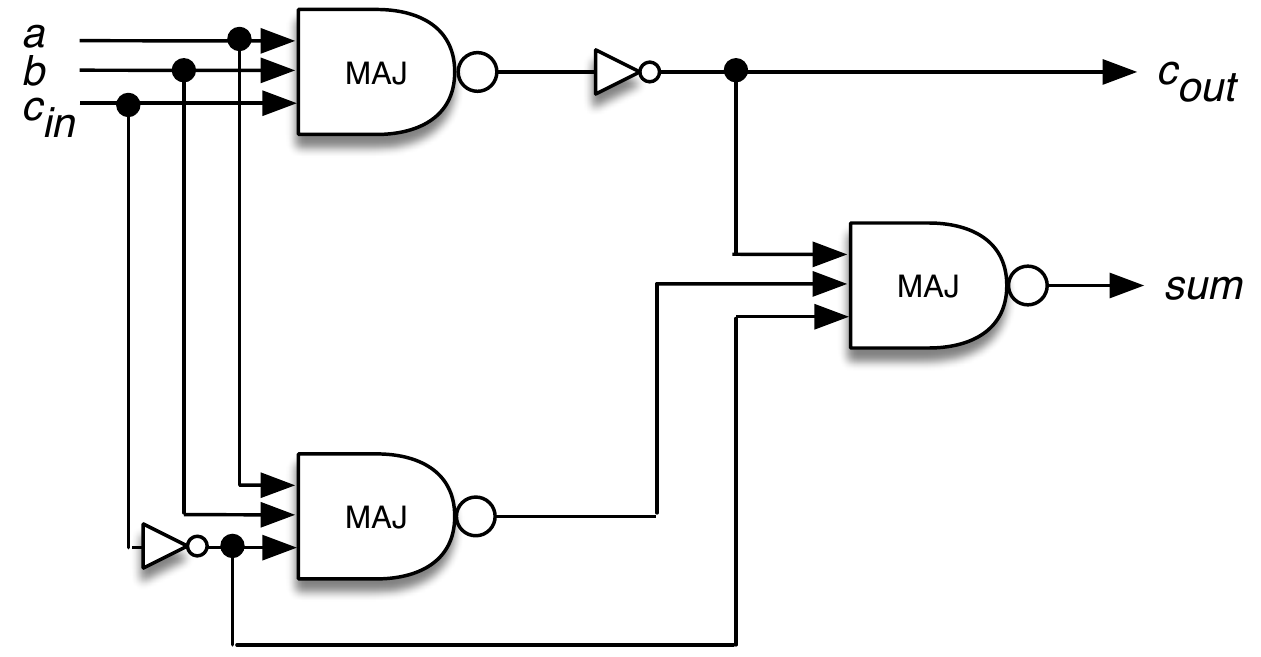}}} \hspace{0.2cm}
\subfigure[]{\scalebox{0.37}{\includegraphics{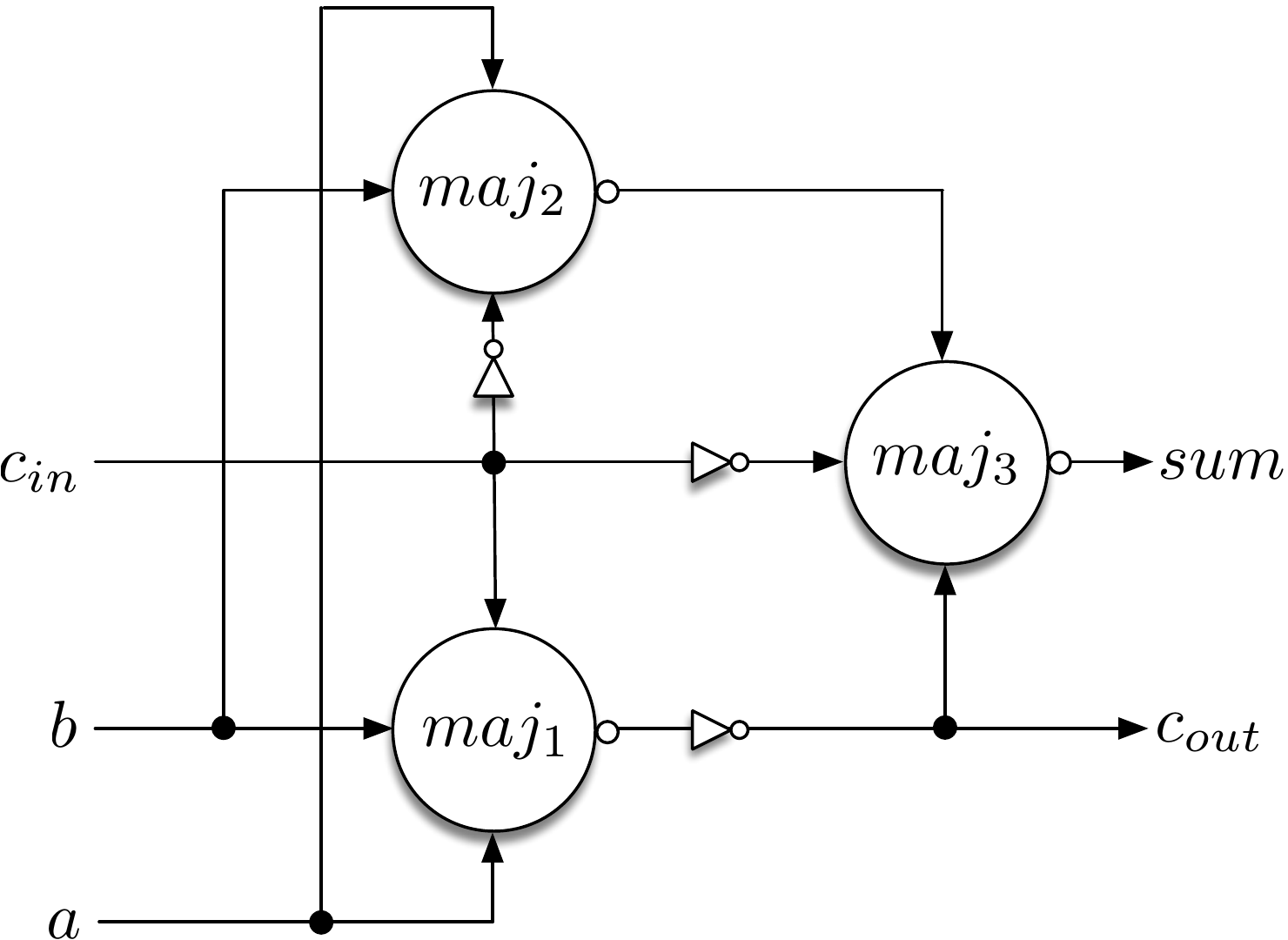}}}
\end{center}
\caption{(a)~A circuit to implement a binary full adder based in {\sc not majority} gates. (b) The diagram for a full binary adder adapted for works with Cubelets robots.}
\label{AdderCircuit}
\end{figure}

Now is time to design a binary full adder robotic. We follow the same idea which was used in the Life-like rule $B2/S2345$ \cite{martinez2010computation, martinez2010majority}. We use a circuit where a binary full adder can be constructed with three {\sc not majority} gates and two {\sc not} gates, illustrated in Fig.~\ref{AdderCircuit}a. To handle this circuit with Cubelets we need to preserve the routes of every signal. The Fig.~\ref{AdderCircuit}b shows how we will propagate each signal. Every {\sc not majority} gate is labelled to recognise better each partial result and signals may be distributed in parallel. Finally, the Tab.~\ref{AdderTable} displays each input, partial outputs (stages of {\sc not majority} gates) and the final results (sum and carry out).

\begin{table}[th]
\caption{Truth table  of the binary full adder, show in Fig.~\ref{AdderCircuit}.}
\label{AdderTable}
\centering
\begin{tabular}{|c|c|c|c|c|c|c|c|}
\hline
$a$ & $b$ & $c_{in}$ & $\sim maj_1$ & $\sim maj_2$ & $\sim maj_3$ & $c_{out}$ & $sum$ \\
\hline \hline
0 & 0 & 0 & 1 & 1 & 0 & 0 & 0 \\
0 & 0 & 1 & 1 & 1 & 1 & 0 & 1 \\
0 & 1 & 0 & 1 & 0 & 1 & 0 & 1 \\
0 & 1 & 1 & 0 & 1 & 0 & 1 & 0 \\
1 & 0 & 0 & 1 & 0 & 1 & 0 & 1 \\
1 & 0 & 1 & 0 & 1 & 0 & 1 & 0 \\
1 & 1 & 0 & 0 & 0 & 0 & 1 & 0 \\
1 & 1 & 1 & 0 & 0 & 1 & 1 & 1 \\
\hline
\end{tabular}
\end{table}

\begin{figure}
\begin{center}
\subfigure[]{\scalebox{0.026}{\includegraphics{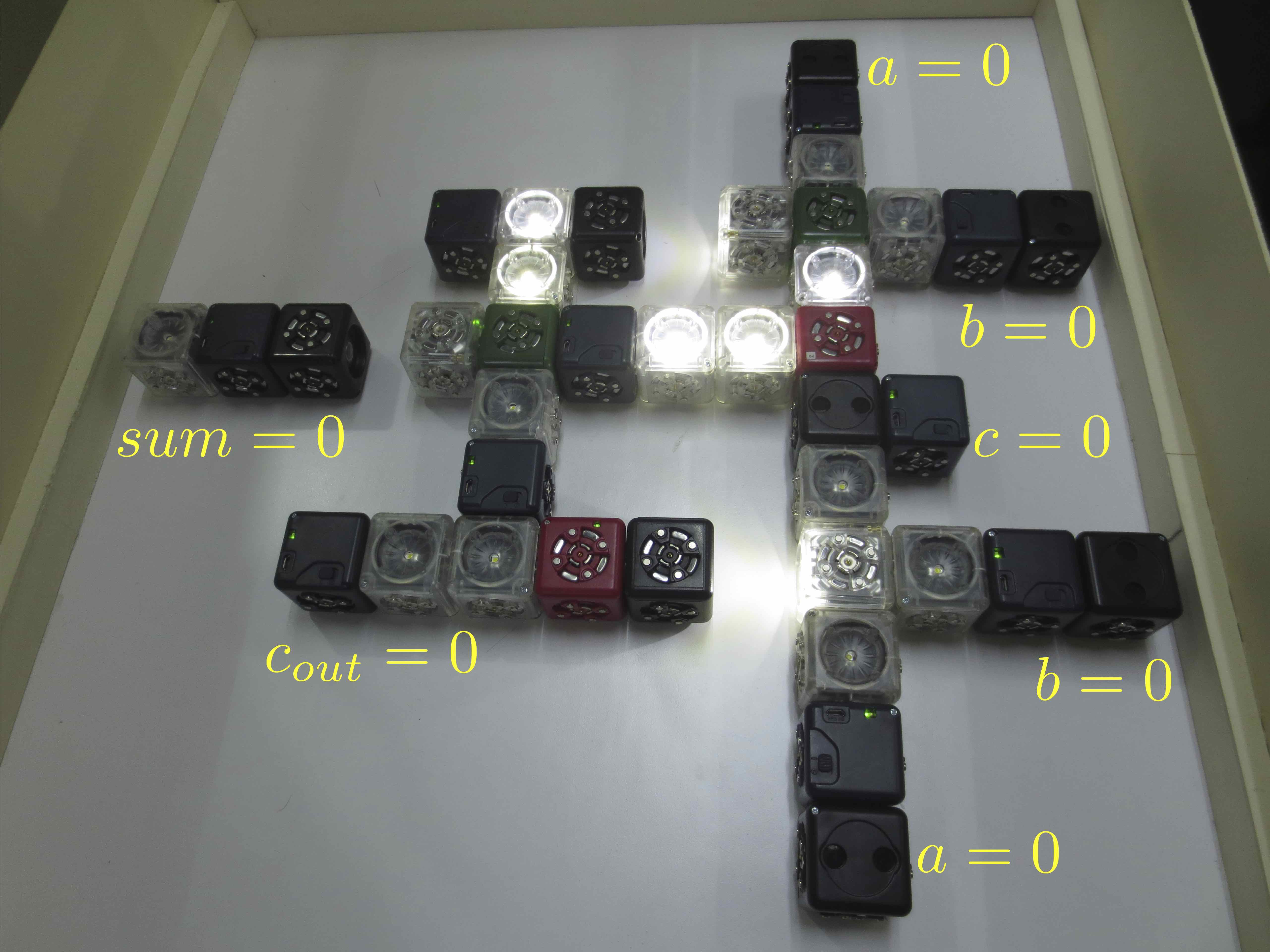}}} \hspace{0.7cm}
\subfigure[]{\scalebox{0.026}{\includegraphics{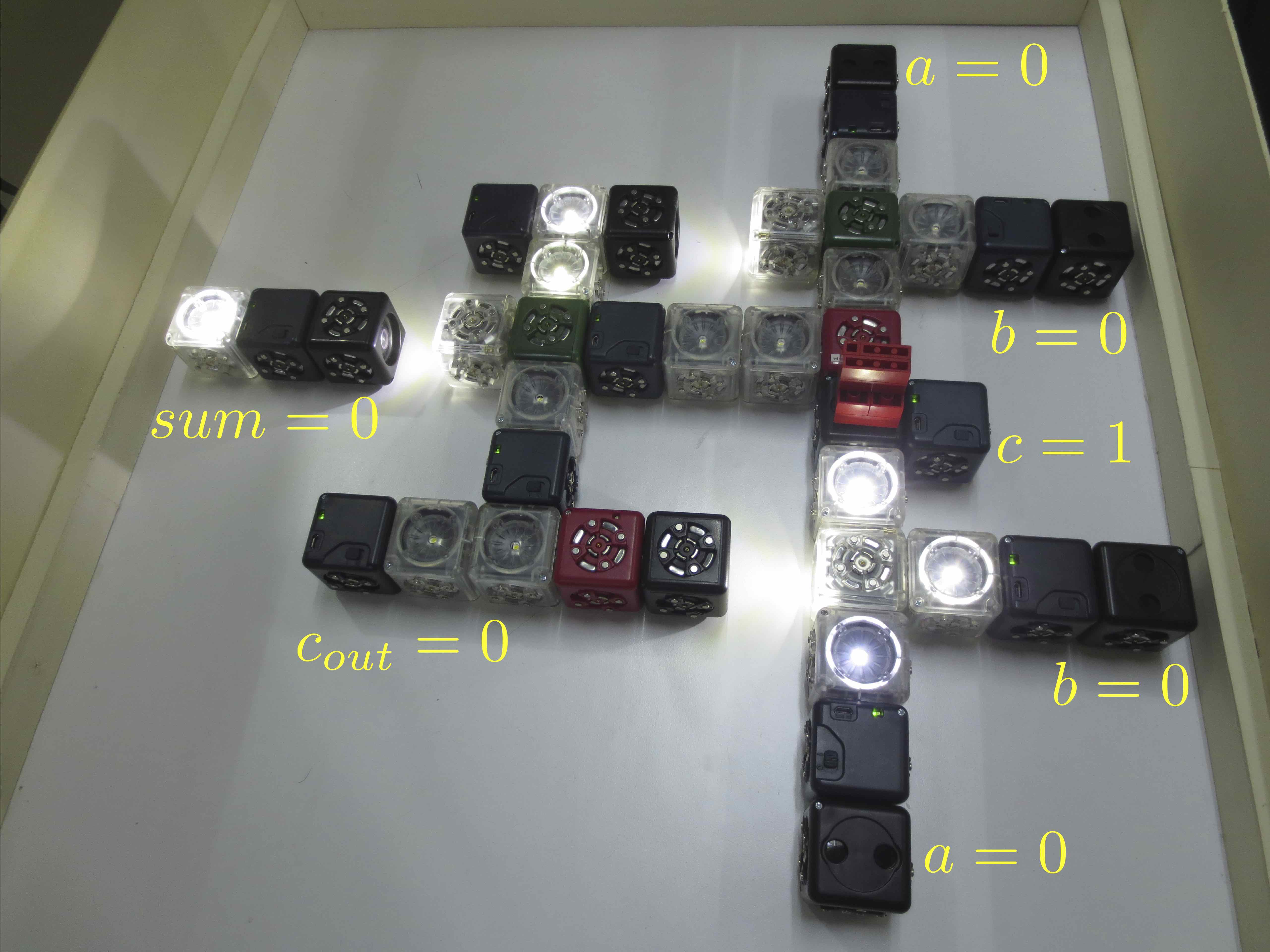}}} \hspace{0.7cm}
\subfigure[]{\scalebox{0.026}{\includegraphics{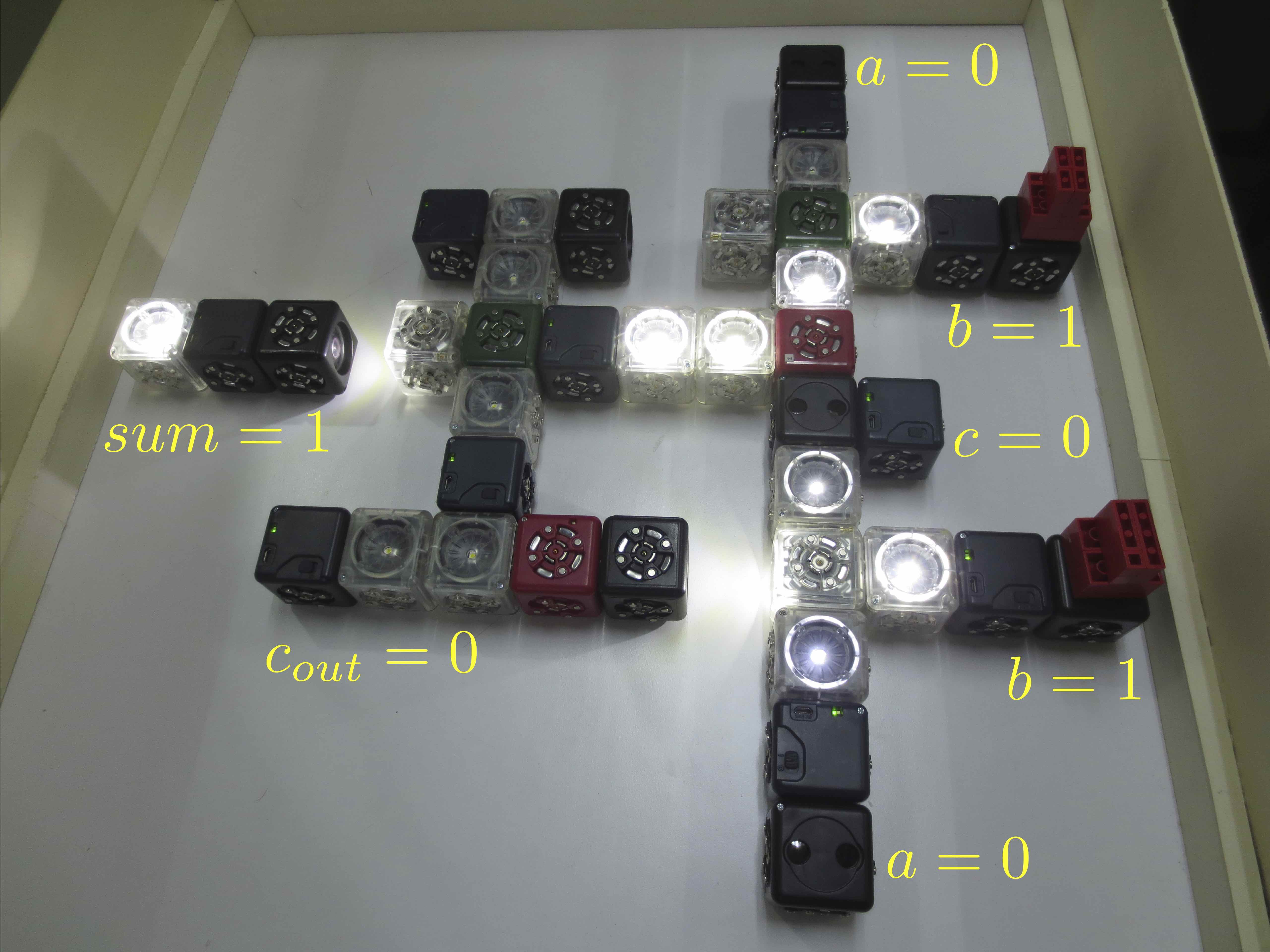}}} \hspace{0.7cm}
\subfigure[]{\scalebox{0.026}{\includegraphics{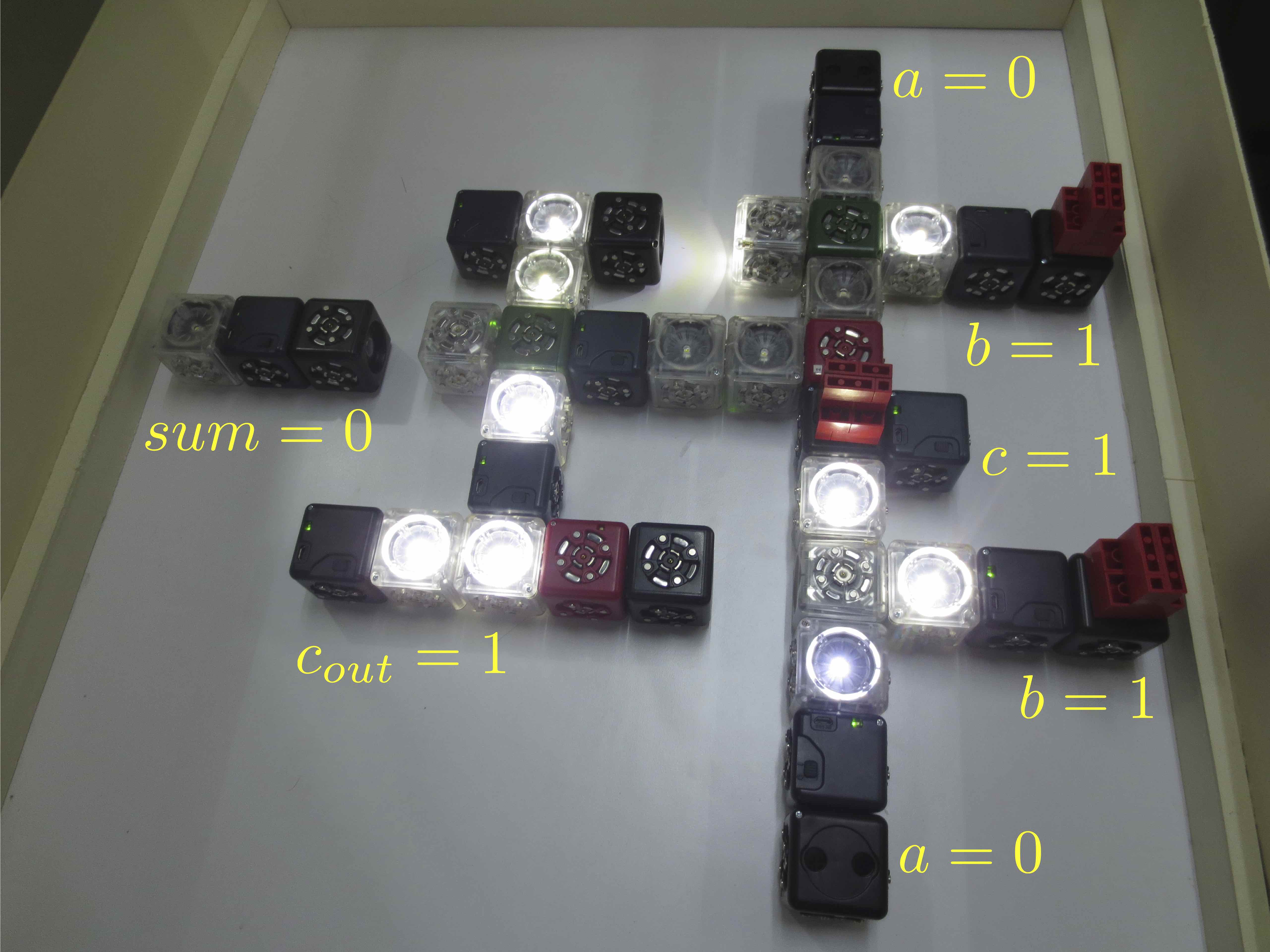}}} \hspace{0.7cm}
\subfigure[]{\scalebox{0.026}{\includegraphics{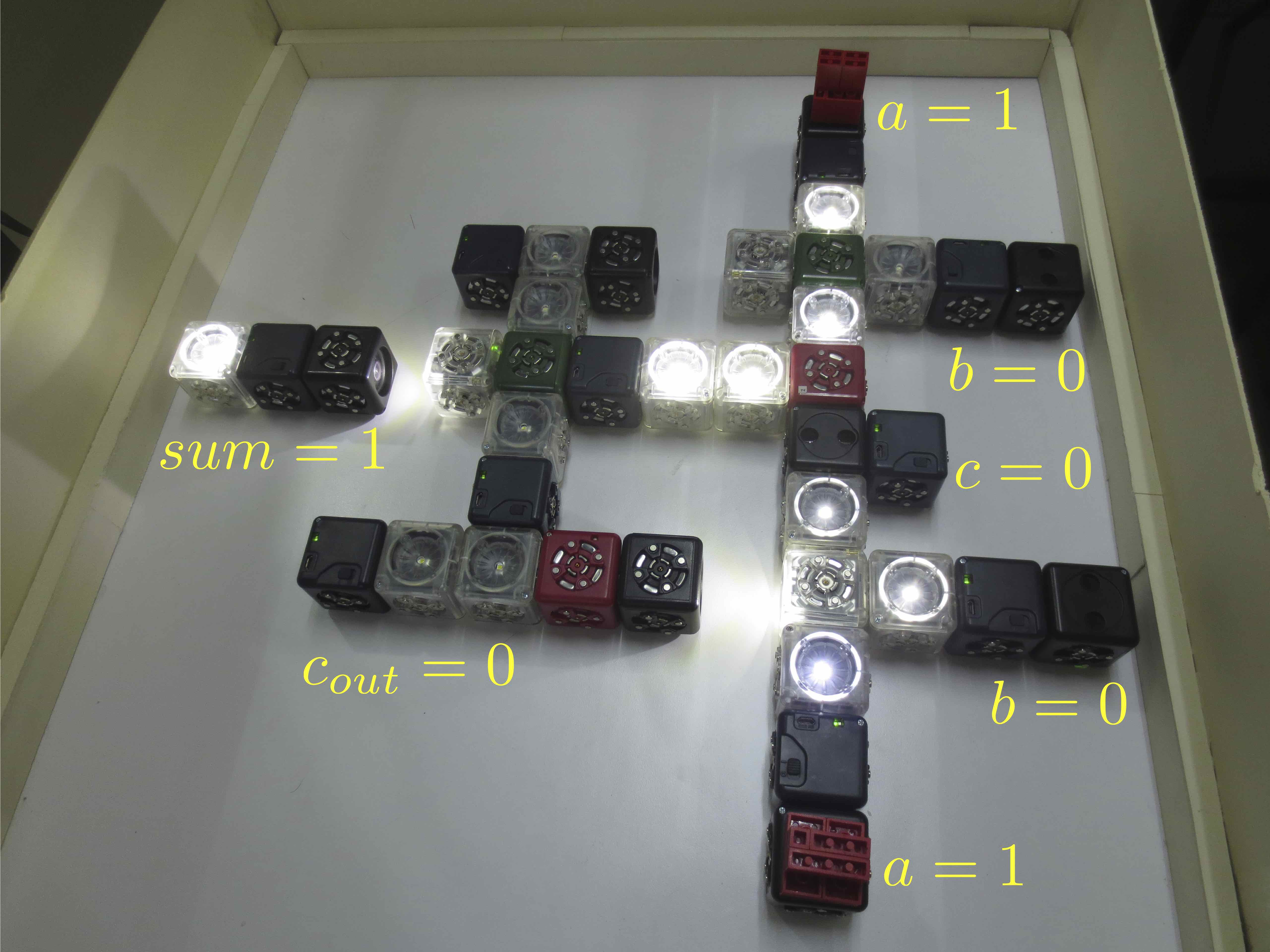}}} \hspace{0.7cm}
\subfigure[]{\scalebox{0.026}{\includegraphics{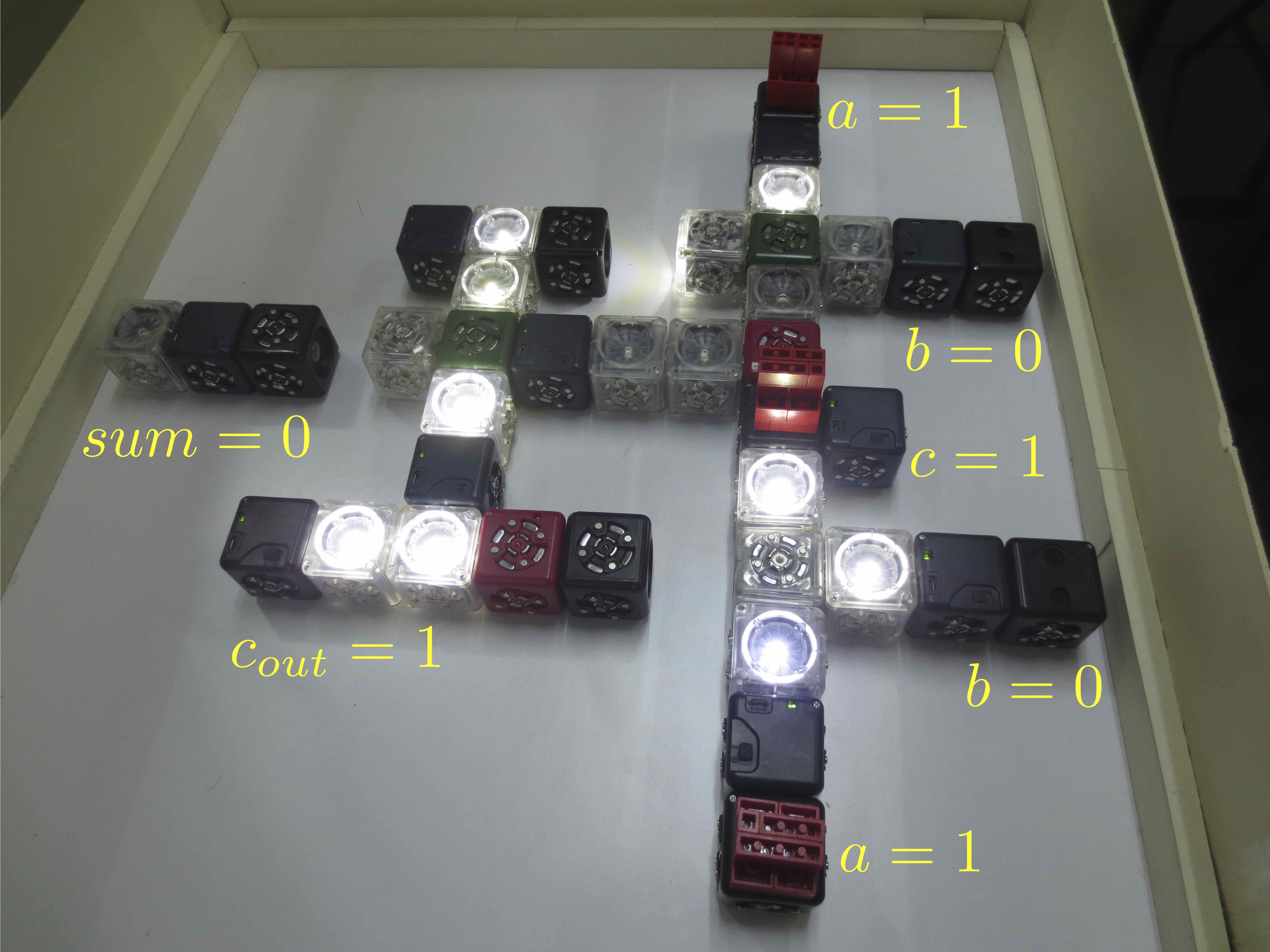}}} \hspace{0.7cm}
\subfigure[]{\scalebox{0.026}{\includegraphics{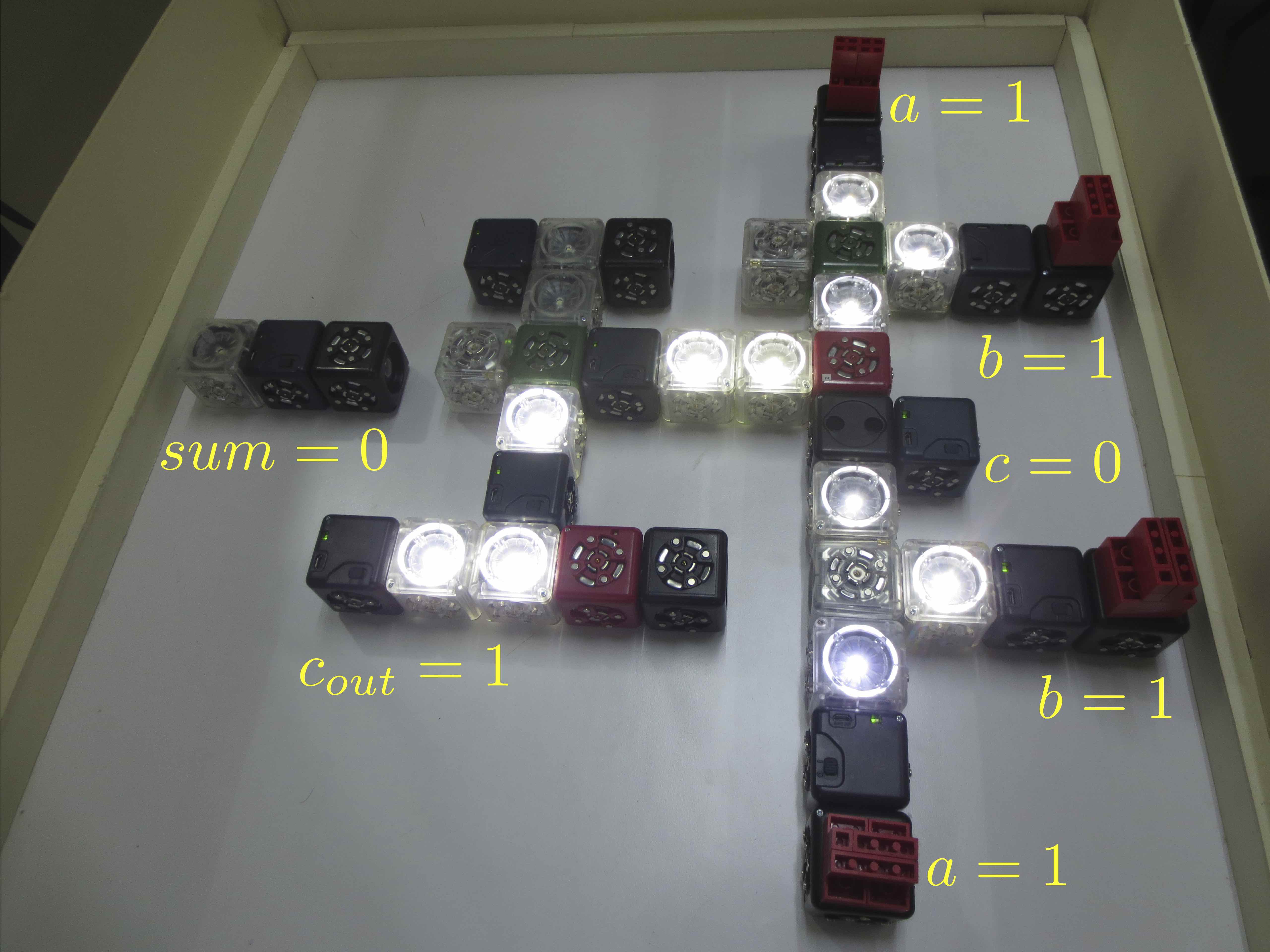}}} \hspace{0.7cm}
\subfigure[]{\scalebox{0.026}{\includegraphics{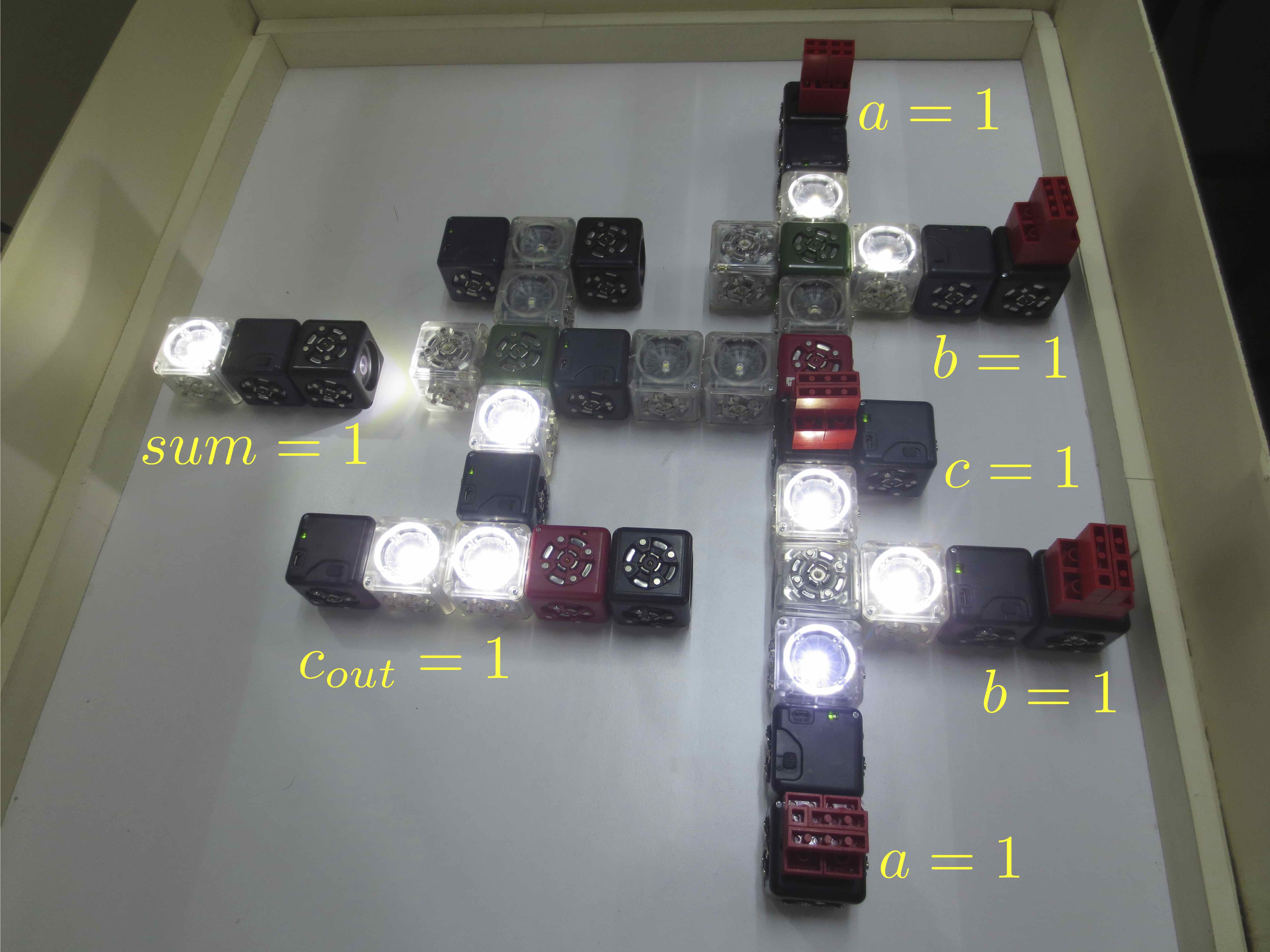}}}
\end{center}
\caption{Binary full adder $w_{BA}$ for all inputs and outputs. We show every input and its results. (a) $a=0$, $b=0$, $c=0$, (b) 001, (c) 010, (d) 011, (e) 100, (f) 101, (g) 110, and (h) 111.}
\label{binaryaddercubelet1}
\end{figure}

Figure~\ref{CubeletsCircuitAdder1} provides a detailed view of the robot that represents the circuit simulating the full binary adder with Cubelets. The robot is assembled of 39 Cubelets (17 flashlight, 10 battery, 5 distance, 3 brightness, 2 inverse, and 2 passive).  The nature of Cubelets is preserved almost completely because you need just reprogram two Cubelets to implement the adder: the input (flashlight cube) and the output (brightness cube). The {\sc not majority} gate is implemented in the self cube and thus we avoid to use extra inverse cubes. The code\footnote{The source code for $w_{BA}$ is available from: \url{https://gist.github.com/RQF7/87b89a3cf21bf2794a77112ea2bb6542\#file-sensor-c}} for the flashlight cube ($maj_1$ in Fig.~\ref{CubeletsCircuitAdder1}) and brightness cube is simple. If the flashlight has a value higher than $(maximum/2)$ then the value is equal to zero and maximum in any other case. Also, this {\sc not majority} gate is implemented in two passive cubes ($maj_2$ and $maj_3$ in Fig.~\ref{CubeletsCircuitAdder1}).

Finally, we have the string $w_{BA}$ which defines a binary full adder with Cubelets:

\begin{eqnarray*}
w_{BA}  =  ({\bf B}^3 \cdot di_{(3,0,0)}^{(F,N,W)}) \cdot ({\bf B}^3 \cdot ba_{(3,1,0)}^{(F,N,W)}) \cdot ({\bf B}^3 \cdot fl_{(3,2,0)}^{(F,N,W)}) \cdot (di_{(0,3,0)}^{(F,N,W)} \cdot \\ ba_{(1,3,0)}^{(F,N,W)} \cdot fl_{(2,3,0)}^{(F,N,W)} \cdot bo_{(3,3,0)} \cdot fl_{(4,3,0)}^{(W,N,F)} \cdot {\bf B} \cdot br_{(6,3,0)}^{(E,N,B)} \cdot fl_{(7,3,0)}^{(F,N,W)} \cdot  ba_{(8,3,0)}^{(F,N,W)}) \cdot \\ ({\bf B}^3 \cdot fl_{(3,4,0)}^{(F,N,W)} \cdot {\bf B}^3 \cdot fl_{(7,4,0)}^{(F,N,W)}) \cdot (in_{(3,5,0)} \cdot fl_{(4,5,0)}^{(F,N,W)} \cdot fl_{(5,5,0)}^{(F,N,W)} \cdot ba_{(6,5,0)}^{(F,N,W)} \cdot \\ bo_{(7,5,0)} \cdot fl_{(8,5,0)}^{(W,N,F)} \cdot {\bf B} \cdot br_{(10,5,0)}^{(E,N,B)} \cdot ba_{(11,5,0)}^{(F,N,W)} \cdot fl_{(12,5,0)}^{(F,N,W)}) \cdot ({\bf B}^2 \cdot ba_{(2,6,0)}^{(F,N,W)} \cdot \\ di_{(3,6,0)}^{(F,N,W)} \cdot {\bf B}^3 \cdot fl_{(5,6,0)}^{(F,N,W)}) \cdot ({\bf B}^3 \cdot fl_{(3,7,0)}^{(F,N,W)} \cdot {\bf B}^3 \cdot ba_{(7,7,0)}^{(F,N,W)}) \cdot (di_{(0,8,0)}^{(F,N,W)} \cdot \\ ba_{(1,8,0)}^{(F,N,W)} \cdot fl_{(2,8,0)}^{(F,N,W)} \cdot fl_{(3,8,0)}^{(W,N,F)} \cdot {\bf B} \cdot br_{(5,8,0)}^{(E,N,B)} \cdot in_{(6,8,0)} \cdot fl_{(7,8,0)}^{(F,N,W)} \cdot fl_{(8,8,0)}^{(F,N,W)} \cdot \\ ba_{(9,8,0)}^{(F,N,W)}) \cdot ({\bf B}^3 \cdot fl_{(3,9,0)}^{(F,N,W)}) \cdot ({\bf B}^3 \cdot ba_{(3,10,0)}^{(F,N,W)}) \cdot ({\bf B}^3 \cdot di_{(3,11,0)}^{(F,N,W)}).
\end{eqnarray*}

Figure~\ref{binaryaddercubelet1} shows the set of initial conditions for each input and output for $w_{BA}$. The computation is asynchronous and every input updates its output changing the state of every distance cube. Finally, you can see a video of this robotic binary adder in action from \url{https://youtu.be/1qhpbPHnzPw}.

\section{Computations on Cubelets by Sleptsov nets}

Computing on cellular automata with propagating patterns can be simulated directly by Infinite Petri nets using technique studied in \cite{CAinfinitePN, UinfPN} and extended for cellular automata with generalized neighborhood \cite{genNeiCA}. Here we formalize Cubelets computing via modeling by Petri and Sleptsov nets the composed Cubelets robots for logical gates. Petri nets is a known tool for specification and verification concurrent processes \cite{verificationIPN}. Their generalization based on multiple transition firing, called a Sleptsov net and capable of fast computations \cite{SleptsovNet}, represents a graphical language for fine granulation massively parallel computing. We simulate Cubelets robots by Sletptsov nets to verify their behavior. 

\begin{figure}
\begin{center}
\subfigure[]{\scalebox{0.5}{\includegraphics{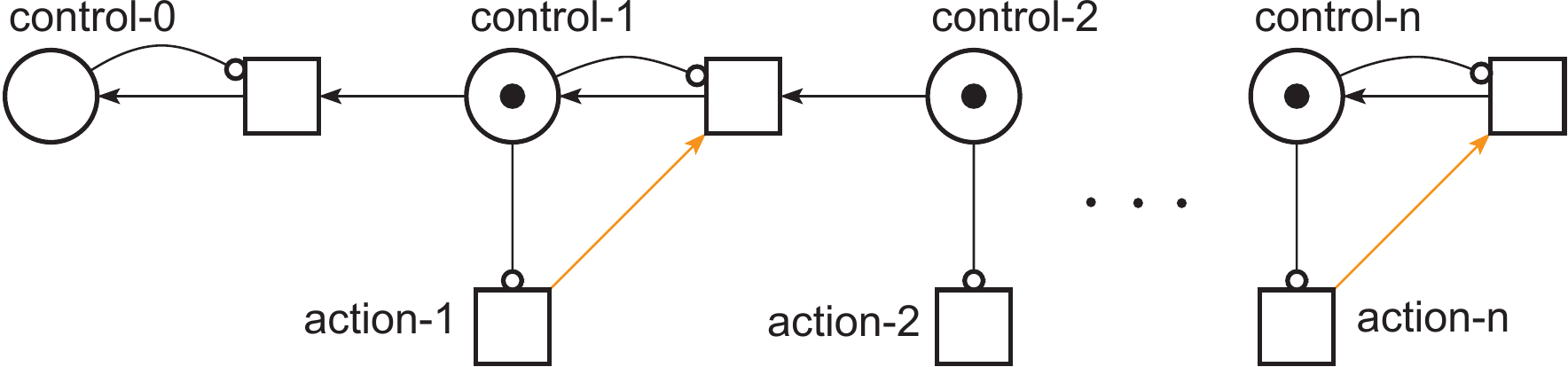}}} \hspace{0.6cm}
\subfigure[]{\scalebox{0.5}{\includegraphics{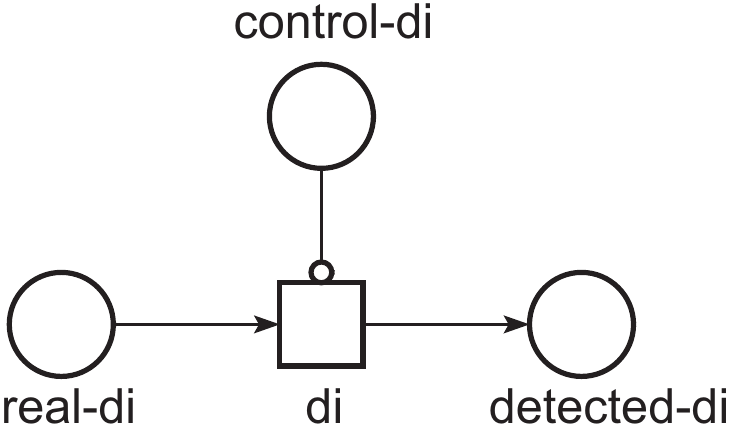}}} \hspace{0.7cm}
\subfigure[]{\scalebox{0.5}{\includegraphics{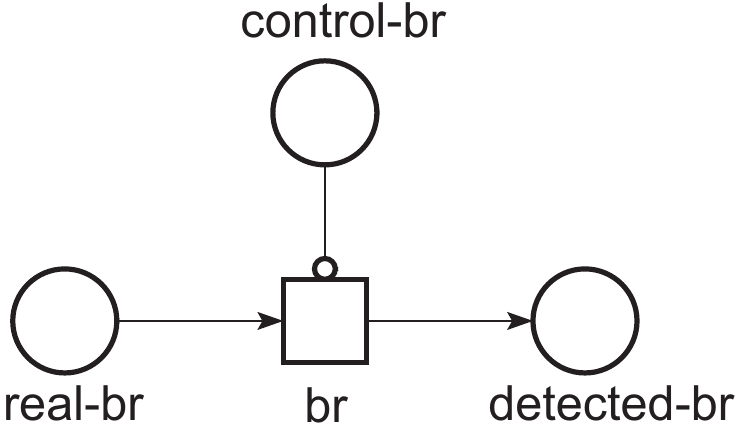}}} \hspace{0.9cm}
\subfigure[]{\scalebox{0.5}{\includegraphics{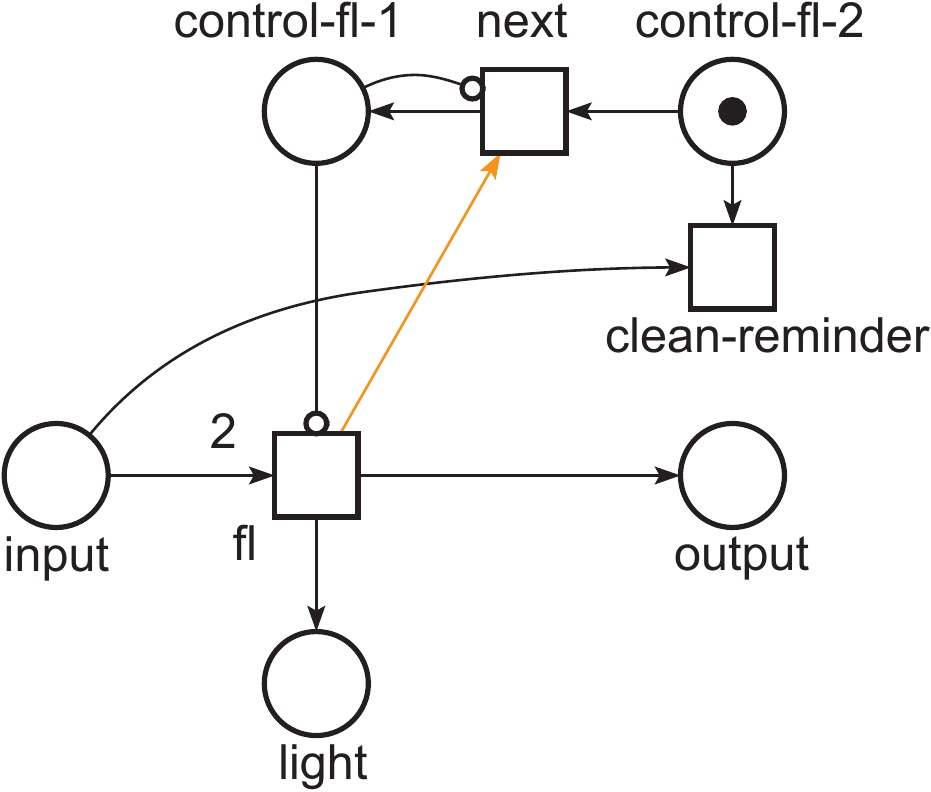}}} 
\subfigure[]{\scalebox{0.5}{\includegraphics{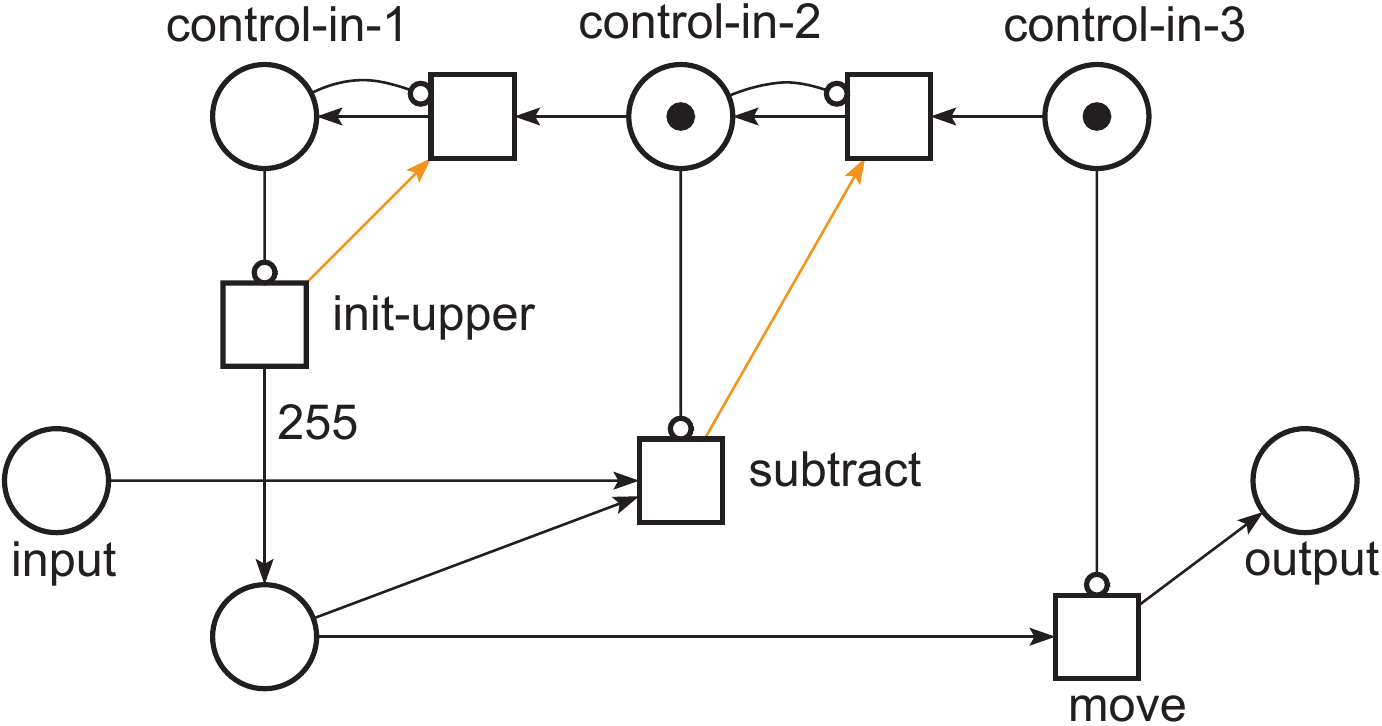}}}
\end{center}
\caption{Simulating Cubelets by Sleptsov nets: (a) battery \textit{ba}, (b) distance sensor \textit{di}, (c) brightness sensor \textit{br}, (d) flashlight \textit{fl}, (e) inverse \textit{in}.}
\label{fig-sn-cub}
\end{figure}

Cubelets behavior have been specified using a great deal of intuitive perception. This especially concerns assembled robots. Even in a simple case of linear connection it is nontrivial process of signals transmission when a battery is attached in the middle of line. When Cubelets are attached to many or even to all facets of a cube, a special technique is required to completely formalize behavior of single Cubelets and assembled robots. When we use Cubelets as a game, probably it is not so significant. Though when we use Cubelet robots as prototypes for real-life robots, especially designed for critical applications, the task of proving correctness of Cubelet robots behavior becomes significant. In the present section, a solution for a restricted class of Cubelet robots which represent computations, is presented. The idea consists in simulating Cubelet robots by Sleptsov nets.

In Section~\ref{cubeletscomputing}, to represent Cubelets computations, only configurations on plane were considered and only the following classes of robots have been employed: \textit{ba, di, fl, br, in}, and the blank cube {\bf B}. Thus only these Cubelets are simulated by Sleptsov nets and we consider two-dimensional configurations only. Note that Cubelet robots which simulate computations are unmovable. Though we believe that the technique can be generalized on moving Cubelet three-dimensional robots which use all the specified classes of cubes. Our motivation is a formal way of proving that the constructed robot implements a definite function. For the considered configurations of the majority gates and adders, the simplest proof technique is the exhaustive search represented by truth table. A sequence of experiments with Cubelets organized in complete accordance with the truth table proves that the robot functioning is correct. We use the considered circuits as tests for our Sleptsov net verification technique. It could be useful for bigger constructs as well to justify the composition rules when definite Cubelet robots are applied as components to compose bigger constructs. 

\begin{figure}[th]
\begin{center}
\includegraphics[width=0.95\textwidth]{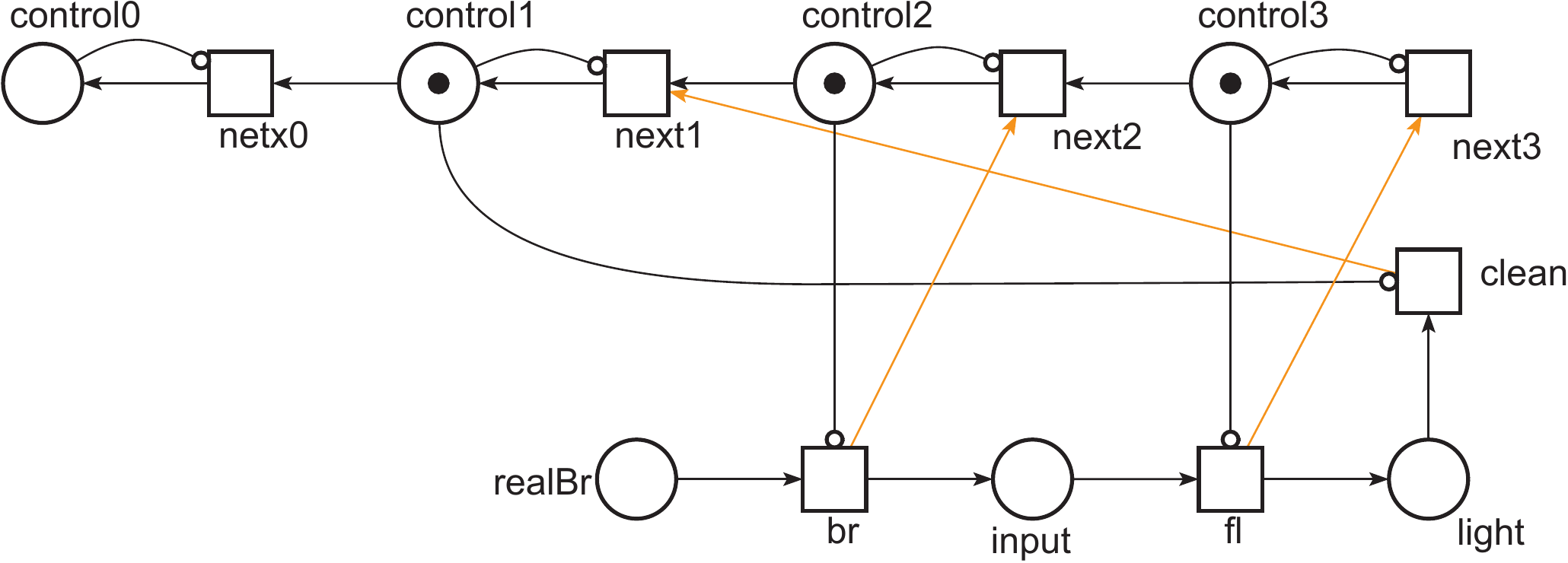}
\end{center}
\caption{Sleptsov net model of lamp Cubelets robot represented in Fig.~\ref{CubeletsLineSimple}b.}
\label{fig-lamp-sn}
\end{figure}

As it was mentioned in previous sections, a threshold technique is applied to obtain binary numbers from one byte counters which represent values of distance and brightness. It seems that visual observation does not allow to tell the brightness difference of a single unit that represents an additional point of our motivation.

Cubelets are based on continuous processes of electrical signals transmission though the signal levels are represented by integer numbers in the range $0\ldots255$. For modeling computations, the switching processes are of interest which are induced by change of input signals -- putting of removing some things from the distance detector \textit{di}. As it is described in \cite{SleptsovNet}, we use inverse control flow which is represented by attached battery. Each change of input is simulated by \textit{reset} procedure which consists in sequential implementation of cleaning computed data \textit{clean} and complete recalculation of output \textit{culc}. 

\begin{figure}
\begin{center}
\subfigure[]{\scalebox{0.277}{\includegraphics{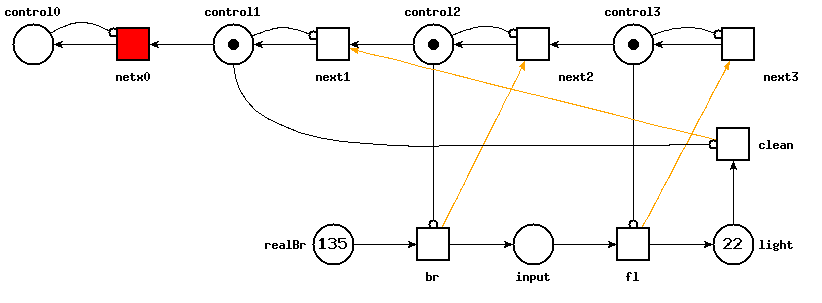}}} 
\subfigure[]{\scalebox{0.277}{\includegraphics{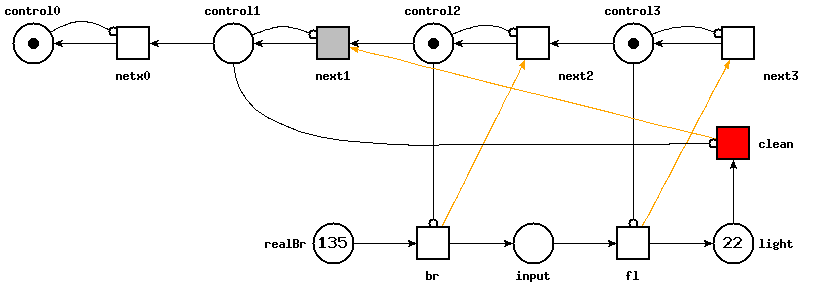}}} 
\subfigure[]{\scalebox{0.277}{\includegraphics{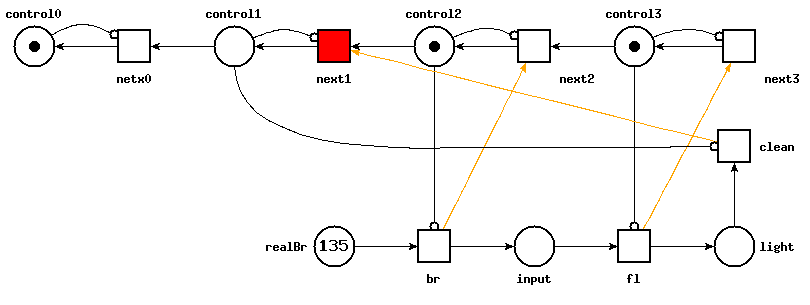}}} 
\subfigure[]{\scalebox{0.277}{\includegraphics{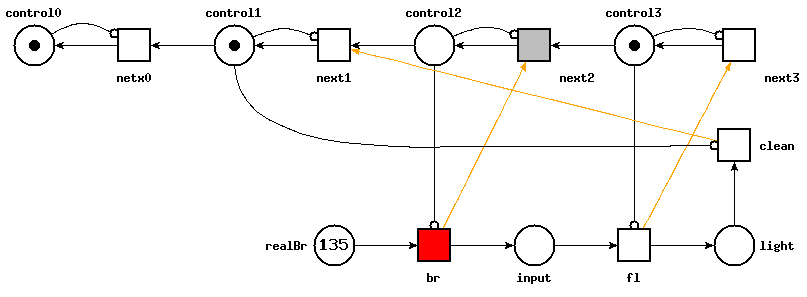}}} 
\subfigure[]{\scalebox{0.277}{\includegraphics{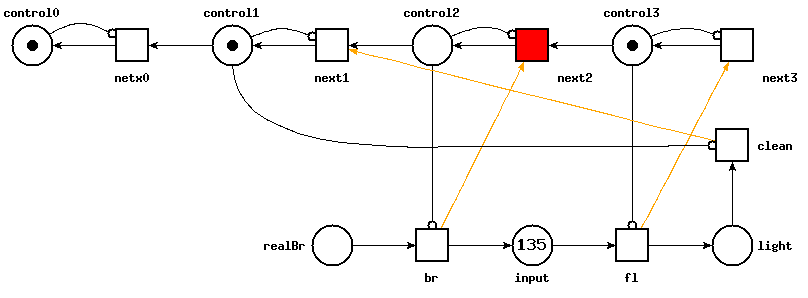}}} 
\subfigure[]{\scalebox{0.277}{\includegraphics{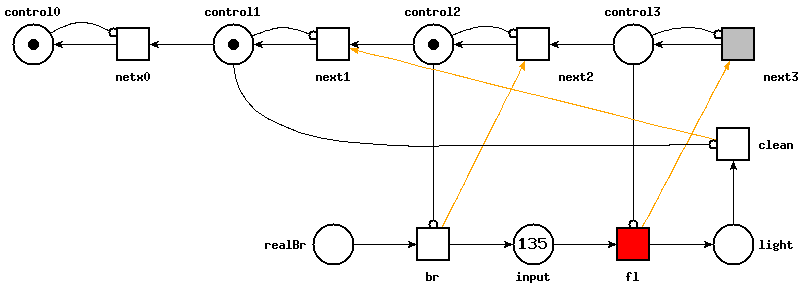}}} 
\subfigure[]{\scalebox{0.277}{\includegraphics{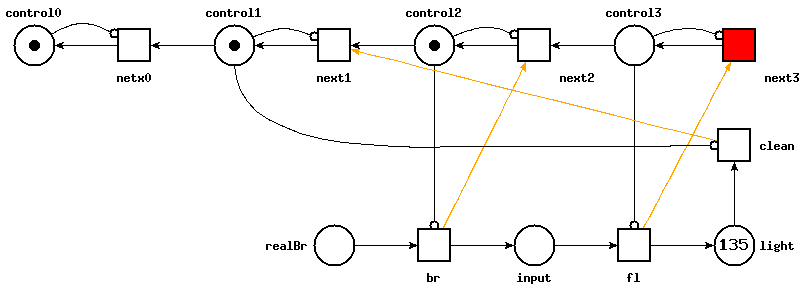}}} 
\subfigure[]{\scalebox{0.277}{\includegraphics{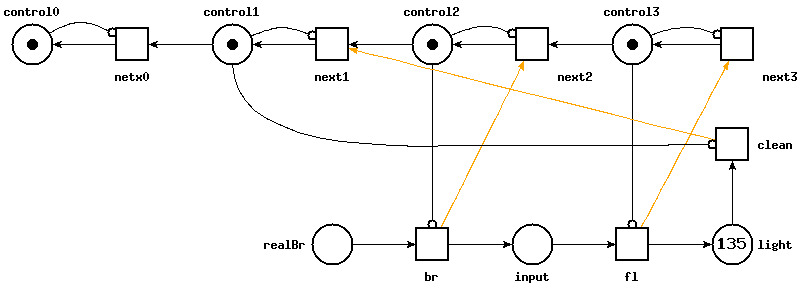}}} 
\end{center}
\caption{Simulating Sleptsov net model of lamp Cubelets robot; the only permitted transition firing sequence is $\sigma = next0 \cdot clean \cdot next1 \cdot br \cdot next2 \cdot fl \cdot next3$: (a) initial state, (b) \textit{next0} fired, (c) \textit{clean} fired, (d) \textit{next1} fired, (e) \textit{br} fired, (f) \textit{next2} fired, (g) \textit{fl} fired, (h) \textit{next3} fired -- final marking.}
\label{fig-lamp-sim}
\end{figure}

Sleptsov net models of employed Cubelets are shown in Fig.~\ref{fig-sn-cub}. The battery model $ba$ is represented by spreading reverse control flow via a sequence of places and transitions $control_i$ which enable the corresponding elements $action_i$. The reverse control flow means moving zero marking provided by combination  of regular and inhibitor arcs. Using reverse control flow allows us firing transition in the maximal multiplicity moving all the tokens in a single tact. Detectors \textit{di} and \textit{br} are represented by a single transition each moving a real value into detected value via using the corresponding transition. A flashlight Cubelet attached into a line is represented by the corresponding transition \textit{fl} which divides the input energy into emitted light energy represented by place \textit{light} and the output energy represented by place \textit{output}. In Fig.~\ref{fig-sn-cub}d we use equal proportion when of 2 input units of energy 1 is emitted as light and 1 is transmitted as output; any proportion $x+y$ can be applied, and extra transition $clean-reminder$ zeroes the reminder of division by $x+y$ (by $2=1+1$ in the figure). We interpret inverse Cubelet in multi-valued (discretized continuous or fuzzy logic) logic sense \cite{continL, fuzzyL} $x =255-x$ as shown in Fig.~\ref{fig-sn-cub}e. Thus, at first, we initialize the upper bound in place \textit{upper} by transition \textit{initUpper}, then we subtract the input value from the upper bound by transition \textit{subtract} and finally we move the obtained result by transition \textit{move} into place \textit{inverse}.

\begin{figure}[th]
\begin{center}
\includegraphics[width=0.99\textwidth]{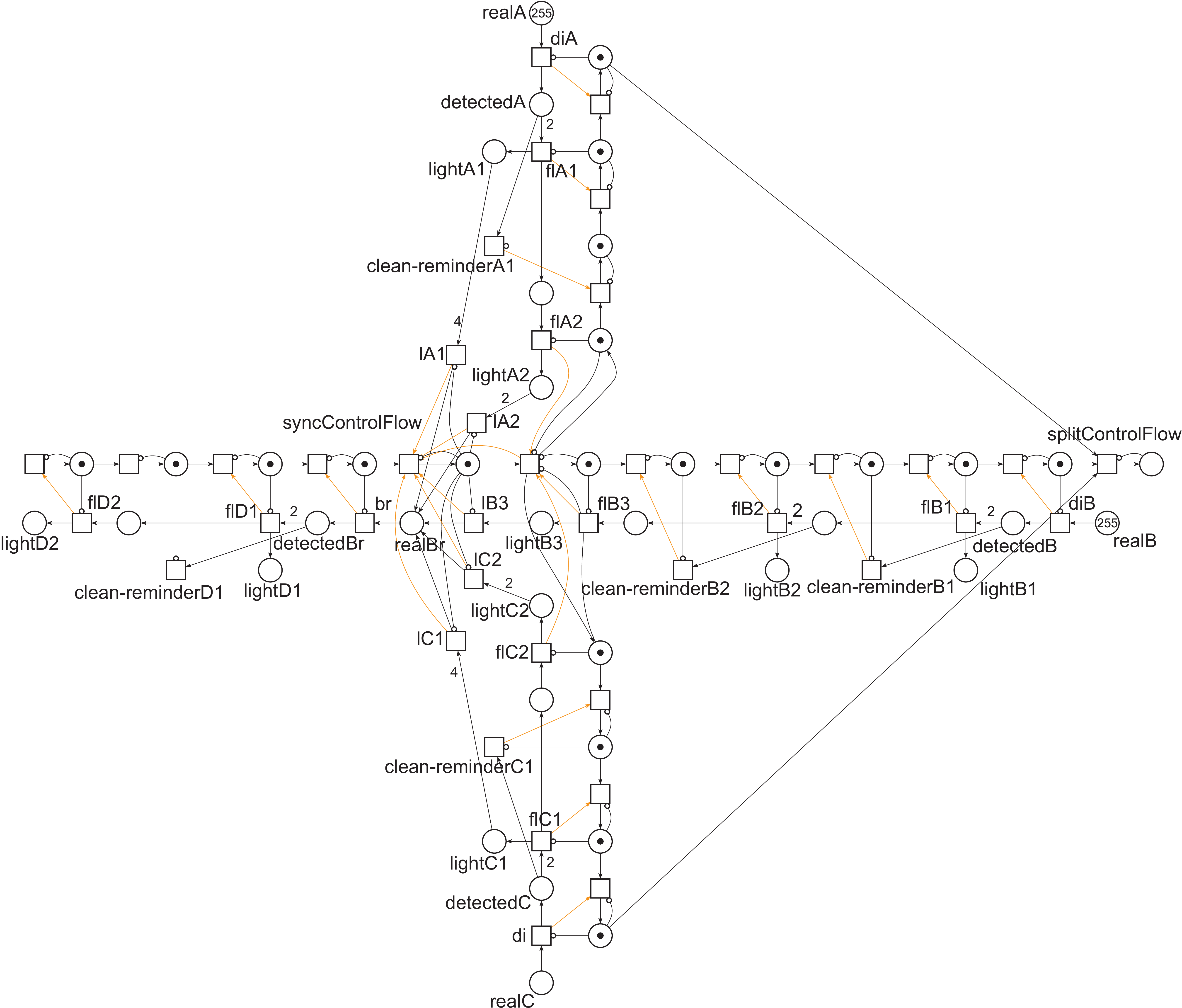}
\end{center}
\caption{Sleptsov net model of the {\sc majority} gate represented in Fig.~\ref{CubeletsMAJgate}. Cleaning part has been omitted for simplicity. For the chosen arc weights specifying the light spreading, the threshold equals 78: if $lightD2 \geq 78$ then $D=1$ else $D=0$. Initial marking $realA=255$, $realB=255$ corresponds to the binary cortege $A=1, B=1, C=0$.}
\label{fig-maj-gate-sn}
\end{figure}

Let us compose a Sleptsov net model of lamp robot $w = br \cdot ba \cdot fl$ which two states are illustrated in Fig.~\ref{CubeletsLineSimple}a and Fig.~\ref{CubeletsLineSimple}b using distance sensor $di$ instead of the brightness sensor $br$. The model is shown in Fig.~\ref{fig-lamp-sn}. We attached the brightness sensor $br$ because the Cubelets combination for spreading and sensoring light is applied further in composition of the majority gate model shown in Fig.~\ref{fig-maj-gate-sn}. The model in Fig.~\ref{fig-lamp-sn} represents a composition of the reverse control flow of battery $ba$ and two more Cubelets for the brightness sensor $br$ and flashlight $fl$ connected sequentially via place {\it input}. Note that since the flashlight Cubelet \textit{fl} is the last one in a line, it does not transmit energy further but consumes it entirely to produce light of the highest brightness, the corresponding Sleptsov net component is truncated, arc weights modified. To explain how the model works, we represent a sequence of Figs.~\ref{fig-lamp-sim}a-h which contain the initial and all the intermediate markings obtained after each transition firing. We suppose that the previous sensored value was 22 that corresponds to the marking of place \textit{light} in Fig.~\ref{fig-lamp-sim}a and the current sensored value is 135 that corresponds to the marking of place \textit{realBr}. The transition firing sequence $\sigma = next0 \cdot clean \cdot next1 \cdot br \cdot next2 \cdot fl \cdot next3$ resulting in the final marking shown in Fig.~\ref{fig-lamp-sim}h is obtained during computer simulation though using the technique studied in paper \cite{SleptsovNet} we can prove this fact in a formal way. Let us illustrate peculiarity of Sleptsov net behavior compared with Petri net: firing transition $br$, permitted in Fig.~\ref{fig-lamp-sim}d, moves all 135 tokens from place {\it realBR} to place {\it input} as shown in Fig.~\ref{fig-lamp-sim}e while only one token is moved at a step by the corresponding Petri net.

Now we compose Sleptsov net model of majority gate $w_{MAJgate}$ shown in Fig.~\ref{CubeletsMAJgate}. It consist of two parts: input $w_{MAJgateI}$ and output $w_{MAJgateO}$ which are not attached to each other. The connection between two parts is established via light emitted by three branches of the input part. The output part detects the light brightness, only flashlight emitted by more than one input branch is enough to switch the gate into true state (logical unit). The model is represented in Fig.~\ref{fig-maj-gate-sn}. Its assembling of the cubes models shown in Fig.~\ref{fig-sn-cub} is considered a routine job according to the technique described when constructing model of the lamp robot shown in Fig.~\ref{fig-lamp-sn}. The model shape replicates the shape of Cubelets robot in Fig.~\ref{CubeletsMAJgate}. Three distance sensors {\it diA}, {\it diB}, {\it diC} model the input  data while the output data is represented by flashlight {\it lightD2}. The peculiarity of the control consists in splitting the control flow by transition $splitControlFlow$ into three subflows for $A$, $B$, and $C$, respectively, and then synchronizing them by transition $syncControlFlow$ to compute the result $D$. Within Cubelets robot model (Fig.~\ref{CubeletsMAJgate}) there is a blank cube in the center separating the input and the output parts of the majority gates. This blank cube is represented by rather sophisticated tangle of Sleptsov net including transition $syncControlFlow$ for synchronization of control flow and transitions which simulate laws of light spreading $lA1$, $lA2$, $lC1$, $lC2$, $lB3$ and incidental arcs. We assume that flashlights $flB1$ and $flB2$ influence the total brightness insignificantly and their influence on the brightness sensor was not considered. If required, it can be done in the same way as for flashlights of branches $A$ and $C$. The specified set of arc weights produces brightness of the $lightD2$ flashlight for input values of distance detectors sensed on $realA$, $realB$, $realC$ represented with Tab.~\ref{tab-maj-gate}. Choosing a threshold, for instance, equal to 78, allows us to implement logical majority gates.

\begin{table}[th]
\caption{Table of simulated resulting brightness depending on sensed distance.}
\label{tab-maj-gate}
\centering
\begin{tabular}{|c|c|c|c|}
\hline
$realA$ & $realB$ & $realC$ & $lightD2$ \\
\hline \hline
0 & 0 & 0 & 0  \\
0 & 0 & 255 & 47  \\
0 & 255 & 0 & 31  \\
0 & 255 & 255 & 78  \\
255 & 0 & 0 & 47  \\
255 & 0 & 255 & 94  \\
255 & 255 & 0 & 78  \\
255 & 255 & 255 & 125  \\
\hline
\end{tabular}
\end{table}

Thus a technique of modeling Cubelets computations by Sleptsov nets has been developed, the advantage of modeling by Sleptsov nets consists in formal methods of the model behavior verification developed for Petri and Sleptsov nets \cite{verificationIPN,SleptsovNet}.

To verify the Cubelets robots behavior, it was offered to simulate Cubelets robots by Sleptsov nets \cite{SleptsovNet}. Only models of Cubelets, involved in simulation of Cubelets computing, have been constructed and principles of their assembling have been studied on an example of modeling majority gates by Sleptsov nets.

In the present paper, propagation of signals represented by cellular automata Life rule $B2/S2345$ \cite{martinez2010majority} was applied to construct logical gated and assemble an adder. An alternative way consists in modeling maltivalue logic operations directly using operations of continuous (fuzzy) logic \cite{continL, fuzzyL} represented by \textit{min, max, inv} Cubelets.

\section{Simple cooperating robots - perspectives and possibilities}

The possible future research on the robotic field like with Cubelets can be focused also into swarm domain, i.e. the set of robots is regarded as the swarm, solving the joined problem. As already simply demonstrated in the paper. Over the years, science and technology have been developing strongly, getting many outstanding achievements \cite{evenson2019science}, especially in the field of artificial intelligence (AI) \cite{haenlein2019brief}, where swarm intelligence optimization algorithm (SA) is an integral part \cite{del2019bio}, such as particle swarm optimization \cite{phung2017enhanced,ghamry2017multiple}, artificial bee colony \cite{pan2017unmanned,tian2018real}, and genetic algorithm \cite{arantes2016hybrid,roberge2018fast}. Along with the general development of AI, the SA has made great leaps and bounds \cite{mavrovouniotis2017survey,wanka2019swarm}, and applied in many fields of science and technology such as \cite{bao2019obstacle} and \cite{diep2018movement}. One of the most important and common applications is path planning for autonomous robots on the ground, underwater and in the space.

The control of the unmanned aerial vehicles (UAVs) is often more complicated, requiring not only the speed of the algorithm, but also real-time accuracy. Many solutions have been proposed and successfully applied to path planning for UAVs such as sampling-based path planning for UAV collision avoidance \cite{lin2017sampling}, cooperative path planning with applications to target tracking and obstacle avoidance for multi-UAVs \cite{yao2016cooperative}, and grid-based coverage path planning with minimum energy over irregular-shaped areas with UAVs \cite{cabreira2019grid}. So contemporary research on the field of the UAVs deals also with SA. One of the possible extension of the SA in UAVs is in its fusion with artificial neural networks (ANN).

\subsection{Incorporate ANN and the robot }\label{sec:LeverageANN}
 Here we propose possibility, how to join ANN with swarm robots, as already tested on alternative swarm system control in \cite{truong2019neural}. The concept of Multi-Layer Perceptron (MLP) neural network can bee used to enhance the swarm robots control, communication and problem solution. More specifically, we can adopt MLP in two cases:

\begin{enumerate}
\item Each of the robot in the swarm will contain an MLP,
\item A number of individuals in the population will simulate the behaviour of the MLP to perform actions, i.e. only selected individuals will be used as the ANN body.
\end{enumerate}

\subsubsection{Each robot comprise a ANN} 
\begin{figure}[h]
\centering
\includegraphics[width=0.6\linewidth]{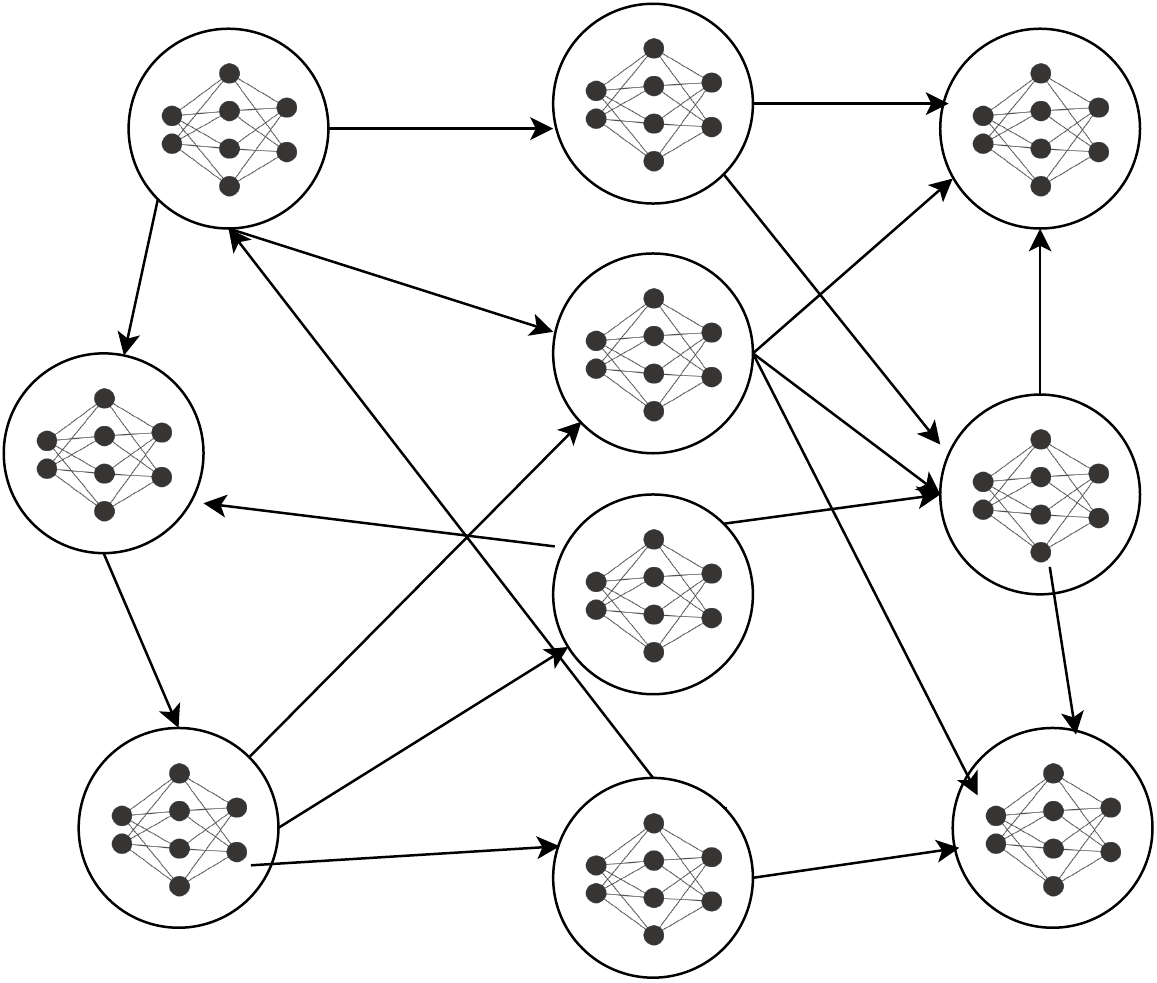}
\caption{Each robot in the swarm contain a MLP.}
\label{fig:nn1}
\end{figure}

~\\In this implementation, each individual in the swarm had its own MLP as shown in Fig. \ref{fig:nn1}.  

\begin{figure}[h]
\centering
\includegraphics[width=0.85\linewidth]{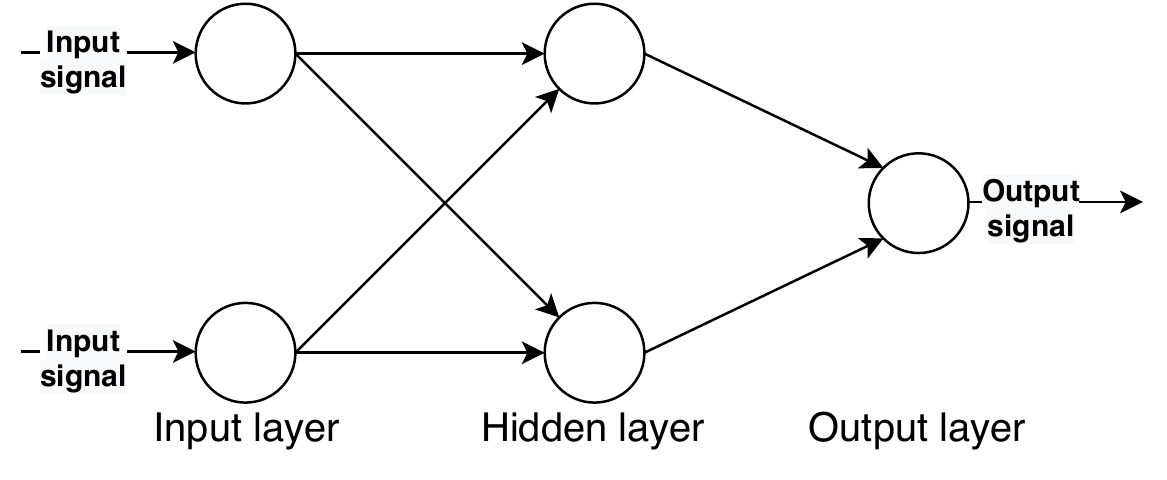}
\caption{Architecture of the MLP embedded in robot.}
\label{fig:mlp}
\end{figure}

The MLP implementation, in this case, is composed of three layers: input layer, hidden layer and output layer. The input can be regarded as the command for the swarm, in this case has MLP network's input layer, which contains a total of 2 neurons. The hidden layer consists of 2 neurons for network training, and an output layer consists of 1 neuron as it produces the result of whether to perform the tasks of the swarm or not. In terms of activation function, there are a variety of methods which could be used for training. Also, the topology of the network can be various and easily reconfigurable by uploading of the new one. In the paper \cite{truong2019neural}, for the purpose of binary classification, the logistic sigmoid function is utilised. Figure \ref{fig:mlp} shows the architecture of the MLP integrated into this prototype.

\begin{figure}[h!]
\centering
\includegraphics[width=0.55\linewidth]{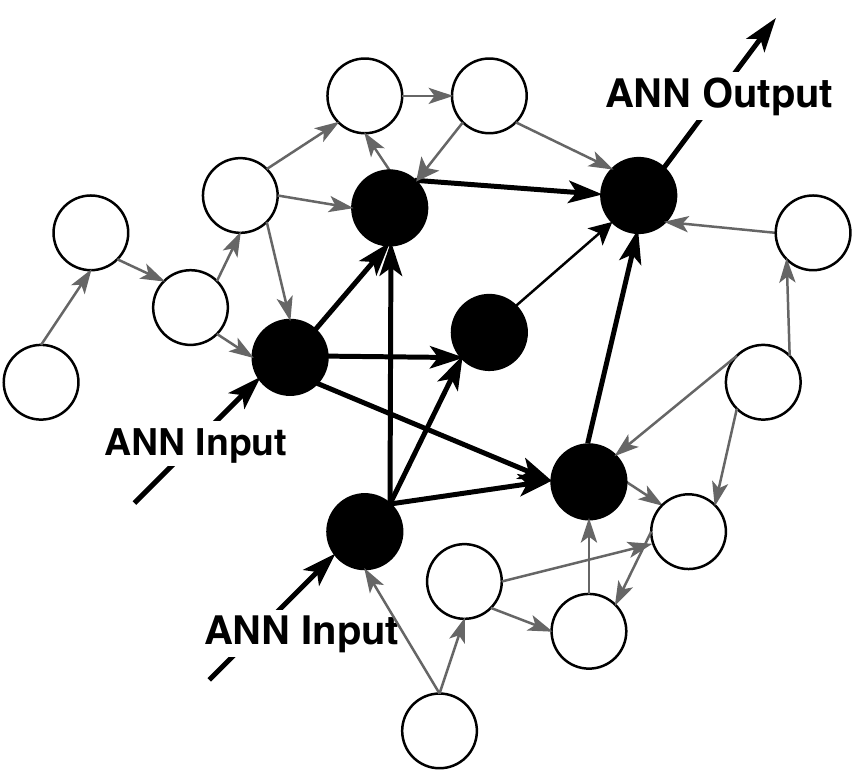}
\caption{Robots in swarm simulate the ANN working mechanism.}
\label{fig:nn2}
\end{figure}

The back-propagation algorithm \cite{rumelhart1986learning} was used to train the MLP in the proposed approach. Back-propagation is short for ``backward propagation of errors.'' It is a supervised training algorithm in the multilayer feed-forward networks using gradient descent \cite{haykin2010neural}. The dataset used for training is collected from system files. After training the model, the optimized network weights are integrated into the swarm. When the swarm executes, these weights are used on the embedded MLP to make the computation. Then, the MLP makes the ``trigger conditions'' to perform the execution, swarm entities, i.e. robots in this proposed case, are trained to perform system searches for finding a suitable target. The ANN then generates signals for conducting a task. Subsequently, reverse engineers analysts can hardly reverse the ANN to work out the target specifications, that can be beneficial if the swarm perform secret tasks.

Thus, swarm can use an ANN model, which is a black box, instead of a traditional {\sf if-then} command line to camouflage its trigger condition.

\subsubsection{Robot act as a node of an ANN}
This kind of implementation uses several robots in the swarm to act as input layers, hidden layers and output layers of an MLP. Figure \ref{fig:nn2} illustrates this idea. As shown in this figure, only some individuals in the swarm (black coloured) act as nodes in the MLP network. More precisely, two individuals are utilised as input nodes to receive signals, three for hidden nodes and one for the output node.

The MLP is also trained to obtain the optimised weights in the same manner as mentioned above. The difference is the weight allocated on individuals that act as nodes in the MLP network. In other words, each node comprises/carry its own weights.

When the robot executes, the swarm simulates the working mechanism of an MLP network. More specifically, the signal values propagated from the ``inputs robot,'' through the connection to the ``hidden robot,'' and then onward through more connection to the ``output robot.'' Following this strategy, the aim of the robot could be distributed, and it is hard to reverse engineer the robot strategy.

\section{Discussion}

Using modular robots Cubelets as a metaphor for swarms of very simple agents --- molecules, nanoparticles, crystals --- we demonstrated how the computing circuits can be designed, developed and executed. A binary full adder has been also designed as an illustrious way to demonstrate a computational power of robotic ensembles.

An advantage to use Cubelets robots is that you can work with other variants. 

\begin{itemize}
\item We can handle propagating signals using rotating Cubelets (signal value is a spin position),
\item We can handle propagating signals using speaker Cubelets (signal value is a sound wave),
\item We can handle propagating signals using thermal Cubelets (signal value is a thermal gradient).
\end{itemize}

Another contribution in the paper is that Cubelets robots have been simulated by Sletptsov nets to verify their behavior and this way we can construct more complex networks to design another non-trivial robots. Also the possibility of the next level of the robots, cooperating as the swarm, is outlined here. Thus cooperating agents (robots, molecules, nanoparticles, $\ldots$) can not only perform mathematical operations by simple interactions, but also behave as the more complex system exhibiting kind of adaptivity and plasticity, when the principles of the swarm intelligence and/or ANN are used. That can rise the computational possibility of the simple cooperating robots (Cubelets --- using T-bugs\footnote{\url{https://youtu.be/1kzht-VDU5w}}, chipibots (hydrabots)\footnote{It is a work in progress in our labs. It is a new robot based Cubelets simulating the move of baby squids and when they evolve self-assemble hydrabots.}, kilobots, $\ldots$) on higher level. Further work could be focused on developing learning devices in swarms of Cubelets and implementing image recognition and pattern classification tasks.

Some limitations and problems during these implementations were:

\begin{itemize}
\item If we increase the number of Cubelets on these designs hence the connectivity between cubes is most difficult because some of them lost the adequate connection. Although they could be fixed with some brick adapters. 
\item We have tested some three-dimensional designs and although they are totally viable and powerful the problem to ``fly'' Cubelets connected in the three-dimensional space is that the gravity difficult hardy its connectivity and stability.
\end{itemize}



\begin{thebibliography}{99}

\bibitem{arantes2016hybrid} Arantes, M.D.S., Arantes, J.D.S., Toledo, C.F.M. \& Williams, B.C. (2016). A hybrid multi-population genetic algorithm for UAV path planning. In: {\em Proceedings of the Genetic and Evolutionary Computation Conference 2016}, (pp. 853--860).

\bibitem{adamatzky2002collision} Adamatzky, A. (Ed.) (2002). {\em Collision-based computing}. Springer, London.

\bibitem{RDC} Adamatzky, A., Costello, B.D.L. \&  Asai, T. (2005). {\em Reaction-diffusion computers}. Elsevier.

\bibitem{adamatzky2009hot} Adamatzky, A. (2009). Hot ice computer, {\em Physics Letters A} {\bf 374(2)} 264--271.

\bibitem{adamatzky2010game} Adamatzky, A. (2010). {\em Game of life cellular automata}. London: Springer.

\bibitem{adamatzky2016advances} Adamatzky, A. (Ed.). (2016). {\em Advances in Unconventional Computing: Volume I Theory}. Springer.

\bibitem{adamatzkyAdvPhysarum} Adamatzky, A. (Ed.). (2016). Advances in Physarum machines: Sensing and computing with slime mould (Vol. 21). Springer.

\bibitem{adamatzky2017plantgates} Adamatzky, A., Sirakoulis, G.C., Mart{\'i}nez, G.J., Baluska, F. \&  Mancuso, S. (2017). On plant roots logical gates. {\em BioSystems} {\bf 156} 40--45.

\bibitem{OnBuildings} Adamatzky, A., Szaciłowski, K., Konkoli, Z., Werner, L.C., Przyczyna, D. \& Sirakoulis, G.C. (2020). On buildings that compute. A proposal. In From Astrophysics to Unconventional Computation (pp. 311--335). Springer, Cham.

\bibitem{bao2019obstacle} Bao, D.Q. \& Zelinka, I. (2019). Obstacle avoidance for swarm robot based on self-organizing migrating algorithm. {\em Procedia Computer Science} {\bf 150} 425--432.

\bibitem{cabreira2019grid} Cabreira, T.M., Ferreira, P.R., Di Franco, C. \& Buttazzo, G.C. (2019). Grid-Based Coverage Path Planning With Minimum Energy Over Irregular-Shaped Areas With Uavs. In: {\em 2019 International Conference on Unmanned Aircraft Systems (ICUAS)} (pp. 758--767). IEEE.

\bibitem{codd1968cellular} Codd, E.F. (1968). {\em Cellular automata}. Academic Press.

\bibitem{dewdney1990column} Dewdney, A.K. (1990). Column ``Computer Recreations'': WireWorld. {\em Scientific American}, January.

\bibitem{del2019bio} Del Ser, J., Osaba, E., Molina, D., Yang, X.S., Salcedo-Sanz, S., Camacho, D., Das, S., Suganthan, P.N., Coello, C.A.C. \& Herrera, F. (2019). Bio-inspired computation: Where we stand and what's next. {\em Swarm and Evolutionary Computation} {\bf 48} 220--250.

\bibitem{diep2018movement} Diep, Q.B. \& Zelinka, I. (2018). The Movement of Swarm Robots in an Unknown Complex Environment. In: {\em International Conference on Advanced Engineering Theory and Applications} (pp. 949--959). Springer, Cham.

\bibitem{evenson2019science} Evenson, R. \& Ranis, G. (2019). Science and Technology: lessons for development policy. Routledge.

\bibitem{figueroa2019turing} Figueroa, R.Q., Zamorano, D.A., Mart{\'i}nez, G.J. \& Adamatzky, A. (2019). A {T}uring machine constructed with Cubelets robots. {\em Journal of Robotics, Networking and Artificial Life} {\bf 5(4)} 265--268.

\bibitem{fischer2017experimental} Fischer, T., Kewenig, M., Bozhko, D.A., Serga, A.A., Syvorotka, I.I., Ciubotaru, F., Adelmann, C., Hillebrands, B. \& Chumak, A.V. (2017). Experimental prototype of a spin-wave majority gate, {\em American Institute of Physics} {\bf 17(2)} 86--91.

\bibitem{ghamry2017multiple} Ghamry, K.A., Kamel, M.A. \& Zhang, Y. (2017). Multiple UAVs in forest fire fighting mission using particle swarm optimization. In: {\em 2017 International Conference on Unmanned Aircraft Systems (ICUAS)}, (pp. 1404--1409). IEEE.

\bibitem{griffeath2003new} Griffeath, D. \& Moore, C. (Eds.). (2003). {\em New constructions in cellular automata}. Oxford University Press.

\bibitem{snider1999quantum} Gregory, L.S., Orlov, A.O., Amlani, I., Bernstein, G.H., Lent, C.S., Merz, J.L. \& Porod, W. (1999). Quantum-Dot Cellular Automata: Line and Majority Logic Gate, {\em Japanese Journal of Applied Physics} {\bf 38} 7227--7229.

\bibitem{gunji2011} Gunji, Y.P., Nishiyama, Y. \&  Adamatzky, A. (2011). Robust soldier crab ball gate. In AIP Conference Proceedings (Vol. 1389, No. 1, pp. 995--998). American Institute of Physics.

\bibitem{haykin2010neural} Huang, G., Huang, G.B., Song, S. \& You, K. (2015). Trends in extreme learning machines: A review. {\em Neural Networks} {\bf 61} 32--48.

\bibitem{haenlein2019brief} Haenlein, M. \& Kaplan, A. (2019). A brief history of artificial intelligence: On the past, present, and future of artificial intelligence. {\em California Management Review} {\bf 61(4)} 5--14.

\bibitem{hutton2010codd} Hutton, T.J. (2010). Codd's self-replicating computer. {\em Artificial Life} {\bf 16(2)} 99--117.

\bibitem{lin2017sampling} Lin, Y. \& Saripalli, S. (2017). Sampling-based path planning for UAV collision avoidance. {\em IEEE Transactions on Intelligent Transportation Systems}, {\bf 18(11)} 3179--3192.

\bibitem{martinez2008logical} Mart{\'i}nez, G.J., Adamatzky, A. \& Costello, B.L. (2008). On logical gates in precipitating medium: cellular automaton model, {\em Physics Letters A} {\bf 1(48)} 1--5.

\bibitem{martinez2010computation} Mart{\'i}nez, G.J., Adamatzky, A., Morita, K. \& Margenstern, M. (2010). Computation with competing patterns in Life-like automaton, In: {\em Game of Life Automata}, A. Adamatzky (Ed.), Springer, chapter 27, pages 547--572.

\bibitem{martinez2018logical} Mart{\'i}nez, G.J., Adamatzky, A. \& Morita, K. (2018). Logical Gates via Gliders Collisions. In: {\em Reversibility and Universality} (pp. 199--220). Springer, Cham.

\bibitem{mazoyer1996computations} Mazoyer, J. (1996). Computations on one-dimensional cellular automata. {\em Annals of Mathematics and Artificial Intelligence} {\bf 16(1)} 285--309.

\bibitem{mcintosh1990programming} McIntosh, H.V. \& Cisneros, G. (1990). The programming languages REC and Convert. {\em ACM SIGPLAN Notices} {\bf 25(7)} 81--94.

\bibitem{mitchell1994evolving} Mitchell, M., Crutchfield, J.P. \& Hraber, P.T. (1994). Evolving cellular automata to perform computations: Mechanisms and impediments. {\em Physica D: Nonlinear Phenomena} {\bf 75(1-3)} 361--391.

\bibitem{minsky1967computation} Minsky, M.L. (1967). {\em Computation: Finite and Infinite Machines}. Englewood Cliffs: Prentice-Hall.

\bibitem{martinez2010majority} Mart{\'i}nez, G.J., Morita, K., Adamatzky, A. \& Margenstern, M. (2010). Majority adder implementation by competing patterns in Life-like rule $B2/S2345$, {\em Lecture Notes in Computer Science} {\bf 6079} 93--104.

\bibitem{martinez2018conservative} Mart{\'i}nez, G.J. \& Morita, K. (2018). Conservative Computing in a One-dimensional Cellular Automaton with Memory. {\em Journal of Cellular Automata} {\bf 13(4)} 325--346.

\bibitem{morita2001simple} Morita, K. (2001). A simple universal logic element and cellular automata for reversible computing. In: {\em International Conference on Machines, Computations, and Universality} (pp. 102--113). Springer, Berlin, Heidelberg.

\bibitem{mavrovouniotis2017survey} Mavrovouniotis, M., Li, C. \& Yang, S. (2017). A survey of swarm intelligence for dynamic optimization: Algorithms and applications. {\em Swarm and Evolutionary Computation} {\bf 33} 1--17.

\bibitem{pan2017unmanned} Pan, T.S., Dao, T.K. \& Pan, J.S. (2017). An unmanned aerial vehicle optimal route planning based on compact artificial bee colony. In: {\em Advances in Intelligent Information Hiding and Multimedia Signal Processing}, (pp. 361--369). Springer, Cham.

\bibitem{phung2017enhanced} Phung, M.D., Quach, C.H., Dinh, T.H. \& Ha, Q. (2017). Enhanced discrete particle swarm optimization path planning for UAV vision-based surface inspection. {\em Automation in Construction} {\bf 81} 25--33.

\bibitem{rumelhart1986learning} Rumelhart, D.E., Hinton, G.E. \& Williams, R.J. (1986). Learning representations by back-propagating errors. {\em Nature} {\bf 323(6088)} 533--536.

\bibitem{roberge2018fast} Roberge, V., Tarbouchi, M. \& Labont\'e, G. (2018). Fast genetic algorithm path planner for fixed-wing military UAV using GPU. {\em IEEE Transactions on Aerospace and Electronic Systems} {\bf 54(5)} 2105--2117.

\bibitem{schweikardt2008learning} Schweikardt, E. \& Gross, M.D. (2008). Learning about complexity with modular robots. In: {\em 2008 Second IEEE International Conference on Digital Game and Intelligent Toy Enhanced Learning} (pp. 116--123). IEEE.

\bibitem{schweikardt2011modular} Schweikardt, E. (2011). Modular robotics studio. In: {\em Proceedings of the fifth international conference on Tangible, embedded, and embodied interaction} (pp. 353--356). ACM.

\bibitem{tian2018real} Tian, G., Zhang, L., Bai, X. \& Wang, B. (2018). Real-time Dynamic Track Planning of Multi-UAV Formation Based on Improved Artificial Bee Colony Algorithm. In: {\em 2018 37th Chinese Control Conference (CCC)}, (pp. 10055--10060). IEEE.

\bibitem{truong2019neural} Truong, T.C., Zelinka, I. \& Senkerik, R. (2019). Neural Swarm Virus. In: {\em Swarm, Evolutionary, and Memetic Computing and Fuzzy and Neural Computing} (pp. 122--134). Springer, Cham.

\bibitem{neumann1966theory} von Neumann, J. (1966). {\em Theory of Self-reproducing Automata} (edited and completed by A.W. Burks), University of Illinois Press, Urbana and London.

\bibitem{wanka2019swarm} Wanka, R. (2019). Swarm intelligence. {\em it-Information Technology} {\bf 61(4)} 157--158.

\bibitem{yao2016cooperative} Yao, P., Wang, H. \& Su, Z. (2016). Cooperative path planning with applications to target tracking and obstacle avoidance for multi-UAVs. {\em Aerospace science and Technology} {\bf 54} 10--22.

\bibitem{continL} Zaitsev, D.A., Sarbei, V.G. \& Sleptsov, A.I. (1998). Synthesis of continuous-valued logic functions defined in tabular form. {\em Cybernetics and Systems Analysis} {\bf 34(2)} 190--195.

\bibitem{verificationIPN} Zaitsev, D.A. (2013). Verification of Computing Grids with Special Edge Conditions by Infinite Petri Nets. {\em Automatic Control and Computer Sciences}  {\bf 47(7)} 403--412.

\bibitem{UinfPN} Zaitsev, D.A. (2015). Universality in Infinite Petri Nets. {\em Lecture Notes in Computer Science} {\bf 9288} 180--197.

\bibitem{genNeiCA} Zaitsev, D.A. (2017). A generalized neighborhood for cellular automata. {\em Theoretical Computer Science} {\bf 666} 21--35.

\bibitem{SleptsovNet} Zaitsev, D.A. (2017). Universal Sleptsov Net. {\em International Journal of Computer Mathematics} {\bf 94(12)} 2396--2408.

\bibitem{fuzzyL} Zaitsev D.A. (2017). A Toolbox for Fuzzy Logic Functions Synthesis on a Choice Table. In: {\em Eurosis: 15th Industrial Simulation Conference Polish Academy of Science}, Warsaw, pp. 11--16.

\bibitem{CAinfinitePN} Zaitsev, D.A. (2018). Simulating Cellular Automata by Infinite Petri Nets. {\em Journal of Cellular Automata} {\bf 13(1-2)} 121--144.


\end{thebibliography}
\end{document}